\documentclass[10pt,twocolumn,letterpaper]{article}

\usepackage[pagenumbers]{cvpr} 

\usepackage{amsmath,amssymb,amsthm}
\usepackage{bm}
\usepackage{graphicx}
\usepackage[font=small,labelfont=bf,skip=4pt]{caption}
\usepackage{subcaption}
\usepackage{booktabs}
\usepackage{multirow}
\usepackage{array}
\usepackage{enumitem}
\usepackage{xcolor}
\usepackage{algorithm}
\usepackage{algpseudocode}
\usepackage{tikz}
\usetikzlibrary{arrows.meta,positioning,fit,backgrounds,calc,patterns,shadows,shapes.geometric}

\theoremstyle{plain}

\newtheorem{proposition}{Proposition}

\theoremstyle{definition}

\providecommand{\R}{\mathbb{R}}

\newcommand{\Span}{\operatorname{span}}

\newcommand{\btau}{\bm{\tau}}
\newcommand{\btheta}{\bm{\theta}}
\newcommand{\mW}{\bm{W}}
\newcommand{\mU}{\bm{U}}
\newcommand{\mV}{\bm{V}}

\newcommand{\mB}{\bm{B}}

\newcommand{\mQ}{\bm{Q}}
\newcommand{\mZ}{\bm{Z}}
\newcommand{\bx}{\bm{x}}
\newcommand{\mI}{\bm{I}}
\newcommand{\mP}{\bm{P}}
\newcommand{\norm}[1]{\left\lVert #1\right\rVert}

\newcommand{\inner}[2]{\left\langle #1, #2\right\rangle}

\providecommand{\etal}{et~al.}

\newcommand{\ours}{\textsc{JNSP}}

\floatname{algorithm}{Algorithm}

\newcolumntype{C}[1]{>{\centering\arraybackslash}p{#1}}
\newcolumntype{L}[1]{>{\raggedright\arraybackslash}p{#1}}

\definecolor{oursblue}{RGB}{31,78,121}
\definecolor{basegrey}{RGB}{140,140,140}
\definecolor{baseorange}{RGB}{217,133,63}
\definecolor{basegreen}{RGB}{90,150,85}
\definecolor{basered}{RGB}{178,80,80}
\definecolor{tint}{RGB}{223,232,242}
\definecolor{tintb}{RGB}{245,231,213}

\definecolor{cvprblue}{rgb}{0.21,0.49,0.74}
\usepackage[pagebackref,breaklinks,colorlinks,allcolors=cvprblue]{hyperref}

\def\confName{CVPR}
\def\confYear{2027}

\title{Joint Null-Space Projection for Data-Free Model Merging}

\author{Tian Xia\\
{\tt\small xia.tian.t4@alumni.tohoku.ac.jp}
}

\makeatletter
\newcommand{\teaserimg}[2]{\csname jnsp@teaser@#2\endcsname{#1}}
\@namedef{jnsp@teaser@mnist}#1{\includegraphics[width=#1,height=#1]{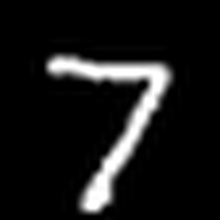}}
\@namedef{jnsp@teaser@eurosat}#1{\includegraphics[width=#1,height=#1]{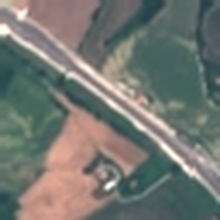}}
\@namedef{jnsp@teaser@gtsrb}#1{\includegraphics[width=#1,height=#1]{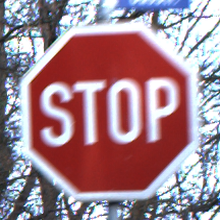}}
\@namedef{jnsp@teaser@svhn}#1{\includegraphics[width=#1,height=#1]{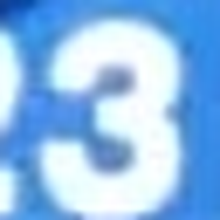}}
\@namedef{jnsp@teaser@dtd}#1{\includegraphics[width=#1,height=#1]{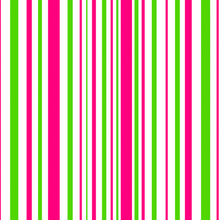}}
\@namedef{jnsp@teaser@resisc45}#1{\includegraphics[width=#1,height=#1]{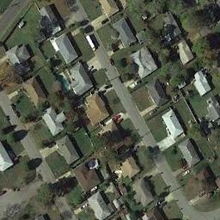}}
\@namedef{jnsp@teaser@cars}#1{\includegraphics[width=#1,height=#1]{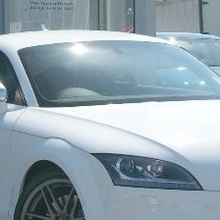}}
\@namedef{jnsp@teaser@sun397}#1{\includegraphics[width=#1,height=#1]{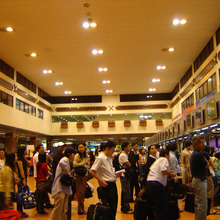}}
\makeatother
\begin{document}
\makeatletter
\let\jnsp@maketitle\@maketitle
\def\@maketitle{\jnsp@maketitle
\vspace*{2pt}
\begin{tikzpicture}[font=\footnotesize,>={Stealth[length=1.8mm]},x=1cm,y=1cm]

\definecolor{tA}{RGB}{31,78,121}
\definecolor{tB}{RGB}{217,133,63}
\definecolor{tC}{RGB}{90,150,85}
\definecolor{tD}{RGB}{178,80,80}
\definecolor{tE}{RGB}{64,140,150}
\definecolor{tF}{RGB}{122,96,155}
\definecolor{tG}{RGB}{198,163,66}
\definecolor{tH}{RGB}{96,96,116}

\foreach \x in {6.62,12.38}{
  \draw[gray!40,densely dotted,line width=0.5pt] (\x,0.62) -- (\x,4.92);}

\foreach \x/\n/\txt in {%
  0.20/1/{one backbone, $N$ experts},
  12.62/3/{sum once: one model, every task}}{
  \node[circle,fill=oursblue,text=white,inner sep=1.05pt,font=\scriptsize\bfseries] at (\x+0.10,4.62) {\n};
  \node[anchor=west,font=\scriptsize\bfseries,text=black!80] at (\x+0.30,4.62) {\txt};
}
\node[circle,fill=oursblue,text=white,inner sep=1.05pt,font=\scriptsize\bfseries] at (6.96,4.62) {2};
\node[anchor=west,align=left,font=\scriptsize\bfseries,text=black!80]
  at (7.16,4.62) {per tensor: keep only\\unclaimed directions};

\begin{scope}[shift={(-0.35,0)}]
\draw[gray!25,fill=gray!10,rounded corners=2.5pt] (0.74,1.62) rectangle (1.98,3.42);
\draw[tA!22,fill=tA!7,rounded corners=2.5pt]      (0.66,1.70) rectangle (1.90,3.50);
\draw[tA!50,fill=tA!11,rounded corners=2.5pt]     (0.58,1.78) rectangle (1.82,3.58);
\foreach \i in {0,1,2}\foreach \j in {0,1,2}{
  \draw[tA!40,fill=white,line width=0.3pt]
    ({0.69+0.29*\i},{2.78+0.27*\j}) rectangle ++(0.26,0.24);}
\foreach \i in {0,1,2,3}{
  \draw[tA!50,fill=white,line width=0.3pt,rounded corners=0.6pt]
    ({0.70+0.24*\i},2.30) rectangle ++(0.19,0.14);}
\node[font=\fontsize{4.6}{5.2}\selectfont,text=black!70,align=center] at (1.20,1.55) {CLIP ViT};

\foreach \y in {3.90,3.12,2.34,1.56}{
  \draw[->,gray!50,line width=0.5pt] (1.95,2.70) .. controls (2.55,2.70) .. (3.10,\y+0.24);}

\foreach \y/\cl/\cr in {3.54/tA/tE, 2.76/tB/tF, 1.98/tC/tG, 1.20/tD/tH}{
  \foreach \x/\col in {3.12/\cl,4.85/\cr}{
    \draw[gray!30,fill=white,rounded corners=1.8pt,line width=0.4pt,
      drop shadow={opacity=0.10,shadow xshift=0.2ex,shadow yshift=-0.2ex}]
      (\x,\y) rectangle ++(1.70,0.66);
    \fill[\col!80,rounded corners=1pt] (\x+0.02,\y+0.03) rectangle ++(0.05,0.60);}}

\foreach \x/\y/\img/\lab in {%
  3.12/3.54/mnist/MNIST, 3.12/2.76/eurosat/EuroSAT,
  3.12/1.98/gtsrb/GTSRB, 3.12/1.20/svhn/SVHN}{
  \node[anchor=south west,inner sep=0pt] at (\x+0.05,\y+0.10)
    {\teaserimg{0.46cm}{\img}};
  \node[font=\fontsize{3.9}{4.6}\selectfont,anchor=west] at (\x+0.53,\y+0.33) {\lab};}

\foreach \x/\y/\img/\lab in {%
  4.85/3.54/dtd/DTD, 4.85/2.76/resisc45/RESISC45,
  4.85/1.98/cars/Cars, 4.85/1.20/sun397/SUN397}{
  \node[anchor=south west,inner sep=0pt] at (\x+0.05,\y+0.10)
    {\teaserimg{0.46cm}{\img}};
  \node[font=\fontsize{3.9}{4.6}\selectfont,anchor=west] at (\x+0.53,\y+0.33) {\lab};}

\end{scope}

\shade[inner color=gray!3,outer color=gray!17] (7.45,2.42) circle (0.88);
\draw[gray!45] (7.45,2.42) ellipse [x radius=0.88,y radius=0.26];
\draw[gray!32] (7.45,2.42) ellipse [x radius=0.27,y radius=0.88];
\draw[tA,fill=tA,fill opacity=0.30,line width=0.5pt,rotate around={18:(7.34,2.72)}] (7.34,2.72) ellipse [x radius=0.40,y radius=0.21];
\draw[tB,fill=tB,fill opacity=0.30,line width=0.5pt,rotate around={-16:(7.62,2.56)}] (7.62,2.56) ellipse [x radius=0.40,y radius=0.21];
\draw[tC,fill=tC,fill opacity=0.30,line width=0.5pt,rotate around={8:(7.54,2.20)}] (7.54,2.20) ellipse [x radius=0.38,y radius=0.20];
\fill[basered,pattern=north east lines,pattern color=basered!80,opacity=0.50,rotate around={5:(7.49,2.58)}]
  (7.49,2.58) ellipse [x radius=0.15,y radius=0.10];
\node[font=\fontsize{5.2}{6.0}\selectfont,text=black!72,anchor=north,align=center] at (7.45,1.34)
  {caps overlap\\$\Rightarrow$ interference};

\shade[inner color=gray!3,outer color=gray!17] (10.3,2.42) circle (0.88);
\draw[gray!45] (10.3,2.42) ellipse [x radius=0.88,y radius=0.26];
\draw[gray!32] (10.3,2.42) ellipse [x radius=0.27,y radius=0.88];
\draw[tA!25,fill=tA,fill opacity=0.06,line width=0.35pt,dashed,rotate around={18:(10.19,2.72)}] (10.19,2.72) ellipse [x radius=0.40,y radius=0.21];
\draw[tB!25,fill=tB,fill opacity=0.06,line width=0.35pt,dashed,rotate around={-16:(10.47,2.56)}] (10.47,2.56) ellipse [x radius=0.40,y radius=0.21];
\draw[tC!25,fill=tC,fill opacity=0.06,line width=0.35pt,dashed,rotate around={8:(10.39,2.20)}] (10.39,2.20) ellipse [x radius=0.38,y radius=0.20];
\draw[tA,fill=tA,fill opacity=0.50,line width=0.5pt,rotate around={18:(10.19,2.72)}] (10.19,2.72) ellipse [x radius=0.40,y radius=0.21];
\draw[tA,fill=tA,fill opacity=0.50,line width=0.5pt,rotate around={-16:(10.47,2.56)}] (10.47,2.56) ellipse [x radius=0.40,y radius=0.21];
\draw[tA,fill=tA,fill opacity=0.50,line width=0.5pt,rotate around={8:(10.39,2.20)}] (10.39,2.20) ellipse [x radius=0.38,y radius=0.20];
\fill[white,opacity=0.80,rotate around={5:(10.34,2.58)}] (10.34,2.58) ellipse [x radius=0.15,y radius=0.10];
\node[font=\fontsize{5.2}{6.0}\selectfont,text=black!72,anchor=north,align=center] at (10.3,1.34)
  {only unclaimed\\directions stay};

\draw[->,oursblue,line width=0.7pt] (8.4,2.42) -- (9.3,2.42);
\node[font=\fontsize{5.0}{5.8}\selectfont,text=oursblue,align=center,anchor=north] at (8.85,2.34)
  {project off\\the others};

\foreach \y/\col in {3.35/tA, 2.42/tB, 1.49/tC}{
  \draw[\col!60,fill=\col!14,rounded corners=1.2pt,line width=0.35pt] (12.84,\y) rectangle ++(0.40,0.24);
  \draw[->,\col!75,line width=0.5pt] (13.26,\y+0.12) .. controls (13.48,\y+0.12) and (13.55,2.42) .. (13.78,2.42);}
\draw[gray!22,fill=gray!10,rounded corners=3pt] (14.02,1.94) rectangle (16.42,3.54);
\draw[oursblue!22,fill=oursblue!7,rounded corners=3pt] (13.94,1.86) rectangle (16.34,3.46);
\draw[oursblue!60,fill=white,rounded corners=3pt] (13.86,1.78) rectangle (16.26,3.38);
\foreach \i/\img/\col in {%
  0/mnist/tA, 1/eurosat/tB, 2/gtsrb/tC, 3/svhn/tD,
  4/dtd/tE, 5/resisc45/tF, 6/cars/tG, 7/sun397/tH}{
  \pgfmathsetmacro{\x}{14.10+0.26*\i}
  \draw[\col!65,fill=\col!10,line width=0.4pt,rounded corners=1.1pt]
    (\x-0.110,3.105) rectangle ++(0.220,0.220);
  \node[inner sep=0pt] at (\x,3.215)
    {\teaserimg{0.172cm}{\img}};
  \draw[oursblue!25,line width=0.3pt] (\x,3.095) -- (15.02,2.775);}
\draw[oursblue!18,fill=oursblue!6,rounded corners=3.2pt] (14.585,1.855) rectangle (15.455,2.725);
\draw[oursblue!85,fill=white,line width=0.7pt,rounded corners=2.6pt] (14.62,1.89) rectangle (15.42,2.69);
\foreach \dx/\dy/\img in {%
  -0.255/0.080/mnist, -0.085/0.080/eurosat, 0.085/0.080/gtsrb, 0.255/0.080/svhn,
  -0.255/-0.080/dtd, -0.085/-0.080/resisc45, 0.085/-0.080/cars, 0.255/-0.080/sun397}{
  \node[inner sep=0pt] at ({15.02+\dx},{2.29+\dy})
    {\teaserimg{0.150cm}{\img}};}
\node[font=\fontsize{4.6}{5.0}\selectfont,text=oursblue!85!black] at (15.02,1.665)
  {one model, all tasks};
\foreach \x/\yy/\t in {13.20/1.44/no images, 14.58/1.44/no labels,
                       13.20/1.10/no gradients, 14.58/1.10/order-free}{
  \node[draw=gray!40,rounded corners=2pt,fill=gray!5,inner sep=1.8pt,
        font=\fontsize{4.6}{5.0}\selectfont,anchor=west] at (\x,\yy) {\t};}

\draw[->,gray!60,line width=0.7pt] (11.62,2.42) -- (12.24,2.42);
\node[font=\fontsize{5.0}{5.6}\selectfont,text=black!62,below] at (11.93,2.38) {sum};

\end{tikzpicture}

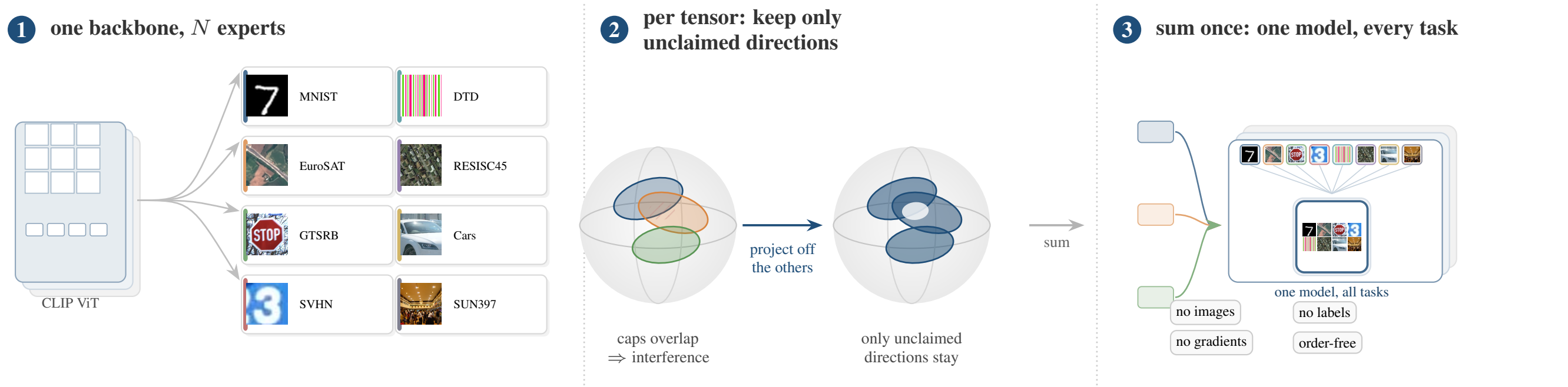
\captionof{figure}{\textbf{\ours{} in three moves.} (1)~One pretrained CLIP encoder is fine-tuned once per task, so the experts share a coordinate system and need no alignment. (2)~A task's update inside a weight tensor concentrates on a few input directions (a cap); where caps overlap, tasks interfere, and entrywise rules cannot see it. \ours{} projects each task off the union of the \emph{other} tasks' caps, keeping only directions no neighbour claims, symmetrically and in no fixed order. (3)~The projected updates sum to one model serving every task, with no images, labels, gradients or optimisation.}
\label{fig:teaser}
}
\makeatother
\maketitle
\begin{abstract}
Merging independently fine-tuned checkpoints into a single multi-task model requires no data, no labels, and no gradient steps. In practice the task vectors of different tasks interfere, and the standard remedies---averaging, magnitude pruning, sign election---operate entry by entry, ignoring that each delta induces a low-dimensional update on the hidden features it acts upon. We recast merging as a \emph{subspace interference} problem. For every two-dimensional weight tensor we take the top-$r$ right-singular subspace of each task's delta---the input directions along which that task moves the representation most---and remove from each task vector the component lying in the union of the \emph{other} tasks' subspaces. The resulting \emph{Joint Null-Space Projection} (\ours) is closed-form and order-independent, needs no forward pass, and is defined entirely by the checkpoint set. On eight CLIP image-classification tasks \ours{} merges ViT-B/16 experts to \textbf{89.45\%} average accuracy without using a single image, \textbf{11.88} points above the strongest parameter-wise training-free baseline (TIES-Merging, 77.57\%), 14.22 above Task Arithmetic, and 0.98 above WUDI's data-free consensus merge (88.47\%). The gain holds across three backbones on the independently trained TALL release ($+10.8$, $+12.6$, $+9.1$ over TIES for ViT-B/16, ViT-B/32, ViT-L/14) and replicates there at 89.20\%; controls confirm the source---a single shared subspace instead of per-task ones loses 2.9 points, and a sequential variant collapses to 69.6\%. On RoBERTa-large the same projection underperforms plain task arithmetic, which we report rather than hide. Code is available at \url{https://github.com/sumorday/JNSP}.
\end{abstract}


\section{Introduction}
\label{sec:intro}

Fine-tuning a foundation model once per task leaves checkpoints $\{\btheta_1,\dots,\btheta_N\}$ that descend from $\btheta_0$ but serve different purposes, and serving them jointly costs $N$ copies plus input routing. \emph{Model merging} synthesizes one parameter vector instead. The direct route adds the \emph{task vectors} $\btau_i = \btheta_i - \btheta_0$~\cite{ilharco2023task,wortsman2022model,matena2022merging}, which degrades as $N$ grows because the updates conflict; existing remedies treat this \emph{interference} entrywise---trimming and sign voting~\cite{yadav2023ties}, random dropping~\cite{yu2024dare}, outlier sparsification~\cite{davari2024model}---or reweight the sum~\cite{yang2024adamerging,daheim2024merging,gargiulo2025tasksingular,cheng2025wudi,shi2025fda} (Appendix~\ref{app:related}).

All of these assume interference is a property of \emph{individual parameters}. Yet a task vector has structure: for a weight matrix $\mW \in \R^{m \times n}$ reading an $n$-dimensional hidden state, the relevant geometry is the row space---which input directions the update amplifies~\cite{ortizjimenez2023task}---and entrywise statistics cannot see it.

We merge in the subspace where each task acts. Per tensor we take the top-$r$ right-singular subspace of $\btau_i$ and define task $i$'s \emph{joint null space} as the orthogonal complement of the union of the other tasks' subspaces; projecting $\btau_i$ onto it keeps what no neighbour claims and deletes what neighbours claim. Summing the projections gives \textbf{Joint Null-Space Projection} (\ours): SVDs and QR factorizations only, no data, no gradients, no optimization, and order-invariant because every task is projected against the same neighbours. The subspace serves as a \emph{conflict map between tasks} rather than a compression of each delta~\cite{gargiulo2025tasksingular} (Section~\ref{sec:theory}); Figure~\ref{fig:teaser} summarizes it.

On eight CLIP-based image-classification tasks, \ours{} merges ViT-B/16 experts to \textbf{89.45\%} average top-1, against 77.57\% for TIES-Merging and 75.23\% for Task Arithmetic, and the lead holds on ViT-B/32, ViT-L/14 and an independently trained checkpoint release (89.20\% versus 78.36\%), without using a single image. It does not transfer to RoBERTa-large on GLUE, although language task-vector subspaces overlap about as much as the vision ones (mean squared overlap $0.142$ versus $0.179$, both far above chance), so overlap alone does not explain the difference.

\paragraph{Contributions.}
\begin{enumerate}[leftmargin=*,itemsep=1pt,topsep=2pt]
  \item Merging cast as \emph{cross-task subspace interference}, a low-rank structure that entrywise rules cannot access (\S\ref{sec:method}).
  \item \ours: closed-form, data-free, order-independent, one rank parameter, a sum of per-task least-squares projections related to a joint least-squares problem (\S\ref{sec:theory}).
  \item $+11.88$ points over the best parameter-wise training-free baseline, $+0.98$ over WUDI's consensus merge, stable across three backbones and an independent release, with ablations isolating the gain (\S\ref{sec:exp},~\S\ref{sec:ablation}).
  \item A diagnosed negative result on language models (\S\ref{sec:negative}).
\end{enumerate}


\section{Related Work}
\label{sec:related}

\paragraph{Weight-space merging and interference.}
Model Soups averages checkpoints of one task~\cite{wortsman2022model}; Task Arithmetic sums \emph{task vectors} across tasks~\cite{ilharco2023task} and Fisher weighting reweights them~\cite{matena2022merging}. Because these are coordinatewise, most subsequent work searches for the right entrywise filter: TIES trims to large magnitudes and elects signs by majority~\cite{yadav2023ties}, DARE drops a random fraction and rescales~\cite{yu2024dare}, Model Breadcrumbs removes outlier and near-zero coordinates~\cite{davari2024model}. Each acts on coordinates independently and therefore commutes with any permutation of a row or column, which is exactly what discards the geometry we exploit.

\paragraph{Learned and subspace merging.}
A second family learns the merge rather than designing it---per-layer coefficients from unlabeled-test entropy~\cite{yang2024adamerging}, gradient or uncertainty matching~\cite{daheim2024merging}, evolutionary search~\cite{akiba2024evolutionary}, distillation~\cite{wan2023fusellm}---at the cost of an optimization loop and, typically, target data. A third works in singular space: TSV compresses each delta to its dominant singular vectors~\cite{gargiulo2025tasksingular}, KnOTS aligns LoRA subspaces~\cite{stoica2025knots}, Tam~\etal{} restrict the merge to a shared subspace~\cite{tam2024matching}, SCORE builds one orthogonal basis from the union of the deltas~\cite{chaves2026score}, Iso-C and Iso-CTS whiten a common subspace~\cite{marczak2025isocts}, and TALL merges through task-vector geometry~\cite{wang2024tall}. AlphaMerging~\cite{bazarvaani2026alphamerging} is closest: it projects each delta off the output-side (left) singular vectors of its peers, compressed to $k$ directions, whereas \ours{} uses the input-side (right) vectors and keeps the union uncompressed---a \emph{conflict map} rather than a compression. Projecting a task vector off its peers' singular subspaces is therefore not new in itself (AlphaMerging here, NUFILT below); what we contribute is a specific construction---right-side bases, an uncompressed leave-one-out peer union, applied per tensor with no merge coefficient---and a controlled comparison against these operators on one checkpoint and evaluator. ESM argues that right singular vectors need not align with function~\cite{li2026esm}; FDA instead reconstructs synthetic inputs from gradients~\cite{shi2025fda}, initialized by WUDI to reduce interference~\cite{cheng2025wudi}. Section~\ref{sec:method_role} takes up ESM's objection directly.

\begin{figure*}[t]
\centering
\includegraphics[width=\textwidth]{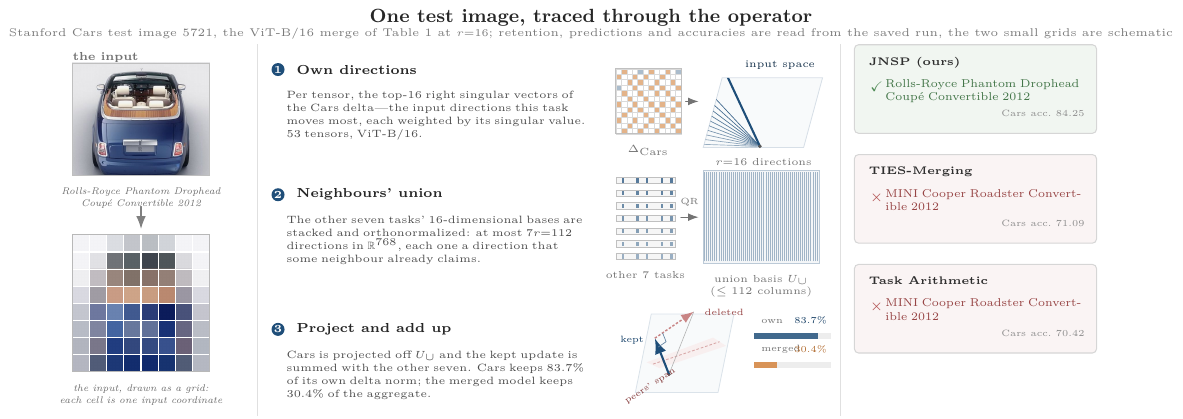}
\caption{\textbf{One merged prediction, traced through the operator.} Image 5721 of the Stanford Cars test split, predicted by the same merged checkpoints as the Cars column of Table~\ref{tab:main}. Left: the test image, and the same input drawn as a cell grid. Middle: the three steps of \ours{} for this task, with schematic graphics for the delta and the basis (no checkpoint is stored for rendering them). The two retention figures are averages over tensors and come from our $r{=}16$ run: Cars keeps 83.7\% of its own delta norm, and the merged model keeps 30.4\% of the aggregate norm. Right: predictions of \ours{}, TIES-Merging and Task Arithmetic on this image, with each method's Cars accuracy. \ours{} is correct; both entrywise baselines predict the same wrong class. The sample is one of 962 Cars test images on which \ours{} is correct and both baselines err (of 8{,}040). WUDI is not shown because its per-sample predictions were not saved.}
\label{fig:cars_example}
\end{figure*}

\paragraph{Null spaces, merge order, and linearization.}
ZipIt~\cite{stoica2024zipit}, Git Re-Basin~\cite{ainsworth2023git}, and optimal-transport fusion~\cite{singh2020model} align neurons across independently initialized models~\cite{frankle2020linear,li2018visual,entezari2022role}; we share one initialization and handle the geometry of the deltas instead. Projecting updates off previously learned directions is standard in continual learning, notably Gradient Projection Memory~\cite{saha2021gpm}, and underlies memory-efficient adaptation~\cite{zhaod2024galore,hu2022lora,houlsby2019adapters}. NUFILT~\cite{qiu2026nufilt} applies the \emph{same} operator, $\mI - \mV\mV^\top$, but accumulates the filtering basis along the merge order, so its result depends on that order; our leave-one-out construction applies it jointly, and Section~\ref{sec:abl_seq} shows the difference is not cosmetic (18.4 points). Appendix~\ref{app:related} develops each family at length, and Figure~\ref{fig:cars_example} traces one test image through the operator.


\section{Method}
\label{sec:method}

\subsection{Preliminaries}
\label{sec:prelim}

Let $\btheta_0 \in \R^d$ denote the pretrained weights and $\btheta_i$ the weights after fine-tuning on task $i \in [N] \equiv \{1,\dots,N\}$. The \emph{task vector} is $\btau_i = \btheta_i - \btheta_0$, and Task Arithmetic merges with $\btheta_0 + \lambda \sum_i \btau_i$ for a scaling coefficient $\lambda$~\cite{ilharco2023task}.

As in this literature, layer norms and biases stay at their pretrained values and each task keeps its own head; the merge applies to two-dimensional tensors whose smaller dimension is at least 64---the attention and MLP matrices plus five projection and embedding tensors (Appendix~\ref{app:setup}). For ViT-B/16 this is 53 tensors, and every baseline in \S\ref{sec:exp} uses the same subset.

Let $\bm{W}^{(k)}_0 \in \R^{m \times n}$ be the pretrained value of the $k$-th merged tensor, $\Delta^{(k)}_i = \bm{W}^{(k)}_i - \bm{W}^{(k)}_0$ the task-$i$ delta, and suppress $k$ below. The tensor reads an $n$-dimensional hidden state $\bm{x}$ and contributes $\Delta \bm{x}$ to the output, so the geometry that matters lives on the \emph{input} side---the row space of $\Delta$, which coincides with its right singular subspace.

\subsection{Cross-Task Subspace Interference}
\label{sec:interference}

For a single tensor, define the rank-$r$ right singular subspace
\begin{equation}
  \mathcal{S}_i \;=\; \Span\!\left\{\bm{v}_{i,1},\dots,\bm{v}_{i,r}\right\},
  \qquad
  \Delta_i \;=\; \sum_{j=1}^{\min(m,n)} \sigma_{i,j}\, \bm{u}_{i,j}\bm{v}_{i,j}^{\top},
  \label{eq:svd}
\end{equation}
where $\sigma_{i,1} \ge \sigma_{i,2} \ge \cdots$ are singular values. By Eckart--Young~\cite{eckart1936approximation,golub2013matrix}, $\mathcal{S}_i$ captures the directions of input space along which task $i$ most strongly perturbs the hidden representation in squared Frobenius norm---the task's own low-rank coordinate system.

Two tasks conflict when they claim the same directions. If $\bm{x}$ has a substantial component in $\mathcal{S}_i \cap \mathcal{S}_j$, then $(\Delta_i + \Delta_j)\bm{x}$ carries two independent contributions along the same input direction; they may add---benign, if both tasks want the same change---or cancel, in which case the smaller is erased. Because the result then passes through a nonlinearity and normalization, the cancellation is not a simple subtraction of accuracies but can re-route downstream computation. This is the mechanism behind TIES-Merging's ``signal conflict''~\cite{yadav2023ties}, stated at the level of subspaces rather than coordinates.

\paragraph{Why entrywise methods miss it.}
The trimming, sign election, and random dropping rules of TIES and DARE act on each entry independently, so they commute with any permutation of the entries of $\Delta$. But permuting the columns of $\Delta$ changes its right singular subspace entirely; two deltas with identical entrywise statistics can therefore have completely different interference structure. Interference is a \emph{geometric} rather than a marginal phenomenon.

The size of this effect is measurable. For tasks $i,j$ and a tensor with orthonormal bases $\mV_i,\mV_j\in\R^{n\times r}$, let $O_{ij}=\lVert \mV_i^{\top}\mV_j\rVert_F^2/r$ be the mean squared cosine, which is $r/n$ in expectation for independently drawn subspaces and $1$ when they coincide. On our checkpoints no pair of tasks is anywhere near orthogonal: the weakest pair has mean squared overlap $0.130$ and the mean over pairs is $0.179$ (Figure~\ref{fig:overlap}). Fine-tuning the same encoder on eight unrelated datasets does not produce independent updates---every task moves the same features, along directions the others also use. Appendix~\ref{app:geometry} shows the overlap is spread across almost every pair rather than concentrated in a few, and is not explained by entrywise similarity alone.

\subsection{Joint Null-Space Projection}
\label{sec:jnsp}

We want each task to give up whatever lies in the directions other tasks move most. Define the joint null space of task $i$ as
\begin{equation}
  \mathcal{N}_i \;=\; \Bigl(\Span\textstyle\bigcup_{j \neq i} \mathcal{S}_j\Bigr)^{\!\perp},
  \label{eq:nullspace}
\end{equation}
the orthogonal complement of all directions claimed by the \emph{other} $N-1$ tasks. Let $\mU_\cup^{(i)} \in \R^{n \times q_i}$, $q_i \le (N-1)r$, be an orthonormal basis of the peer union $\Span\bigcup_{j\ne i}\mathcal{S}_j$, obtained by stacking the peers' top-$r$ right singular vectors and applying QR; $\mI-\mU_\cup^{(i)}\mU_\cup^{(i)\top}$ is then the orthogonal projector onto $\mathcal{N}_i$. The projected task vector is
\begin{equation}
  \tilde{\btau}_i \;=\; \btau_i \bigl(\mI - \mP_i\bigr),
  \qquad
  \mP_i \;=\; \mU_{\cup}\mU_{\cup}^{\top},
  \label{eq:proj}
\end{equation}
where $\mU_\cup=\mU_\cup^{(i)}$ (we drop the superscript when $i$ is clear). The merged model is
\begin{equation}
  \btheta_{\mathrm{merge}} \;=\; \btheta_0 + \sum_{i=1}^{N} \tilde{\btau}_i .
  \label{eq:merge}
\end{equation}
Equation~\eqref{eq:proj} is the only substantive step: it deletes from $\Delta_i$ the part lying in the dominant-update directions of any other task. This includes the part of task $i$'s own $\mathcal{S}_i$ that lies inside the peer union; $\mathcal{S}_i$ is not exempt from the projection, it is only not added to $\mU_\cup^{(i)}$.

\begin{figure}[t]
\centering
\includegraphics[width=\columnwidth]{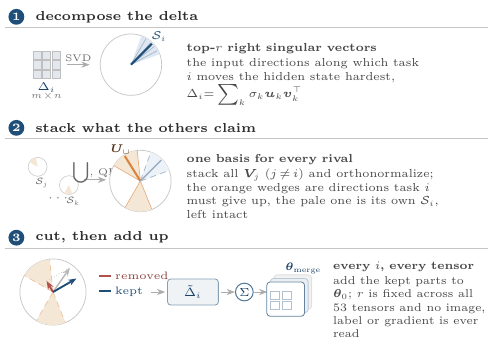}
\caption{\textbf{How \ours{} merges one tensor.} (1)~The top-$r$ right singular vectors $\mathcal{S}_i$ of $\Delta_i$ are the input directions task $i$ moves most. (2)~The other tasks' bases are stacked and orthonormalized to $\mU_\cup$; the orange wedges are directions task $i$ must give up; $\mathcal{S}_i$ itself is not added to $\mU_\cup$, but any part of it inside the peer union is removed. (3)~$\Delta_i$ is projected onto $\mathcal{N}_i$, keeping $\tilde\Delta_i$ and removing the rest; the kept parts are summed into $\btheta_{\mathrm{merge}}$. The construction never depends on task order.}
\label{fig:jnsp}
\end{figure}

Several properties follow directly. \emph{Data-freedom}: no input, label, or forward pass enters~\eqref{eq:proj}. \emph{Closed form}: one SVD per task per tensor plus $N$ QR factorizations, with no optimization. \emph{Symmetry}: since $\mathcal{N}_i$ depends on the set $\{j : j \ne i\}$ rather than on an ordering, permuting the task list leaves~\eqref{eq:merge} unchanged---and Section~\ref{sec:theory} shows the natural sequential alternative fails precisely because it lacks this. Figure~\ref{fig:jnsp} draws the construction; Appendix~\ref{app:worked} traces it numerically on a single tensor.

The projection removes only the component of $\Delta_i$ lying inside another task's subspace: if $\mathcal{S}_i$ and $\mathcal{S}_j$ are disjoint, $\Delta_i$ is untouched by task $j$; if they coincide, $\Delta_i$ loses everything it shares. At $r{=}16$ with $n{=}768$ each $\mathcal{S}_i$ is a small fraction of $\R^n$, yet every pair of task subspaces overlaps well above the random value $r/n$, with mean squared overlap between $0.130$ and $0.338$ (Figure~\ref{fig:overlap}), so the seven-neighbour union---at most $7r{=}112$ directions of a $768$-wide tensor---lands where the deltas concentrate rather than spreading thinly over the input space (Appendix~\ref{app:geometry}).

\begin{algorithm}[t]
\caption{Joint Null-Space Projection (\ours)}
\label{alg:jnsp}
\begin{algorithmic}[1]
\Require $\btheta_0$, $\{\btheta_i\}_{i=1}^N$, rank $r$
\Ensure merged weights $\btheta_{\mathrm{merge}}$
\State $\btau_i \gets \btheta_i - \btheta_0$; $\btheta_{\mathrm{merge}} \gets \btheta_0$
\For{each 2D tensor $\bm{W}$ with $\min(\mathrm{shape}) \ge 64$}
  \For{$i \gets 1$ \textbf{to} $N$}
    \State $[\cdot,\cdot,\bm{V}_i^{\top}] \gets \mathrm{SVD}(\Delta_i)$;\; $\mV_i \gets \bm{V}_i[:, 1{:}r]$
  \EndFor
  \For{$i \gets 1$ \textbf{to} $N$}
    \State $\bm{U}_\cup \gets \mathrm{QR}\bigl([\,\mV_j \;\text{for}\; j \neq i\,]\bigr)$
    \State $\btheta_{\mathrm{merge}}[\bm{W}] \mathrel{+}= \Delta_i\,(\mI - \bm{U}_\cup\bm{U}_\cup^{\top})$
  \EndFor
\EndFor
\State \Return $\btheta_{\mathrm{merge}}$
\end{algorithmic}
\end{algorithm}

\paragraph{Complexity.}
For an $m \times n$ tensor with $m \ge n$, a full SVD costs $\mathcal{O}(mn^2)$; with $N$ tasks and $K$ merged tensors the merge costs $\mathcal{O}(K N mn^2)$, with negligible QR overhead. On ViT-B/16 the merge takes about 61 seconds on CPU for 53 tensors and $N{=}8$ (Appendix~\ref{app:runtime}).

\subsection{A Least-Squares View}
\label{sec:theory}

We now justify the construction. Suppose the merged delta for tensor $k$ is constrained to lie in a subspace $\mathcal{W} \subseteq \R^{m \times n}$ that excludes the interfering directions, and that we want it to be as close as possible to the sum of the task deltas:
\begin{equation}
  \min_{\mZ \in \mathcal{W}} \; \Bigl\lVert \mZ - \sum_{i=1}^N \Delta_i \Bigr\rVert_F^2 .
  \label{eq:ls}
\end{equation}
If $\mathcal{W}$ is the set of matrices whose rows are orthogonal to a subspace with orthonormal basis $\mB$, the minimizer is $\mZ^\star = \bigl(\sum_i \Delta_i\bigr)(\mI - \mB\mB^\top)$. Three properties relate this object to \ours{}; Appendix~\ref{app:proofs} gives the statements in full and proves each.

\emph{(i) Per-task least squares, not the joint solution.} Each projected delta is itself a least-squares solution: $\tilde\btau_i$ is the matrix closest to $\Delta_i$ in Frobenius norm whose rows are orthogonal to the peer union $\mathcal{T}_i=\Span\bigcup_{j\ne i}\mathcal{S}_j$. The joint problem~\eqref{eq:ls} with $\mB$ spanning the full union $\mathcal{T}=\Span\bigcup_j\mathcal{S}_j$ instead projects the \emph{sum} once; \ours{} differs from that solution exactly by $\sum_i\Delta_i\mP_{\mathcal{R}_i}$, where $\mathcal{R}_i=\mathcal{T}\ominus\mathcal{T}_i$ (dimension at most $r$) is the part of $\mathcal{T}$ that only task $i$ contributes. \ours{} therefore keeps, for each task, the directions of the union that no peer claims, and deletes everything the peers claim, including the part of $\mathcal{S}_i$ that falls inside $\mathcal{T}_i$. The shared-basis control (Section~\ref{sec:abl_global}) projects off a different basis and loses 2.94 points; $\mZ^\star$ is not evaluated. This is a design rationale, not an optimality claim.

\emph{(ii) Sequential projection is directed.} Projecting $\hat\btau_i = \btau_i(\mI - \mQ_{<i}\mQ_{<i}^\top)$ against a nested basis $\mQ_{<i}$ containing only earlier tasks is a contraction with a bound $\norm{\sum_i\hat\btau_i}_F \le \sum_i\norm{\btau_i}_F$, deletes a growing dimension from later tasks, and is not permutation-invariant---consistent with the 18.4-point collapse of Section~\ref{sec:abl_seq}. The stronger claim that the increments are exactly orthogonal is false, and we checked it rather than assumed it (Appendix~\ref{app:proof_sequential}).

\emph{(iii) Invariance of the joint merge.} Because $\mathcal{N}_i$ is defined from the \emph{set} of neighbours and $\mathcal{S}_i$ depends on $\Delta_i$ only through the span of its right singular vectors, merge~\eqref{eq:merge} is invariant to permutations of the task list and to the sign ambiguity of the computed singular vectors. It is \emph{not} invariant to negating a task vector: $\btau_i\mapsto-\btau_i$ leaves every projector unchanged and therefore maps $\tilde\btau_i\mapsto-\tilde\btau_i$.

Together these state what the method is: a symmetric, closed-form set of per-task least-squares projections that deletes whatever peers claim and keeps what only the task itself contributes. They are consistent with the two failure modes we observe---removing the consensus component and sequentializing---and identify $r$ as the knob trading retained signal against removed interference.

\subsection{What the Singular Basis Is Used For}
\label{sec:method_role}

ESM~\cite{li2026esm} argues against right singular vectors as a merging basis: the singular value ordering of $\Delta_i$ need not track functional importance, so a delta compressed to its own top-$r$ directions can discard a direction that matters to the task but carries little energy. We accept the premise, but it does not apply unchanged to~\eqref{eq:proj}, because our basis never compresses the delta it is computed from. $\mathcal{S}_i$ is never used to project $\Delta_i$---it is excluded from $\mU^{(i)}_\cup$ and only enters the other tasks' projectors---so the only thing deleted is $\Delta_i$'s component inside \emph{other} tasks' top-$r$ spaces (which can include part of $\mathcal{S}_i$); a direction that is low-energy for task $i$ and not dominant for any neighbour is retained in full. The objection does survive in one place, and we state it rather than argue it away: a direction can be low-$\sigma_i$, dominant for exactly one neighbour, and at the same time carry real function for task $i$. Equations~\eqref{eq:svd}--\eqref{eq:proj} cannot tell that case apart from a genuinely shared direction. \emph{Verified}: the operator does not truncate $\Delta_i$ to $\mathcal{S}_i$. \emph{Unverified}: whether the deleted directions carry function for the task that loses them, and whether an activation-space basis of the kind ESM fits would select a better deletion set (Appendix~\ref{app:role} takes up both).

The same limitation applies to the meaning of overlap. \ours{} treats a direction two tasks both move along as a \emph{conflict proxy}, on the assumption that two tasks cannot both spend the same capacity. Overlap alone does not establish that the direction is contested rather than shared-beneficially. What the method measures is the geometric redundancy of the updates, not the sign of their interaction; Section~\ref{sec:negative} gives a case where this reading fails outright.


\section{Experiments}
\label{sec:exp}

\subsection{Setup}
\label{sec:setup}

\paragraph{Tasks and checkpoints.}
We follow the eight-task image-classification protocol now standard in model merging: MNIST~\cite{lecun1998mnist}, EuroSAT~\cite{helber2019eurosat}, GTSRB~\cite{stallkamp2012gtsrb}, SVHN~\cite{netzer2011svhn}, DTD~\cite{cimpoi2014dtd}, RESISC45~\cite{cheng2017resisc45}, Stanford Cars~\cite{krause2013cars}, and SUN397~\cite{xiao2016sun}. Each task starts from the same pretrained CLIP image encoder~\cite{radford2021clip,cherti2023openclip} and is fine-tuned end to end with its own head, giving $N{=}8$ checkpoints per backbone (ViT-B/16, ViT-B/32, ViT-L/14~\cite{dosovitskiy2021vit}).

\paragraph{Second checkpoint release.}
Because our ViT-B/16 checkpoints are our own, we repeat every experiment on the task vectors released by TALL~\cite{wang2024tall}, produced by an independent pipeline with different fine-tuning recipes and heads.

\paragraph{Evaluation.}
We report top-1 accuracy with each task's own head on the official test split. Merging applies to the 53 two-dimensional tensors with $\min(\text{shape}) \ge 64$ (Section~\ref{sec:prelim}); every method in Table~\ref{tab:main} operates on this subset except WUDI, whose released operator merges 48 of the 53 and leaves five embedding and projection tensors at their pretrained values (Appendix~\ref{app:datausing}), so the comparison largely isolates the merge rule rather than the parameter scope.

\paragraph{Baselines.}
\emph{Entrywise and consensus, data-free:} Task Arithmetic~\cite{ilharco2023task}, TIES-Merging~\cite{yadav2023ties}, and DARE~\cite{yu2024dare}, plus WUDI's consensus subtraction~\cite{cheng2025wudi} scored as a merge rule without adaptation. For each sweepable method we sweep the merge coefficient $\lambda$ and report the best score, the setting most favourable to the baseline. \emph{Subspace-based (data-free):} TSV-M~\cite{gargiulo2025tasksingular}, Iso-C, Iso-CTS~\cite{marczak2025isocts}, and AlphaMerging~\cite{bazarvaani2026alphamerging}, re-implemented in our harness (Section~\ref{sec:repro_gap}). All baselines in Table~\ref{tab:main} are data-free. \ours{} has no hyperparameter other than $r$, fixed at 16 everywhere: it was chosen once on the ViT-B/16 rank sweep of these same eight tasks (Section~\ref{sec:abl_rank}) and never retuned.

\begin{figure*}[t]
\centering
\includegraphics[width=\textwidth]{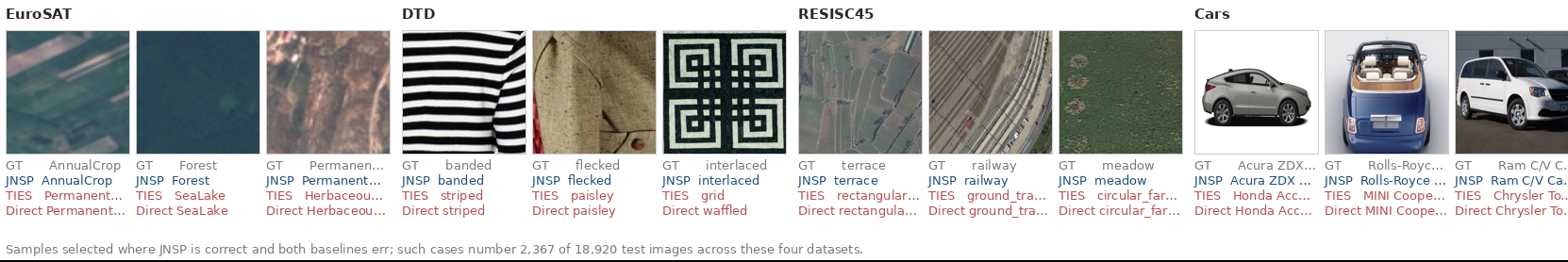}
\caption{\textbf{Qualitative comparison on held-out test images.} Each column is one image; rows give the ground-truth label and the predictions of \ours{}, TIES-Merging, and Task Arithmetic (blue: correct, red: incorrect). Samples are drawn at random from cases where \ours{} is correct and both baselines err. The merge is entirely data-free in all three methods; only the merge rule differs. Per-image agreement counts are in Appendix~\ref{app:qual}.}
\label{fig:qual}
\end{figure*}

\begin{table*}[t]
\centering
\caption{\textbf{Main comparison on ViT-B/16}, eight image-classification tasks, our checkpoints. All methods are data-free and merge the same 53 tensors, except WUDI (48 of them; Appendix~\ref{app:datausing}). \emph{Avg} is what our harness produced for every row, on one checkpoint and one evaluator. \emph{Pub.} is the average quoted from the original paper (TSV-M~\cite{gargiulo2025tasksingular}, AlphaMerging~\cite{bazarvaani2026alphamerging}), from a different checkpoint, tensor scope and pipeline; we do not reproduce it (Section~\ref{sec:repro_gap}). The subspace block is our re-implementation on the identical checkpoint and scope.}
\label{tab:main}
\setlength{\tabcolsep}{3.6pt}
\begin{tabular}{@{}l*{8}{c}cc@{}}
\toprule
Method & MNIST & EuroSAT & GTSRB & SVHN & DTD & RESISC45 & Cars & SUN397 & \textbf{Avg} & Pub. \\
\midrule
Task Arithmetic~\cite{ilharco2023task} & 97.15 & 81.81 & 71.08 & 83.02 & 52.55 & 77.51 & 70.42 & 68.29 & 75.23 & --- \\
DARE~\cite{yu2024dare}                & 97.06 & 81.30 & 70.91 & 82.40 & 52.07 & 77.10 & 70.10 & 68.31 & 74.91 & --- \\
TIES-Merging~\cite{yadav2023ties}     & 98.71 & 82.78 & 71.86 & 90.35 & 57.98 & 81.10 & 71.09 & 66.70 & 77.57 & --- \\
\midrule
WUDI~\cite{cheng2025wudi}             & \textbf{99.44} & 98.00 & \textbf{96.55} & \textbf{95.30} & 73.62 & 90.13 & 81.58 & 73.13 & 88.47 & --- \\
\midrule
TSV-M$^\dagger$~\cite{gargiulo2025tasksingular} & 99.27 & 96.67 & 93.96 & 93.93 & 72.61 & 89.16 & 80.67 & 71.49 & 87.22 & 89.01 \\
Iso-C$^\dagger$                        & 98.78 & 96.78 & 91.74 & 88.27 & 71.65 & 92.16 & 82.54 & 75.28 & 87.15 & n/a \\
Iso-CTS$^\dagger$~\cite{marczak2025isocts} & 98.25 & 96.37 & 91.50 & 84.89 & 70.90 & 92.13 & 83.27 & \textbf{75.44} & 86.59 & --- \\
AlphaMerging$^\dagger$~\cite{bazarvaani2026alphamerging} & 99.00 & 94.56 & 92.34 & 88.24 & 77.07 & 91.30 & 81.42 & 73.84 & 87.22 & 89.9 \\
\midrule
\textbf{\ours{} ($r{=}16$)}           & 99.07 & \textbf{98.44} & 96.26 & 92.16 & \textbf{77.23} & \textbf{93.22} & \textbf{84.25} & 74.97 & \textbf{89.45} & --- \\
\bottomrule
\multicolumn{11}{@{}p{\textwidth}@{}}{\footnotesize $^\dagger$Our re-implementation of the published operator on this checkpoint. AlphaMerging is the projection of its Algorithm~1 (per-task rank $r_1$, shared rank $k$) followed by the closed-form WUDI solve. Its row is the best of a sweep: $r_1\in\{8,16,32,64\}$ at $k{=}64$ gives 86.60, 87.11, 87.22, 86.78 (the $r_1{=}32$ setting is shown), and $k\in\{16,32,64,112\}$ at $r_1{=}16$ gives 86.94, 86.72, 87.11, 86.32. The best setting of each sweep is reported, which favours the baseline.}
\end{tabular}
\end{table*}

\subsection{Main Results}
\label{sec:main_results}

Table~\ref{tab:main} is the central result. \ours{} merges the eight ViT-B/16 experts to 89.45\% average accuracy, 11.88 points above TIES-Merging (77.57\%), the strongest of the parameter-wise baselines, and 14.22 points above Task Arithmetic (75.23\%). It also leads the subspace family: TSV-M, AlphaMerging, Iso-C and Iso-CTS, re-implemented on the identical checkpoint and tensor scope, reach 87.22, 87.22, 87.15 and 86.59, so \ours{} is 2.23--2.86 points ahead. Against WUDI's consensus subtraction (88.47\%) it leads by 0.98 points without tuning a threshold, a sign vote, or a drop rate.

The per-dataset columns show where the margin comes from. TIES is competitive on MNIST (98.71) but degrades on visually diverse tasks (57.98 on DTD, 71.09 on Cars, 66.70 on SUN397); \ours{} loses far less on exactly these (77.23, 84.25, 74.97), the signature one expects if the failure is interference between conflicting low-rank updates rather than accumulated magnitude. Against the subspace family the pattern differs: Iso-C and Iso-CTS are \emph{stronger} on SUN397 (75.3--75.4) but fall below \ours{} on SVHN (84.9--88.3 versus 92.16) and DTD. Appendix~\ref{app:gain} decomposes both margins task by task; against the subspace family \ours{} wins on five tasks and trails by at most 1.77. Figure~\ref{fig:qual} shows the margin on individual test images.

\begin{table*}[t]
\centering
\caption{\textbf{Scaling across backbones} on the independently trained TALL release; every number is on that release, so cross-row comparisons are apples-to-apples. \ours{} uses $r{=}16$ throughout, untuned per backbone. $^\dagger$Re-implementation (Section~\ref{sec:repro_gap}). All methods are data-free and training-free.}
\label{tab:backbones}
\setlength{\tabcolsep}{4.9pt}
\footnotesize
\begin{tabular}{@{}ll*{8}{c}c@{}}
\toprule
Backbone & Method & MNIST & EuroSAT & GTSRB & SVHN & DTD & RESISC45 & Cars & SUN397 & \textbf{Avg} \\
\midrule
\multirow{8}{*}{ViT-B/16}
 & Task Arithmetic & 97.10 & 84.89 & 72.04 & 80.45 & 52.87 & 76.95 & 71.44 & 68.30 & 75.51 \\
 & DARE            & 97.08 & 84.70 & 72.43 & 80.20 & 52.77 & 76.43 & 71.16 & 68.09 & 75.36 \\
 & TIES-Merging    & 98.69 & 89.41 & 73.34 & 89.19 & 56.17 & 79.71 & 73.64 & 66.69 & 78.36 \\
 & TSV-M$^\dagger$  & \textbf{99.33} & 97.52 & 93.99 & \textbf{93.12} & 71.44 & 88.95 & 82.19 & 71.44 & 87.25 \\
 & Iso-C$^\dagger$  & 98.86 & 97.56 & 92.78 & 87.05 & 72.07 & 92.21 & 83.74 & 75.42 & 87.46 \\
 & Iso-CTS$^\dagger$ & 98.55 & 96.78 & 92.15 & 83.35 & 70.96 & 92.06 & 83.61 & \textbf{75.50} & 86.62 \\
 & AlphaMerging$^\dagger$ & 98.91 & 92.93 & 90.09 & 84.54 & 74.31 & 89.92 & 79.29 & 73.18 & 85.40 \\
 & \textbf{\ours{} $r{=}16$} & 99.23 & \textbf{98.63} & \textbf{96.03} & 91.43 & \textbf{75.90} & \textbf{92.70} & \textbf{84.85} & 74.85 & \textbf{89.20} \\
\midrule
\multirow{8}{*}{ViT-B/32}
 & Task Arithmetic & 90.00 & 66.74 & 63.44 & 69.20 & 49.26 & 69.05 & 60.41 & 62.39 & 66.31 \\
 & DARE            & 90.06 & 65.56 & 61.63 & 68.37 & 48.72 & 68.92 & 59.94 & 62.08 & 65.66 \\
 & TIES-Merging    & 97.17 & 69.74 & 71.96 & 81.93 & 51.65 & 71.06 & 60.50 & 58.46 & 70.31 \\
 & TSV-M$^\dagger$ & \textbf{98.95} & 89.11 & \textbf{91.01} & \textbf{89.37} & 65.32 & 81.35 & 69.44 & 64.21 & 81.09 \\
 & Iso-C$^\dagger$ & 96.83 & 84.26 & 85.21 & 79.54 & 63.88 & 84.24 & 72.35 & 69.53 & 79.48 \\
 & Iso-CTS$^\dagger$ & 95.66 & 81.96 & 81.38 & 72.81 & 63.51 & 83.68 & \textbf{72.90} & \textbf{70.17} & 77.76 \\
 & AlphaMerging$^\dagger$ & 97.80 & 82.44 & 87.47 & 75.31 & 66.54 & 83.52 & 69.45 & 69.67 & 79.03 \\
 & \textbf{\ours{} $r{=}16$} & 98.91 & \textbf{92.52} & 90.36 & 87.04 & \textbf{68.03} & \textbf{85.89} & 72.21 & 68.40 & \textbf{82.92} \\
\midrule
\multirow{8}{*}{ViT-L/14}
 & Task Arithmetic & 97.74 & 82.48 & 79.41 & 81.91 & 60.64 & 79.89 & 79.91 & 70.38 & 79.05 \\
 & DARE            & 97.69 & 82.15 & 79.26 & 82.09 & 60.32 & 79.71 & 79.64 & 70.13 & 78.87 \\
 & TIES-Merging    & 98.88 & 87.37 & 82.67 & 88.98 & 61.22 & 83.32 & 81.21 & 70.16 & 81.73 \\
 & TSV-M$^\dagger$ & 99.40 & 97.33 & 95.46 & 94.45 & 73.72 & 90.00 & 87.15 & 73.28 & 88.85 \\
 & Iso-C$^\dagger$ & 99.08 & 95.70 & 93.98 & 90.73 & 74.95 & 91.25 & 88.59 & 76.79 & 88.89 \\
 & Iso-CTS$^\dagger$ & 98.86 & 94.78 & 93.64 & 88.71 & 74.31 & 91.38 & 88.93 & \textbf{76.87} & 88.44 \\
 & AlphaMerging$^\dagger$ & 99.34 & 97.81 & 95.91 & 89.39 & 76.33 & 92.57 & 87.55 & 76.44 & 89.42 \\
 & \textbf{\ours{} $r{=}16$} & \textbf{99.52} & \textbf{97.96} & \textbf{97.49} & \textbf{95.23} & \textbf{78.62} & \textbf{92.67} & \textbf{89.39} & 76.09 & \textbf{90.87} \\
\bottomrule
\end{tabular}
\end{table*}

\subsection{Reproduction Gap for Published Subspace Baselines}
\label{sec:repro_gap}

TSV-M reports 89.01~\cite{gargiulo2025tasksingular} and AlphaMerging 89.9~\cite{bazarvaani2026alphamerging} on eight-task ViT-B/16; our re-implementations reach 87.22 for both. Every comparison here is on \emph{Avg}, where all rows share one checkpoint, tensor scope and evaluator (Appendix~\ref{app:repro}).

All rows use one evaluation code, tensor scope, heads and test splits; the four subspace baselines were implemented from their published operators---top-$r$ task singular vectors with a per-task coefficient, a common subspace, its centred variant, and AlphaMerging's peer-output projection followed by the WUDI solve---and swept over their free parameters. Re-running TSV-M, Iso-C and Iso-CTS on the TALL checkpoint moves them by at most 0.31 (87.25, 87.46, 86.62 against 87.22, 87.15, 86.59), so the checkpoint is not the missing variable for these three. AlphaMerging was run on TALL at its best own-checkpoint setting ($r_1{=}32$, $k{=}64$) without re-sweeping and drops to 85.40 (87.22 at the same setting on our checkpoint), so for it the setting does not transfer unchanged. We therefore claim a 2.23--2.86 point lead under a protocol we control end to end.

Table~\ref{tab:backbones} reports the TALL release. On ViT-B/16, \ours{} reaches 89.20\% against 78.36\% for TIES, a 10.84-point lead (11.88 on our own checkpoints): the effect is a property of independently fine-tuned CLIP experts, not of one set of hyperparameters.

\subsection{Backbone Scaling}
\label{sec:scaling}

The gain does not depend on model size. On ViT-B/32 \ours{} reaches 82.92\% versus 70.31\% for TIES (+12.61) and 81.09\% for TSV-M, the best subspace baseline there (+1.83); on ViT-L/14, 90.87\% versus 81.73\% for TIES (+9.14) and 89.42\% for AlphaMerging (+1.45). The best subspace baseline changes across backbones (Iso-C on B/16, TSV-M on B/32, AlphaMerging on L/14), but \ours{} places first at every backbone. The margin over the subspace family (1.45--1.83) is much narrower than over TIES, and narrowest on the largest model.

\subsection{Statistical Precision}
\label{sec:stats}

\ours{} is deterministic: for fixed checkpoints and $r$ the merge is closed-form, so seed variance is identically zero. We report instead a stratified bootstrap over the evaluation set~\cite{efron1993bootstrap}: in each of 2{,}000 replicates, every task's test set is resampled with replacement independently at its original size, the eight task accuracies are recomputed and averaged with equal weight, and the 95\% interval is the 2.5th--97.5th percentile of the resulting macro-averages; point estimates are the observed (non-resampled) accuracies. This gives $89.45\%$ $[89.17, 89.75]$ on our checkpoints and $89.20\%$ $[88.90, 89.51]$ on TALL. The interval measures uncertainty in the \emph{evaluation}, not in training; it is narrow because the test sets are large (87{,}432 images in total).


\section{Ablations}
\label{sec:ablation}

The main table answers whether \ours{} works; this section asks why, using the ViT-B/16 checkpoints and the same protocol.

\subsection{Rank}
\label{sec:abl_rank}

The rank $r$ is the only hyperparameter. Figure~\ref{fig:rank} (Appendix~\ref{app:rank}) sweeps it from 8 to 32 together with the shared-basis control: 88.03\% at $r{=}8$, a peak of 89.45\% at $r{=}16$, 89.22\% at $r{=}20$, and 87.08\% at $r{=}32$. The $[12,20]$ plateau is flat---its endpoints differ by 0.08 points---and $r{=}16$ is chosen once, never retuned per backbone or dataset.

The two tails behave differently. At $r{=}32$ the neighbours' union covers much of each delta (26.0\% of the aggregate norm survives) and accuracy falls by 2.37 points. At $r{=}8$ the aggregate retained ratio is \emph{higher} (33.4\%), yet accuracy is 1.42 points lower than at $r{=}16$; retained norm alone therefore does not explain the low-rank tail, and a plausible reading---not tested here---is that an 8-dimensional peer basis misses interfering directions that a 16-dimensional one removes. The sweep extends to ViT-B/32 and ViT-L/14 on the TALL checkpoints (Table~\ref{tab:rank_sweep}, Appendix~\ref{app:rank}): the plateau shape is stable and accuracy peaks at or within 0.01 points of $r{=}16$ everywhere.

\subsection{Mechanism Control: Shared versus Per-Task Subspace}
\label{sec:abl_global}

The claim of Section~\ref{sec:method} is that the gain comes from using \emph{each task's} subspace to predict where \emph{another} task interferes. A simpler alternative computes one shared subspace---the top-$r$ right singular vectors of the \emph{sum} of all deltas---and projects every task against it, which equals projecting the summed delta once. That removes the direction all tasks share~\cite{ilharco2023task,ortizjimenez2023task} but has no notion of which task conflicts with which. It loses 2.94 points at $r{=}16$ (86.51 versus 89.45) and 2.14--3.15 at every rank. At 86.51\% it still beats the entrywise baselines (TIES 77.57\%) but trails WUDI and the subspace re-implementations (86.59--88.47\%), so shared-direction removal is not useless: it recovers about three quarters of \ours{}'s margin over TIES (8.94 of 11.88 points). The remainder is consistent with pairwise structure mattering---task $j$'s dominant directions being specifically the ones task $i$ must avoid---though this control does not isolate that explanation.

\subsection{Order Dependence}
\label{sec:abl_seq}

Continual-learning practice suggests merging tasks one at a time, projecting each against the basis accumulated so far. The sequential analysis of Section~\ref{sec:theory} predicts that this is order-dependent and progressively destructive, and the measurement is unambiguous: at matched rank $r{=}8$ the sequential variant scores 69.64\% against 88.03\% for the joint variant, a collapse of 18.4 points. The per-dataset pattern (Figure~\ref{fig:order}, Appendix~\ref{app:order}) is consistent with that prediction: early tasks survive nearly intact (MNIST 99.59\%, EuroSAT 93.81\%), while later ones are gutted---RESISC45 64.78\%, Cars 51.14\%, SUN397 36.14\%. The sequential variant also retains \emph{more} aggregate norm (35.90\% versus 33.44\%), so the amount deleted does not explain the collapse. Each task loses only its overlap with \emph{earlier} tasks, and early tasks are never filtered against later ones, so their overlap with later tasks enters the sum unprojected. The joint form avoids this asymmetry by construction: every task is projected against the same seven neighbours, so the merge is order-invariant (Section~\ref{sec:theory}).

\subsection{What the Projection Removes}
\label{sec:abl_norm}

At $r{=}16$ the merge keeps on average 77.8\% of each task's own delta norm (from 71.6\% on SVHN to 83.7\% on Cars), but only 30.4\% of the aggregate norm $\norm{\sum_i\tilde\btau_i}/\sum_i\norm{\btau_i}$ (Appendix~\ref{app:retention}). The aggregate is low mainly because the eight deltas are nearly orthogonal to begin with (mean entrywise cosine $0.083$), so their sum is already well below the sum of their norms; the projection lowers the ratio further. Per-tensor rank allocation gains only 0.21 points (Appendix~\ref{app:rank}).


\section{Where the Method Fails: Language Models}
\label{sec:negative}

We test a second architecture family. The result is negative, and it bounds what \ours{} can claim.

\paragraph{Setup and result.}
We merge eight RoBERTa-large~\cite{liu2019roberta} checkpoints fine-tuned on GLUE~\cite{wang2019glue} from one pretrained checkpoint, with the same $r$ and projection as in vision and the same $\min(\text{shape})\ge 64$ filter, which here selects 146 tensors (the 144 attention and FFN matrices plus the word and position embeddings). Task arithmetic at its usual scaling $\lambda{=}0.2$ reaches 59.70 and the shared-basis control 63.28; \ours{}, which has no merge coefficient ($\lambda{=}1$), reaches 31.43. It degrades most tasks, STS-B collapses to a negative score, and every diagnostic that explains the vision result inverts. Part of this gap is a scaling effect: unlike in vision, the unscaled sum is far from optimal on GLUE (task arithmetic at $\lambda{=}1$ scores 39.76, Appendix~\ref{app:nlu}), so the comparison above mixes two scalings.

\paragraph{It is not a matter of overlap.}
Language task vectors do not overlap more: at $r{=}16$ the mean squared overlap is $0.142$ for RoBERTa-large ($9.0\times$ its random baseline of $0.0157$) against $0.179$ for ViT-B/16, marginally \emph{lower}.

\paragraph{What the failure does say.}
In vision the joint merge retains 30.4\% of the delta norm and improves accuracy; in language it retains 42.7\%---\emph{more} mass---and yet destroys the merge. Projecting deletes less and hurts more. One untested reading is that in the language encoder the removed directions carry structure every task needs; our experiments cannot separate this from differences in architecture, objective, tokenization or scaling. A consensus--residual decomposition that projects the task-specific residual instead recovers 69.45\%, 6.17 points above the global control (Appendix~\ref{app:nlu}); we report it as a repair, not a contribution.

\section{Conclusion}
\label{sec:conclusion}

We cast model merging as subspace interference and proposed Joint Null-Space Projection, which removes from each task vector its components in the other tasks' dominant-update directions. It is closed-form and data-free, has one hyperparameter, and is flat over a wide rank range. On eight image-classification tasks it merges ViT-B/16 experts to 89.45\%, 11.88 points above the strongest parameter-wise training-free baseline and 0.98 above WUDI's consensus merge, a lead that survives three backbones and an independent checkpoint release. Generative objectives are untested, and the language failure is diagnosed but not explained.


\clearpage
\appendix
\section{Extended Related Work}
\label{app:related}

This appendix expands Section~\ref{sec:related} (see~\cite{yang2024survey} for a broader survey) and states, family by family, which precise operation each baseline performs on a task vector and where \ours{} departs from it.

\paragraph{Task arithmetic and its descendants.}
Model Soups~\cite{wortsman2022model} averages checkpoints fine-tuned from one initialization on the \emph{same} task and shows that the average is more robust than any member. Task Arithmetic~\cite{ilharco2023task} turns the averaging into a merge primitive for \emph{different} tasks: with $\btau_i = \btheta_i - \btheta_0$ it proposes $\btheta_0 + \lambda \sum_i \btau_i$ and demonstrates that a single scalar $\lambda$ controls the tradeoff between task performance and general capability; the earlier PAINT~\cite{ilharco2022patching} interpolates a fine-tuned and a zero-shot CLIP model for the same purpose. Fisher-weighted merging~\cite{matena2022merging} replaces the uniform average by a diagonal-Fisher weighting, which requires an estimate of the Fisher information per task. All three treat the sum coordinate by coordinate and have no mechanism for deciding that two coordinates belong to different tasks; the failure is visible in Table~\ref{tab:main} as the collapse on DTD and SUN397.

\paragraph{Entrywise interference resolution.}
TIES-Merging~\cite{yadav2023ties} applies three successive operations per tensor: trim each delta to its largest-$k$ magnitudes, elect a sign per coordinate by summing the signed normalized values, and average only the coordinates whose sign agrees with the election. DARE~\cite{yu2024dare} zeroes a fraction $p$ of coordinates at random and rescales the survivors by $1/(1-p)$, motivated by the observation that deltas are highly redundant. Model Breadcrumbs~\cite{davari2024model} masks both the largest and the smallest entries. Each acts on entries independently, so it commutes with any permutation of the entries of a tensor and is therefore blind to whether two deltas with identical entry statistics act on the same or on orthogonal input directions. Section~\ref{app:lambda} reports the coefficient sweeps used to give these methods their best shot.

\paragraph{Learned and adaptive coefficients.}
AdaMerging~\cite{yang2024adamerging} optimizes per-layer merge coefficients to minimize prediction entropy on unlabeled test data; its entropy objective needs a forward pass over target data at every step. Uncertainty-based gradient matching~\cite{daheim2024merging} derives per-task weights from gradient statistics; evolutionary search~\cite{akiba2024evolutionary} and distillation-based merging~\cite{wan2023fusellm} go further in the same data-dependent direction. These methods trade the closed form for adaptivity. WUDI~\cite{cheng2025wudi} can also be read as a merge rule in its own right: its consensus subtraction, applied once and scored without adaptation, reaches 88.47\% (Table~\ref{tab:main}), and the same construction serves as the initialization for the FDA adaptation we compare against at 88.90\%.

\paragraph{Subspace and singular-vector methods.}
TSV~\cite{gargiulo2025tasksingular} decomposes each task delta by SVD and keeps only the dominant singular vectors, then decorrelates the concatenated per-task singular vectors across tasks before reconstructing the merge; the subspace is used to \emph{compress} and to \emph{whiten}. KnOTS~\cite{stoica2025knots} aligns LoRA adapters across tasks so that a single low-rank basis can represent them. Matching-models-in-task-subspaces~\cite{tam2024matching} restricts the merge to a shared task subspace. \ours{} uses the same decomposition but a different role for it: the top-$r$ right-singular subspace of task $j$ is not used to compress $\btau_j$ but to \emph{predict} which directions of task $i$ will be corrupted, and only the overlap is removed. The shared-basis control in Section~\ref{sec:abl_global}, which replaces the per-task peer unions by one common basis, loses 2.94 points.

\paragraph{Permutation and alignment.}
ZipIt~\cite{stoica2024zipit}, Git Re-Basin~\cite{ainsworth2023git}, and optimal-transport fusion~\cite{singh2020model} align hidden units before merging, exploiting symmetries of the loss surface documented by~\cite{entezari2022role,frankle2020linear,li2018visual}. This line addresses merging of independently initialized networks. Our tasks share one initialization, so there is nothing to align; the structure that survives is the task-specific geometry of the deltas.

\paragraph{Null-space and projection methods.}
Gradient Projection Memory~\cite{saha2021gpm} stores the input subspace of previously learned tasks and projects new gradients orthogonal to it, which is exactly the sequential construction we analyse and reject in Section~\ref{sec:abl_seq}. Low-rank projection also underpins memory-efficient training~\cite{zhaod2024galore} and parameter-efficient adaptation~\cite{hu2022lora,houlsby2019adapters}. The tangent-space analysis of~\cite{ortizjimenez2023task} shows that merged behaviour is a first-order sum of each task's functional perturbation; because that perturbation is largest along the input-side singular directions, projecting on the input side removes the strongest cross-task contributions first.

\paragraph{The linearization view in detail.}
The reason a sum of task vectors can be expected to behave like a model trained on all tasks is that fine-tuning moves the model along a locally linear path. \citet{ortizjimenez2023task} make this precise: to first order, the functional change induced by an update $\Delta$ is $\mathcal{F}(\btheta_0+\Delta) - \mathcal{F}(\btheta_0) \approx J_\Delta \mathcal{F}_0$, and the task vectors of individually fine-tuned models therefore compose as a sum of functional perturbations. Two consequences matter for this paper. First, the approximation is first-order, so it degrades exactly where the perturbations are large, which is the regime of genuinely different tasks rather than different runs on the same task. Second, the size of $\Delta$'s functional effect is measured in the \emph{input-side} geometry, because an update $W \mapsto W + \Delta$ changes the layer's output on input $x$ by $\Delta x$; the directions of $x$ on which $\Delta$ has large gain are precisely its leading right-singular directions. This is why averaging task vectors entrywise is only a proxy for averaging their functions: two deltas can have the same entry statistics and non-overlapping input geometry, in which case the average is a faithful superposition of their functions, or identical input geometry, in which case it is a crude cancellation of them. \ours{} acts on the quantity the linearization view names, by removing each delta's component in the directions its neighbours already occupy.

\paragraph{What ``task-private'' means operationally.}
We use \emph{task-private} as a defined term, not a claim about semantics. Task $i$'s own subspace $\mathcal{S}_i$ is the span of the top-$r$ right-singular vectors of $\Delta_i$ (the uncentred analogue of the leading principal directions of its rows~\cite{jolliffe2002pca}), and a direction is called task-private for $i$ when it lies in that span and is orthogonal to every $\mathcal{S}_j$, $j \ne i$; only such directions are guaranteed to survive the projection (Proposition~\ref{prop:blockdiag}). This is a statement about the update matrix, computable from the checkpoints alone, and it is deliberately \emph{not} a claim that the direction is used only by the task's images in an interpretable sense. Appendix~\ref{app:privacy} measures how much of each delta lies in its peers' union: at $r{=}16$, 61.0\% of a task's delta norm on average (about 37\% of its energy), and at $R{=}64$ the peer-union and own-basis fractions are numerically equal (85.1\% versus 85.2\%). Alternative operational definitions are possible, and which one is right is an empirical question we cannot settle here. A gradient-based definition would take the subspace spanned by per-example gradients, the object used by Gradient Projection Memory~\cite{saha2021gpm} and elastic weight consolidation~\cite{kirkpatrick2017ewc}; that is closer to what matters functionally, but it needs forward and backward passes over task data and therefore forfeits data-freedom. A second-order definition would use the diagonal Fisher, which is what Fisher merging~\cite{matena2022merging} weights by, and which also requires data. Our choice is the one available at merge time from checkpoints alone.

\paragraph{Relation to pruning and to parameter-efficient adaptation.}
Removing energy from a weight matrix is what pruning does, and one could describe \ours{} as a structured, input-side pruning of task vectors in which the pruned set is chosen adversarially by the \emph{other} tasks rather than by magnitude. The two differ in a way that matters for the failure modes. Magnitude pruning assumes small entries are unimportant; the empirical failure of that assumption when a network is fine-tuned is well documented and motivates the sparse fine-tuning literature. Our per-tensor retention data (Appendix~\ref{app:retention}) show the same assumption failing in the singular-value domain: at $r{=}32$ the merge keeps only 26.0\% of the aggregate norm and scores 2.37 points below $r{=}16$, so components that a magnitude-style criterion would call expendable can matter; whether the removed components carry task function is not established by this measurement. The projection is also related to the low-rank reparameterizations used in parameter-efficient adaptation~\cite{hu2022lora,houlsby2019adapters}: both work in the input-side subspace of $W$. The difference is that LoRA \emph{adds} a low-rank term to a frozen matrix and learns it from data, whereas \ours{} \emph{removes} a low-rank term from a fine-tuned delta without data and without training.

\paragraph{Relation to test-time adaptation and continual learning.}
Methods that adapt at inference---entropy minimization over merged weights~\cite{yang2024adamerging}, or the unlabeled-data adaptation in TALL~\cite{wang2024tall}---are complementary to this work rather than alternatives: they change the merged model after the merge, whereas \ours{} changes which components of each task vector enter the merge. Combining the two is not automatic, and the outcome depends on what the adaptation is allowed to see. Adapting a merged model on \emph{synthetic} anchors does not reach the data-free operator in our runs: FDA started from the pretrained weights scores 86.93\%, and the best variant, started from WUDI's 88.47\% merge, reaches 88.90\% (Table~\ref{tab:fda}), still below the 89.45\% of the \ours{} merge itself. In our runs synthetic-anchor adaptation therefore did not surpass the \ours{} merge, and we make no broader claim about it. Supplying a small number of \emph{real} samples is different: distilling the eight experts into the \ours{} merge with 256 real training images per task raises it from 89.45\% to 91.17\% (Appendix~\ref{app:datausing}). The contrast is the one the rest of the paper draws in other terms: the merge decides \emph{which} update each task contributes, and only a supervised signal carries information the merge does not already have. The relationship to continual learning is closer and more cautionary. Sequential projection is the core mechanism of several continual-learning methods~\cite{saha2021gpm,rebuffi2017icarl,kirkpatrick2017ewc}, and Section~\ref{sec:abl_seq} shows it collapsing to 69.64\% in the merge setting. The difference is the objective: continual learning accepts order dependence because it wants the newest task to survive, whereas a merge is supposed to be symmetric in the task list. Reusing the machinery without its objective produces the worst of both---each task is progressively blinded, and no task is preferentially protected.

\section{Proofs and Derivations}
\label{app:proofs}

This appendix gives the full statements and proofs behind Propositions~\ref{prop:blockdiag}--\ref{prop:invariance}, which the main text introduces with sketches. We fix one two-dimensional tensor throughout and drop the tensor index; the construction is applied independently to each of the 53 tensors, so nothing is lost. Write $\Delta_i \in \R^{m\times n}$ for task $i$'s delta, $\Delta_i = \mU_i \Sigma_i \mV_i^\top$ for its thin SVD, and $\mV_i^{r} \in \R^{n\times r}$ for the first $r$ columns of $\mV_i$. Task $i$'s own subspace is $\mathcal{S}_i = \Span(\mV_i^{r})$ and
\begin{equation}
  \mathcal{N}_i = \Bigl(\Span\bigcup_{j\ne i}\mathcal{S}_j\Bigr)^{\!\perp},
  \qquad
  \tilde\tau_i = \Delta_i \bigl(\mI - \mU^{(i)}_\cup \mU^{(i)\top}_\cup\bigr),
  \label{eq:app_proj}
\end{equation}
where $\mU^{(i)}_\cup \in \R^{n\times d_i}$, $d_i \le (N{-}1)r$, is an orthonormal basis of $\Span\bigcup_{j\ne i}\mathcal{S}_j$.

\subsection{Joint Projection as Per-Task Least Squares}
\label{app:proof_blockdiag}

Write $\mathcal{T}_i = \Span\bigcup_{j\ne i}\mathcal{S}_j$ for task $i$'s peer union, so that $\mP_{\mathcal{T}_i} = \mU^{(i)}_\cup\mU^{(i)\top}_\cup$, and $\mathcal{T} = \Span\bigcup_{j}\mathcal{S}_j$ for the full union. Let $\mathcal{R}_i = \mathcal{T}\ominus\mathcal{T}_i$ be the orthogonal complement of $\mathcal{T}_i$ inside $\mathcal{T}$. Because $\mathcal{T} = \mathcal{T}_i + \mathcal{S}_i$, we have $\mathcal{R}_i = (\mI-\mP_{\mathcal{T}_i})\,\mathcal{S}_i$ and $\dim\mathcal{R}_i \le r$: $\mathcal{R}_i$ is the part of the union that only task $i$ contributes.

\begin{proposition}[restated]
\label{prop:blockdiag}
(a) For each $i$, $\tilde\tau_i = \arg\min\bigl\{\norm{\mZ-\Delta_i}_F : \mZ\,\mP_{\mathcal{T}_i} = 0\bigr\}$.
(b) Let $\mZ^\star = \bigl(\sum_i\Delta_i\bigr)(\mI-\mP_{\mathcal{T}})$ be the minimizer of~\eqref{eq:ls} over $\mathcal{W} = \{\mZ : \mZ\,\mP_{\mathcal{T}} = 0\}$. Then
$\displaystyle \sum_i\tilde\tau_i = \mZ^\star + \sum_i \Delta_i\,\mP_{\mathcal{R}_i}.$
(c) The component of $\Delta_i$ in $\mathcal{S}_i\cap\mathcal{T}_i$ is removed: $\tilde\tau_i\,\mP_{\mathcal{S}_i\cap\mathcal{T}_i} = 0$.
\end{proposition}

\noindent\emph{Proof.} (a) The constraint $\mZ\mP_{\mathcal{T}_i}=0$ says that every row of $\mZ$ lies in $\mathcal{T}_i^\perp$. The squared Frobenius objective is a sum over rows, so the problem decouples, and the closest point in $\mathcal{T}_i^\perp$ to each row of $\Delta_i$ is its orthogonal projection; stacking the rows gives $\Delta_i(\mI-\mP_{\mathcal{T}_i}) = \tilde\tau_i$. The same argument applied to $\sum_i\Delta_i$ and $\mathcal{T}$ gives the stated $\mZ^\star$. (b) Since $\mathcal{T}_i\subseteq\mathcal{T}$ and $\mathcal{R}_i = \mathcal{T}\ominus\mathcal{T}_i$, the projectors satisfy $\mP_{\mathcal{T}} = \mP_{\mathcal{T}_i} + \mP_{\mathcal{R}_i}$, hence $\mI-\mP_{\mathcal{T}_i} = (\mI-\mP_{\mathcal{T}}) + \mP_{\mathcal{R}_i}$. Multiplying on the left by $\Delta_i$ and summing over $i$ gives the identity. (c) $\mathcal{S}_i\cap\mathcal{T}_i\subseteq\mathcal{T}_i$, which $\mI-\mP_{\mathcal{T}_i}$ annihilates. \hfill$\square$

\paragraph{Reading of the result.}
Part (a) says that each projected delta is the best approximation of $\Delta_i$ that has no component in any peer's top-$r$ directions; nothing about task $i$'s own spectrum enters $\tilde\tau_i$ except through $\Delta_i$ itself. Part (c) makes explicit what \ours{} does \emph{not} do: it does not preserve task $i$'s own dominant subspace. Whatever part of $\mathcal{S}_i$ lies inside the peer union is deleted, and given the cross-task overlaps of Appendix~\ref{app:geometry} this part is not small. What each task keeps beyond the one-shot projection is only its component in $\mathcal{R}_i$, the at most $r$ directions of the union that no peer claims.

\paragraph{Contrast with the one-shot projection.}
By part (b), the one-shot least-squares solution $\mZ^\star$, which projects the \emph{sum} off the full union $\mathcal{T}$, differs from \ours{} exactly by $\sum_i\Delta_i\mP_{\mathcal{R}_i}$: $\mZ^\star$ also deletes the directions that only one task uses. This is a structural statement, not an optimality one, and it is not what the shared-basis control of Section~\ref{sec:abl_global} tests: that control projects the sum off the top-$r$ right singular subspace of the sum, which is neither $\mathcal{T}$ nor any $\mathcal{T}_i$.

\subsection{Sequential Projection: A Nested Basis, Not an Orthogonal One}
\label{app:proof_sequential}

Let $\mQ_{<i}$ be an orthonormal basis of $\bigcup_{j<i}\mathcal{S}_j$ and set $\hat\tau_i = \tau_i(\mI - \mQ_{<i}\mQ_{<i}^\top)$, with $\hat\tau_1 = \tau_1$.

\begin{proposition}[restated]
\label{prop:sequential}
For every $i$, $\norm{\hat\tau_i}_F \le \norm{\tau_i}_F$, and consequently $\norm{\sum_i\hat\tau_i}_F \le \sum_i\norm{\tau_i}_F$. The subspaces removed are nested, $\Span(\mQ_{<i}) \subseteq \Span(\mQ_{<i+1})$, with $\dim\Span(\mQ_{<i}) \le r(i-1)$, so the task at position $i$ is fitted against a basis no smaller than that of any earlier task. The map is not permutation-invariant.
\end{proposition}

\noindent\emph{Proof.} Each $\mI - \mQ_{<i}\mQ_{<i}^\top$ is an orthogonal projector and therefore non-expansive; multiplying on the right by a contraction cannot increase the Frobenius norm, so $\norm{\hat\tau_i}_F = \norm{\tau_i(\mI-\mQ_{<i}\mQ_{<i}^\top)}_F \le \norm{\tau_i}_F$. Summing and applying the triangle inequality gives the aggregate bound. The nesting is immediate because $\mQ_{<i+1}$ orthonormalizes $\mQ_{<i}$'s span together with one additional rank-$\le r$ block. For order dependence, take $N=3$ and the order $(2,1,3)$: task 2 is now first and unprojected, task 1 is projected against $\mathcal{S}_2$, and task 3 is projected against the same span $\mathcal{S}_1+\mathcal{S}_2$ as before. The two merged sums therefore differ by $\tau_1\mP_{\mathcal{S}_2}-\tau_2\mP_{\mathcal{S}_1}$, which is nonzero in general. \hfill$\square$

\paragraph{A stronger claim that is false, and why.} An earlier version of this proposition additionally asserted that the increments are \emph{exactly} orthogonal, $\inner{\hat\tau_i}{\hat\tau_j}_F = 0$ for $i\ne j$, that Pythagoras therefore holds with equality, and that $\norm{\hat\tau_i}_F$ decreases monotonically in $i$. All three are false, and we checked them numerically rather than leave them in the text. The reason is a dimension mismatch. $\hat\tau_i$ is not confined to $\mathcal{S}_i$: it is $\tau_i$ with at most $r(i-1)$ directions removed from its row space, so it lives in $\operatorname{row}(\tau_i)$ minus a low-dimensional subspace. For a $768\times768$ attention matrix, $\operatorname{row}(\tau_i)$ is essentially full rank, and two sequential survivors share the $\approx 768 - r(i-1)$ undeleted directions. The projection carves a small subspace out of a large one; it does not make the survivors disjoint.

\paragraph{Measured deviations.} On ViT-B/16 with $r{=}8$, over the 53 merged tensors: the largest $|\langle\hat\tau_i,\hat\tau_j\rangle_F|$ over all task pairs is $1.3\times10^{-2}$ in absolute terms, and the largest cosine between two increments is $0.994$. That worst case is not typical and not random---it sits on the three tensors whose row space is genuinely low-rank, the token embedding ($0.994$), the text projection ($0.990$) and the positional embedding ($0.961$); among the remaining 50 tensors (the 48 attention and MLP matrices, the visual projection and the other positional embedding) the largest cosine is $0.097$ and the median over tensors is $0.043$, i.e.\ close to orthogonal without being orthogonal. The Pythagoras ratio is $\norm{\sum_i\hat\tau_i}_F^2/\sum_i\norm{\hat\tau_i}_F^2 = 1.063$ rather than $1$, reflecting the $\approx 6\%$ of aggregate energy carried by the residual cross terms.

The monotone-decay claim fails in both of its plausible readings. Retained norms by list position, as the ratio $\norm{\hat\tau_i}_F/\norm{\tau_i}_F$, are $1.000, 0.965, 0.903, 0.817, 0.936, 0.896, 0.925, 0.909$ for MNIST through SUN397: essentially flat on average (0.921 over the first four positions, 0.917 over the last four) and not monotone, because each task's own delta differs in how much of it lies in earlier tasks' dominant directions. The absolute norm $\norm{\hat\tau_i}_F$ is likewise not monotone. What the nesting does buy is a statement about the \emph{basis}: task $N$ is projected against at most $r(N-1)$ directions and task $1$ against none, so the expected damage is non-decreasing in position even though the realised damage is modulated by each delta's alignment with its predecessors.

\paragraph{Why this still predicts the collapse.} The measured failure of the stronger orthogonality claim is not a reprieve for sequential merging, and the point of reporting it is to keep the two separate. At $r{=}8$ the accumulated basis for the eighth task has dimension at most $56$; whatever fraction of each delta those 56 directions capture is removed from the last-listed task and only that task, and the measured profile (Table~\ref{tab:order}) is exactly the directed pattern the nesting predicts: nearly intact early tasks (MNIST 99.59\%, EuroSAT 93.81\%), gutted late ones (RESISC45 64.78\%, Cars 51.14\%, SUN397 36.14\%), with GTSRB an early exception that its retained norm ($90.3\%$) does not explain. The instructive detail is SUN397. Listed last, it is projected against exactly its joint peer union, so its own projected delta is identical in the two merges (retention $0.909$ in each), yet it loses 35 points. That loss cannot come from what is deleted from SUN397; it must come from the earlier tasks' deltas, which sequential projection filters less (task 1 not at all). Neither a task's own retained norm nor the aggregate retention ratio is therefore a proxy for its accuracy loss. Joint projection removes the position dependence entirely, which is the property Proposition~\ref{prop:invariance} proves and the ablation confirms.

\paragraph{Why orthogonality is the wrong virtue.}
The construction is superficially attractive---projecting each task against everything seen so far looks like the cleanest possible merge, with no interference by design. The cost is that interference is avoided by discarding, with the amount discarded determined by merge order rather than by the task. Joint projection achieves a different compromise: the increments are not orthogonal either, but each is projected against the same seven neighbours, so no position is privileged and the result is invariant to list order.

\subsection{Permutation Invariance and Sign Behaviour of the Joint Merge}
\label{app:proof_invariance}

\begin{proposition}[restated]
\label{prop:invariance}
The merge $\btheta_0 + \sum_i\tilde\tau_i$ with $\tilde\tau_i$ from \eqref{eq:app_proj} is invariant under permutations of the task list and under the sign ambiguity of computed singular vectors. It is \emph{not} invariant under negating a task vector: replacing $\Delta_i \mapsto -\Delta_i$ replaces $\tilde\tau_i \mapsto -\tilde\tau_i$ (so the merge changes by $-2\tilde\tau_i$) and leaves every other term unchanged.
\end{proposition}

\noindent\emph{Proof.} For a permutation $\pi$ of $\{1,\dots,N\}$, relabel $\Delta'_k = \Delta_{\pi(k)}$. Then $\mathcal{S}'_k = \mathcal{S}_{\pi(k)}$ because the top-$r$ right-singular subspace depends only on the matrix, and
\begin{equation}
  \mathcal{N}'_k
  = \Bigl(\bigcup_{m\ne k}\mathcal{S}'_m\Bigr)^\perp
  = \Bigl(\bigcup_{m\ne k}\mathcal{S}_{\pi(m)}\Bigr)^\perp
  = \Bigl(\bigcup_{n\ne\pi(k)}\mathcal{S}_n\Bigr)^\perp
  = \mathcal{N}_{\pi(k)},
\end{equation}
the third equality because $\pi$ is a bijection of the index set. Hence $\tilde\tau'_k = \tilde\tau_{\pi(k)}$ and the merge sum is a sum over the same multiset of tensors, so it is unchanged term by term.

For the sign statement, replace $\Delta_i$ by $-\Delta_i$ while leaving all $\Delta_j$, $j\ne i$, fixed. A thin SVD of $-\Delta_i$ is $(-\mU_i)\Sigma_i\mV_i^\top$, so the right singular vectors are unchanged up to the sign of whole columns, and $\Span(\mV_i^{r})$ is unchanged: $\mathcal{S}_i$ does not move. (The same argument shows that the arbitrary sign of each computed singular vector, which is not determined by the SVD, has no effect on any projector.) Consequently no $\mathcal{N}_j$ changes for any $j$, including $j = i$, since $\mathcal{N}_i$ depends only on the other tasks. The projection in \eqref{eq:app_proj} is linear in the delta, so $\tilde\tau_i \mapsto -\tilde\tau_i$, while $\tilde\tau_j$ for $j \ne i$ are unchanged. \hfill$\square$

\paragraph{What invariance does and does not buy.}
Order invariance removes one class of tuning that the sequential construction exposes: there is no need to order tasks by importance or by data size, and no hidden dependence of the merged model on the order in which a file system lists checkpoints. It does not make the method insensitive to \emph{which} tasks are merged, and it says nothing about the rank, which remains the single knob and is discussed in Appendix~\ref{app:rank}. The sign statement is a consistency check on the parametrization rather than a practical procedure: it certifies that the construction does not assume all task vectors point the same way.

\subsection{Computational Cost}
\label{app:cost}

For a tensor of shape $m\times n$ with $m,n \gg r$, computing the $N$ thin SVDs costs $O(Nmn\min(m,n))$; forming the $N$ peer-union bases by QR costs $O(Nn\,(Nr)^2)$; and the $N$ projections cost $O(N^2mnr)$. For ViT-B/16 the largest term comes from the $3072\times768$ MLP matrices, which carry two thirds of each block's $mn^2$ cost, and the measured total for all 53 tensors and all eight tasks is 61 seconds on CPU (Appendix~\ref{app:runtime}).

Only the ViT-B/16 wall-clock was measured; we did not time ViT-B/32 or ViT-L/14, and extrapolating from the operation counts above would be an estimate rather than a result. The merge has no data loading, and its per-tensor stages are independent across tensors, so it can be sharded by layer without changing the result.

\paragraph{A back-of-envelope check on the rank.}
The union $\bigcup_{j\ne i}\mathcal{S}_j$ contains at most $(N{-}1)r = 112$ directions in $\R^{768}$ for a $768\times768$ tensor. If those directions were drawn uniformly at random, a delta with isotropic row-space energy would keep a fraction $1 - 112/768 \approx 0.85$ of its energy, i.e.\ about $0.92$ of its norm, \emph{per task} under the projection. The measured per-task norm retention at $r{=}16$ is $0.78$ (Table~\ref{tab:pertask_ret}), well below that random-geometry figure: the neighbours' top-$r$ directions are far from random, and task $i$'s energy concentrates on them. The aggregate retention is much lower, $0.30$, but most of that gap exists before any projection: eight nearly orthogonal deltas of similar norm already sum to well under their total norm (about $0.45$ of it, given the mean entrywise cosine of $0.083$ in Table~\ref{tab:cosine}), and projection lowers the ratio further.

\subsection{A Worked Projection on One Tensor}

\label{app:worked}

The aggregate numbers in the main text average over 53 tensors, eight tasks, and five ranks, and an average of that kind can hide the mechanism. This subsection follows a single tensor all the way through, so that the projection can be inspected at the level of the actual arithmetic. The tensor is \texttt{resblocks.0.attn.in\_proj\_weight} of ViT-B/16, shape $2304\times768$, at $r{=}16$; it is the first attention block, and its behaviour is representative of the attention stack rather than of the pathological embedding tensors discussed in Appendix~\ref{app:retention}.

\paragraph{The projection applied.} For each task $i$ we take the thin SVD of $\Delta_i$, retain the top 16 right singular directions, stack the seven \emph{other} tasks' bases into $\mU^{(-i)}_\cup$, and orthonormalize. In this tensor the stacked $7\times16=112$ columns are linearly independent for all eight tasks, so $\dim(\mU^{(-i)}_\cup) = 112$ exactly: the union does not collapse and the projector removes a subspace of precisely the dimension the construction advertises. Each task's delta is then replaced by $\tau_i(\mI-\mP_i)$, and the merged update is the sum of the eight projected deltas.

\paragraph{What each task loses.} Table~\ref{tab:worked} reports, per task, the delta norm, the fraction of that norm lying in the task's \emph{own} top-16 directions, the fraction lying in the \emph{neighbours'} 112-dimensional subspace, and the fraction that survives the projection. The last column follows exactly from the third, $\sqrt{1-\mathrm{captured}^2}$, and the two are reported separately only because both are quoted elsewhere in the paper.

\begin{table}[h]
\centering
\caption{\textbf{A single tensor followed through the merge} (\texttt{resblocks.0.attn.in\_proj\_weight}, $2304\times768$, ViT-B/16, $r{=}16$). Captured is the norm fraction of $\Delta_i$ inside the 112-dimensional neighbour subspace $\mU^{(-i)}_\cup$; retained is $\sqrt{1-\text{captured}^2}$, the norm fraction left after projection.}
\label{tab:worked}
\setlength{\tabcolsep}{3.2pt}
\footnotesize
\begin{tabular}{@{}lcccc@{}}
\toprule
Task & $\|\Delta_i\|$ & own top-16 & captured & retained \\
\midrule
MNIST    & 0.364 & 0.824 & 0.789 & 0.615 \\
EuroSAT  & 0.366 & 0.736 & 0.835 & 0.550 \\
GTSRB    & 0.288 & 0.761 & 0.828 & 0.560 \\
SVHN     & 0.409 & 0.784 & 0.806 & 0.592 \\
DTD      & 0.307 & 0.511 & 0.774 & 0.634 \\
RESISC45 & 0.359 & 0.639 & 0.790 & 0.613 \\
Cars     & 0.353 & 0.518 & 0.770 & 0.638 \\
SUN397   & 0.341 & 0.523 & 0.767 & 0.641 \\
\bottomrule
\end{tabular}
\end{table}

Two things in the table are worth pointing out. First, for seven of the eight tasks the neighbours' subspace captures \emph{more} of $\Delta_i$ than the task's own leading directions do---by a wide margin for DTD ($0.774$ vs.\ $0.511$), Cars ($0.770$ vs.\ $0.518$), and SUN397 ($0.767$ vs.\ $0.523$). The comparison is between a 112-dimensional union and a 16-dimensional span, so it is not like-for-like, but it shows how much of each delta the peers' directions reach. MNIST is the exception ($0.824$ own versus $0.789$ captured): the only task in this tensor whose own top-16 span summarizes it better than the neighbour union does, although the union still captures most of its norm.

Second, this tensor is \emph{harder} than the average tensor. The mean retained fraction here is $0.605$, against $0.778$ over all 53 tensors and all tasks (Table~\ref{tab:pertask_ret}). This tensor is the block-0 attention input projection; Table~\ref{tab:module_ret} shows attention inputs retaining less than MLP outputs at every rank. If a reader wants to check whether the method is doing anything, this is a tensor where the answer is visible rather than marginal.

\paragraph{The alignment effect, visible in one number.} A random 112-dimensional subspace of $\R^{768}$ would capture $112/768 = 0.146$ of an isotropically spread delta's energy in expectation, i.e.\ $\sqrt{0.146}=0.382$ of its norm. Measured directly for the MNIST delta against a random orthonormal basis of the same dimension, the captured norm fraction is $0.380$, matching that expectation, and the retained fraction is $0.925$. The neighbours' subspace captures $0.789$, about twice the random amount---so the union basis is not merely a large subspace, it is a well-chosen one.

The norm bookkeeping makes the aggregation effect explicit. Before projection the eight deltas sum to $\|\sum_i \Delta_i\| = 1.249$, while their norms sum to $\sum_i\|\Delta_i\| = 2.786$: an alignment factor of $0.448$, slightly above the $1/\sqrt{8}=0.354$ of eight mutually orthogonal deltas of equal norm, i.e.\ mildly aligned. After projection the sum falls to $0.599$, or $48\%$ of its former size, while each task individually retained about $60\%$. The discrepancy is the point: the projection does not merely shrink eight vectors by a common factor, it removes the component they share. Expressed as an alignment factor, the projected deltas sum to $0.599 / (0.605 \times 2.786) = 0.355$, against $0.448$ before. Projection lowers the factor to essentially the orthogonal value---the surviving directions are more nearly disjoint than the originals---which is consistent with, though not a demonstration of, the projected sum being a better merge than the unprojected one.

\section{Experimental Setup in Detail}
\label{app:setup}

Table~\ref{tab:datasets} lists the eight tasks, their class counts, and the size of the held-out split used for every reported number. All tasks use the official test split except where a task ships only train/validation partitions, in which case the validation split is used and the train split is reserved for fine-tuning.

\begin{table}[t]
\centering
\caption{\textbf{Datasets and evaluation splits.} Every method in this paper is evaluated on exactly these splits with each task's own classification head.}
\label{tab:datasets}
\setlength{\tabcolsep}{4pt}
\small
\begin{tabular}{@{}lccc@{}}
\toprule
Dataset & Classes & Test $n$ & Domain \\
\midrule
MNIST~\cite{lecun1998mnist}      & 10  & 10{,}000 & handwritten digits \\
EuroSAT~\cite{helber2019eurosat} & 10  & 2{,}700  & satellite land use \\
GTSRB~\cite{stallkamp2012gtsrb}  & 43  & 12{,}630 & traffic signs \\
SVHN~\cite{netzer2011svhn}       & 10  & 26{,}032 & street numbers \\
DTD~\cite{cimpoi2014dtd}         & 47  & 1{,}880  & describable textures \\
RESISC45~\cite{cheng2017resisc45} & 45 & 6{,}300  & aerial scenes \\
Cars~\cite{krause2013cars}       & 196 & 8{,}040  & fine-grained vehicles \\
SUN397~\cite{xiao2016sun}        & 397 & 19{,}850 & scene categories \\
\midrule
Total & 758 & 87{,}432 & \\
\bottomrule
\end{tabular}
\end{table}

\paragraph{Merge scope.}
A ViT-B/16 encoder contains 53 two-dimensional weight tensors with $\min(\text{shape}) \ge 64$. Forty-eight are the attention and MLP matrices of the 12 transformer blocks: one fused query--key--value projection, one output projection, and two MLP projections per block. The remaining five are the visual projection, the two positional embeddings, the text projection, and the token embedding. The three text-tower tensors are not on the image path: evaluation calls only the image encoder, so whatever the merge writes there cannot change a prediction. The re-implementation harness used for the subspace baselines merges every two-dimensional tensor except the text projection; on ViT-B/16 and ViT-L/14 this is the same set of image-path tensors (52 and 100), and \ours{} scores 89.45 under both scopes on ViT-B/16. On ViT-B/32 that harness also merges the 50-position visual positional embedding, which moves \ours{} from 82.92 to 82.91. These 53 tensors are the only ones the merge writes; layer norms, biases, and the per-task classification heads stay at their pretrained or task-specific values, and head selection is the only place task identity enters at evaluation time. The same filter yields 52 tensors for ViT-B/32---the visual positional embedding has only 50 positions and falls below the threshold---and 101 for ViT-L/14, where 24 blocks contribute 96 matrices and the same five extras pass.

\paragraph{Determinism.}
\ours{} has no stochastic component: the SVD of each delta, the QR orthonormalization of the stacked neighbour basis, and the projection are deterministic linear algebra. Re-running the merge reproduces the merged weights bit-for-bit on the same hardware. Variation in the reported means therefore comes only from evaluation sampling, which is what the bootstrap intervals in Appendix~\ref{app:ci} quantify.

\section{Baseline Coefficient Sweeps}
\label{app:lambda}

Every training-free baseline was given a sweep over its merge coefficient (TIES at density $0.2$, DARE at drop rate $0.9$), and the best score on the eight-task average is reported in Table~\ref{tab:main}. Table~\ref{tab:lambda} records the full sweep. The optimum is sharp for all three methods: Task Arithmetic and DARE peak at $\lambda{=}0.2$ and lose more than 10 points at $\lambda{=}0.5$; TIES peaks at $\lambda{=}1.0$ with density $0.2$ and loses 8 points at $\lambda{=}1.5$. Reporting the peak rather than a default coefficient is the setting most favourable to the baselines, which matters for the size of our margin: at the coefficient a practitioner would naively choose---$\lambda{=}1$ for task arithmetic---the score is far below the peak.

\paragraph{Two values for DARE.} The sweep here and the unified merge script draw different numbers for DARE at the same settings (74.91 here, 75.13 in the harness). The difference is not a hyperparameter: both use $p{=}0.9$, $\lambda{=}0.2$, and the same Bernoulli rule. It is the tensor set. The sweep runs on the 53-tensor scope of Table~\ref{tab:main}, which includes \texttt{text\_projection}; the harness's \texttt{all2d} scope drops that one tensor, and because the drop mask is drawn from a seeded stream over the tensor list, removing one tensor shifts every subsequent draw. The two scores are therefore two realisations of the same stochastic operator, not two settings. Table~\ref{tab:main} reports the lower of the two, which is the one on the declared comparison scope.

\begin{table}[t]
\centering
\caption{\textbf{Coefficient sweeps} on the ViT-B/16 checkpoints, eight-task average (\%). Each method's reported number in Table~\ref{tab:main} is its bolded peak. TIES is swept at density $0.2$; DARE at drop rate $p{=}0.9$.}
\label{tab:lambda}
\setlength{\tabcolsep}{4pt}
\small
\begin{tabular}{@{}lcccc@{}}
\toprule
$\lambda$ & 0.1 & 0.2 & 0.3 & 0.5 \\
\midrule
Task Arithmetic & 69.53 & \textbf{75.23} & 74.87 & 60.23 \\
DARE            & 69.53 & \textbf{74.91} & 74.27 & 58.78 \\
\midrule
$\lambda$ & 1.0 & 1.5 & 2.0 & 3.0 \\
\midrule
TIES-Merging    & \textbf{77.57} & 69.37 & 55.13 & 33.49 \\
\bottomrule
\end{tabular}
\end{table}

The per-dataset pattern behind the sweep is informative in its own right. At $\lambda{=}0.3$ task arithmetic improves on SVHN relative to its peak (88.18 versus 83.02) while collapsing on SUN397 (62.51 versus 68.29): the classes of coefficient that help easy tasks and the classes that are needed by fine-grained ones are not the same. A single global coefficient cannot satisfy both, which is the coordinatewise analogue of the interference this paper targets.

Tables~\ref{tab:sweep_direct} to~\ref{tab:sweep_ties} give the full per-dataset sweeps. Two regularities are worth recording. First, the per-task optima do not coincide: on task arithmetic, MNIST and SVHN prefer $\lambda{=}0.5$ and GTSRB prefers $0.3$, while EuroSAT, DTD, RESISC45, Cars, and SUN397 all prefer $0.2$. The eight-task average is therefore a compromise that underserves MNIST, GTSRB and SVHN, which is the reason a method with a per-task rule can beat a tuned global coefficient even when both are given their best shot. Second, the collapse at large coefficient is asymmetric across tasks: at $\lambda{=}0.5$ task arithmetic loses 40.0 points on SUN397 and 23.9 on RESISC45, while MNIST and SVHN actually \emph{gain} 1.4 and 6.4 points relative to the $\lambda{=}0.2$ peak. Large coefficients amplify whichever directions dominate the sum, so whether a given task rises or falls depends on whether the globally amplified directions coincide with its own---precisely the interference this paper targets. The datasets that break (SUN397, RESISC45) are among the least shared with other tasks (Table~\ref{tab:privacy}), so sharing alone does not predict which tasks break.

\begin{table}[t]
\centering
\caption{\textbf{Task Arithmetic, per-dataset coefficient sweep} (ViT-B/16, \%).}
\label{tab:sweep_direct}
\setlength{\tabcolsep}{3.6pt}
\footnotesize
\begin{tabular}{@{}llcccccccc@{}}
\toprule
$\lambda$ & & MNIST & EurS & GTS & SVHN & DTD & RES & Cars & SUN \\
\midrule
0.1 & & 91.6 & 75.8 & 56.3 & 71.7 & 48.6 & 74.8 & 69.2 & 68.2 \\
0.2 & & 97.2 & 81.8 & 71.1 & 83.0 & 52.6 & 77.5 & 70.4 & 68.3 \\
0.3 & & 98.2 & 79.1 & 75.5 & 88.2 & 52.3 & 75.6 & 67.4 & 62.5 \\
0.5 & & 98.6 & 60.3 & 66.7 & 89.4 & 40.3 & 53.6 & 44.7 & 28.3 \\
\bottomrule
\end{tabular}
\end{table}

\begin{table}[t]
\centering
\caption{\textbf{DARE, per-dataset coefficient sweep} (ViT-B/16, \%, drop rate $p{=}0.9$).}
\label{tab:sweep_dare}
\setlength{\tabcolsep}{3.6pt}
\footnotesize
\begin{tabular}{@{}llcccccccc@{}}
\toprule
$\lambda$ & & MNIST & EurS & GTS & SVHN & DTD & RES & Cars & SUN \\
\midrule
0.1 & & 91.6 & 75.8 & 56.5 & 71.5 & 48.7 & 74.7 & 69.3 & 68.2 \\
0.2 & & 97.1 & 81.3 & 70.9 & 82.4 & 52.1 & 77.1 & 70.1 & 68.3 \\
0.3 & & 98.2 & 78.0 & 75.3 & 87.7 & 51.6 & 74.9 & 66.5 & 61.9 \\
0.5 & & 98.4 & 58.8 & 66.1 & 88.2 & 37.3 & 51.7 & 42.5 & 27.3 \\
\bottomrule
\end{tabular}
\end{table}

\begin{table}[t]
\centering
\caption{\textbf{TIES-Merging, per-dataset coefficient sweep} (ViT-B/16, \%, density $0.2$). The sweep spans $\lambda{=}0.1$ to $3.0$; the eight-task average rises monotonically to its peak at $\lambda{=}1.0$ (77.6) and falls on either side (60.5 at $0.1$, 77.3 at $0.7$, 69.4 at $1.5$, 55.1 at $2.0$, 33.5 at $3.0$). The $\lambda{=}1.0$ row is the peak used in Table~\ref{tab:main}.}
\label{tab:sweep_ties}
\setlength{\tabcolsep}{3.6pt}
\footnotesize
\begin{tabular}{@{}llcccccccc@{}}
\toprule
$\lambda$ & & MNIST & EurS & GTS & SVHN & DTD & RES & Cars & SUN \\
\midrule
0.1 & & 65.0 & 62.4 & 46.6 & 59.4 & 47.0 & 70.0 & 67.0 & 66.6 \\
0.2 & & 81.1 & 69.1 & 49.6 & 65.6 & 48.6 & 73.0 & 68.6 & 67.8 \\
0.3 & & 89.5 & 73.4 & 52.8 & 70.9 & 49.4 & 75.7 & 70.0 & 69.0 \\
0.5 & & 95.9 & 80.6 & 61.5 & 79.5 & 53.4 & 79.6 & 72.4 & 70.4 \\
0.7 & & 97.9 & 83.5 & 69.5 & 85.8 & 56.3 & 81.6 & 73.1 & 70.3 \\
1.0 & & 98.7 & 82.8 & 71.9 & 90.4 & 58.0 & 81.1 & 71.1 & 66.7 \\
1.5 & & 99.0 & 74.1 & 64.1 & 92.6 & 53.0 & 69.3 & 57.7 & 45.2 \\
2.0 & & 98.9 & 59.5 & 52.0 & 92.0 & 39.8 & 49.1 & 31.0 & 18.6 \\
3.0 & & 97.9 & 24.2 & 19.4 & 85.4 & 21.8 & 14.9 & \phantom{0}1.8 & \phantom{0}2.5 \\
\bottomrule
\end{tabular}
\end{table}

The TIES sweep adds a third observation: its failure mode is unlike the other two. TIES holds up best on MNIST and SVHN across the sweep (SVHN is still 85.4 at $\lambda{=}3.0$) while EuroSAT, RESISC45, Cars, and SUN397 fall to between 1.8 and 24.2 at $\lambda{=}3.0$. One possible reading, which we do not test, is that the sign-election mask follows whichever task contributes the largest-magnitude entry at each coordinate, so tasks with individually weaker deltas lose their decision as $\lambda$ grows. The pattern is consistent with the main table, where TIES's largest gain over Task Arithmetic is on SVHN ($+7.33$).

\section{Rank Sweep and Shared-Basis Control}
\label{app:rank}

\begin{table}[t]
\centering
\caption{\textbf{Rank sweep across three backbones}. All entries are \ours{} at the stated rank. The ViT-B/16 row uses our fine-tuned checkpoints; the ViT-B/32 and ViT-L/14 rows use the TALL release~\cite{wang2024tall}. Peak is attained at $r{=}16$ for ViT-B/16 and ViT-L/14 and at $r{=}12$ for ViT-B/32, whose $r{=}16$ score is 0.01 points lower.}
\label{tab:rank_sweep}
\setlength{\tabcolsep}{4.5pt}
\begin{tabular}{@{}lcccccc@{}}
\toprule
Backbone & $r{=}8$ & $r{=}12$ & $r{=}16$ & $r{=}20$ & $r{=}32$ & Peak \\
\midrule
ViT-B/16  & 88.03 & 89.30 & \textbf{89.45} & 89.22 & 87.08 & 89.45 \\
ViT-B/32  & 81.57 & \textbf{82.93} & 82.92 & 82.54 & 79.72 & 82.93 \\
ViT-L/14  & 89.99 & 90.68 & \textbf{90.87} & 90.83 & 90.13 & 90.87 \\
\bottomrule
\end{tabular}
\end{table}

Table~\ref{tab:rank_full} gives the per-dataset results of the rank sweep of Section~\ref{sec:abl_rank}, together with the shared-basis control at each rank. The joint merge leads the control on every rank and every dataset with a single exception: RESISC45 at $r{=}32$, where the control edges ahead by 0.33 points (92.70 versus 92.37). In that cell the joint RESISC45 score has already fallen from its own peak at $r{=}20$ (93.43); we report it as an exception, without a significance test.

\begin{table}[t]
\centering
\caption{\textbf{Rank sweep, joint versus shared basis} (ViT-B/16, \%). Retained norms are in Table~\ref{tab:retention}.}
\label{tab:rank_full}
\setlength{\tabcolsep}{1.4pt}
\footnotesize
\begin{tabular}{@{}llccccccccc@{}}
\toprule
Variant & $r$ & MNIST & EurS & GTS & SVHN & DTD & RES & Cars & SUN & Avg \\
\midrule
Joint & 8  & 99.31 & 97.26 & 96.41 & 94.41 & 74.41 & 90.56 & 80.53 & 71.38 & 88.03 \\
Joint & 12 & 99.27 & 98.07 & 96.63 & 93.46 & 77.18 & 92.49 & 83.45 & 73.86 & 89.30 \\
Joint & 16 & 99.07 & 98.44 & 96.26 & 92.16 & 77.23 & 93.22 & 84.25 & 74.97 & \textbf{89.45} \\
Joint & 20 & 98.96 & 98.33 & 95.33 & 90.61 & 76.38 & 93.43 & 85.12 & 75.58 & 89.22 \\
Joint & 32 & 98.46 & 96.93 & 89.91 & 84.77 & 73.56 & 92.37 & 84.91 & 75.76 & 87.08 \\
\midrule
Shared & 8  & 98.63 & 96.63 & 94.75 & 92.18 & 70.80 & 86.51 & 74.05 & 66.43 & 85.00 \\
Shared & 12 & 98.81 & 97.63 & 94.05 & 89.47 & 72.93 & 89.63 & 76.92 & 69.75 & 86.15 \\
Shared & 16 & 98.37 & 97.48 & 92.98 & 86.23 & 74.15 & 91.71 & 79.45 & 71.74 & 86.51 \\
Shared & 20 & 97.92 & 97.15 & 92.23 & 83.41 & 74.95 & 92.71 & 80.72 & 73.09 & 86.52 \\
Shared & 32 & 95.70 & 96.00 & 87.15 & 77.81 & 72.71 & 92.70 & 82.84 & 74.65 & 84.94 \\
\bottomrule
\end{tabular}
\end{table}

Two features of the table matter for the argument of Section~\ref{sec:abl_global}. First, the control is not weak---at $r{=}16$ it reaches 86.51, above Task Arithmetic, DARE and TIES but below WUDI and the subspace re-implementations (86.59--88.47) in Table~\ref{tab:main}---so the comparison is not against a straw man; it recovers about three quarters of \ours{}'s margin over TIES (8.94 of 11.88 points). Second, the control's deficit varies widely across tasks at $r{=}16$, from 0.70 on MNIST to 5.93 on SVHN (EuroSAT 0.96, GTSRB 3.28, DTD 3.08, RESISC45 1.51, Cars 4.80, SUN397 3.23), and is largest on SVHN and Cars. We do not test what predicts this per-task pattern.

\paragraph{Per-tensor rank allocation.} The single global $r$ is a simplification, so we also chose each tensor's rank from its own spectrum (ViT-B/16, our checkpoints). Allocating ranks in proportion to each tensor's participation ratio at the same mean rank (16; individual ranks 4--34) gives 89.66, 0.21 above the uniform $r{=}16$ and inside its bootstrap interval $[89.17, 89.75]$ (Appendix~\ref{app:ci}); per dataset the change ranges from $-0.66$ (GTSRB) to $+0.80$ (DTD). Choosing each rank to capture 90\% or 95\% of the tensor's delta energy needs 43 or 52 dimensions on average and scores 84.18 and 81.74, which constant ranks of 43 and 52 reproduce (84.09 and 81.53): the loss comes from the larger budget, not from the allocation rule. Non-uniform allocation at a fixed budget therefore buys little, and we keep the uniform rank.

\section{How Much of a Task Delta Is Private}
\label{app:privacy}

Appendix~\ref{app:geometry} reports pairwise overlap between task subspaces. That is a summary of \emph{direction}, and it leaves open the quantity the merge actually acts on: what fraction of a task's update \emph{norm} lies in territory its neighbours claim. We measure it directly. For each task $i$ and each rank $R$, let $\mU^{(-i)}_\cup(R)$ be the orthonormalized union of the top-$R$ right-singular bases of the \emph{seven other} tasks, and report
\begin{equation}
\mathrm{capt}_i(R) \;=\; \norm{\Delta_i\,\mU^{(-i)}_\cup(R)} \;/\; \norm{\Delta_i},
\end{equation}
averaged over the 53 merged tensors in the norm sense (numerator and denominator accumulated separately, then divided once). This is the fraction of task $i$'s update that the joint projection would delete if the rank were $R$. For reference we also report $\mathrm{own}_i(R) = \norm{\Delta_i \mV_i(R)}/\norm{\Delta_i}$, the share its own top-$R$ basis carries. Table~\ref{tab:privacy} gives both.

\begin{table}[t]
\centering
\caption{\textbf{Private versus contested energy} (ViT-B/16, \%). Upper panel: share of task $i$'s delta norm inside the union of the other seven tasks' top-$R$ bases; this is what the projection at rank $R$ removes. Lower panel: share inside its own top-$R$ basis. Entries are norm ratios averaged over the 53 tensors, not squared energies.}
\label{tab:privacy}
\setlength{\tabcolsep}{1.6pt}
\footnotesize
\begin{tabular}{@{}lcccc|cccc@{}}
\toprule
 & \multicolumn{4}{c|}{Captured by neighbours} & \multicolumn{4}{c}{Own top-$R$} \\
Task & $R{=}8$ & $16$ & $32$ & $64$ & $R{=}8$ & $16$ & $32$ & $64$ \\
\midrule
MNIST    & 63.0 & 71.5 & 80.2 & 89.3 & 75.4 & 83.7 & 89.6 & 94.0 \\
EuroSAT  & 53.6 & 64.4 & 75.9 & 87.6 & 62.8 & 72.8 & 81.6 & 89.0 \\
GTSRB    & 53.8 & 63.4 & 74.1 & 86.0 & 66.1 & 76.7 & 85.5 & 91.8 \\
SVHN     & 61.7 & 70.5 & 79.5 & 89.1 & 70.9 & 80.4 & 87.4 & 92.7 \\
DTD      & 44.3 & 54.7 & 67.6 & 82.4 & 50.1 & 60.0 & 70.6 & 80.4 \\
RESISC45 & 49.3 & 59.9 & 72.2 & 85.2 & 52.9 & 62.7 & 72.9 & 82.3 \\
Cars     & 40.9 & 51.4 & 64.7 & 80.5 & 49.8 & 59.2 & 69.2 & 79.0 \\
SUN397   & 41.7 & 52.2 & 65.3 & 80.7 & 41.3 & 50.1 & 60.6 & 72.0 \\
\midrule
Mean     & 51.0 & 61.0 & 72.4 & 85.1 & 58.7 & 68.2 & 77.2 & 85.2 \\
\bottomrule
\end{tabular}
\end{table}

\paragraph{How much of each delta the peers reach.}
At $R{=}16$, the seven neighbours' bases capture 61.0\% of a task's delta norm on average, while the task's own top-16 captures 68.2\%. In energy terms the neighbours' share is about 37\% ($0.61^2$), so most of a task's energy lies outside its peers' union; the paper's premise is only that a substantial part does not. The 112-dimensional union and the 16-dimensional own basis are not like-for-like, so the closeness of the two numbers should not be read as a private share. As $R$ grows the two converge---at $R{=}64$ they are numerically equal, 85.1\% and 85.2\%---because both bases then describe most of the space.

\paragraph{Consistency with the measured retention.}
The number is not free of checks. The per-task norm retention of Table~\ref{tab:pertask_ret} is 77.8\% at $r{=}16$, which implies a removed-component norm fraction of $\sqrt{1-0.778^2} = 0.628$, against the 0.610 measured here. The two agree to within 0.02; the gap reflects averaging per-tensor ratios (Table~\ref{tab:pertask_ret}) versus pooling numerators and denominators (this table). Both quantities come from the same projection, so this is a consistency check, not an independent confirmation.

\paragraph{Which pairs are contested.}
The pairwise decomposition is informative about the failure structure of coordinatewise methods. The most entangled pair is MNIST--SVHN: 59.9\% of MNIST's norm lies in SVHN's top-16 and 55.4\% of SVHN's lies in MNIST's, so the two tasks claim the same input directions. Next are GTSRB--SVHN (46.1/49.2) and MNIST--GTSRB (48.5/43.7) (Table~\ref{tab:pairwise}). MNIST and SVHN are among the tasks where TIES-Merging and Task Arithmetic score \emph{highest} in Table~\ref{tab:main}, so pairwise sharing does not by itself predict where coordinatewise methods fail. The least invasive tasks are SUN397 and Cars (row means 27.2 and 27.3): their deltas land in other tasks' bases less than any others', the mirror image of MNIST and SVHN (38.4 and 37.7), which occupy the most. We note that this is a statement about the deltas, not about margins---\ours{}'s per-dataset gain over TIES is 13.16 on Cars but 8.27 on SUN397, and it is smallest (0.36, 1.81) on MNIST and SVHN, the two most invasive tasks.

\begin{table}[t]
\centering
\caption{\textbf{Pairwise captured fraction} at $r{=}16$ (ViT-B/16, \%): entry $(i,j)$ is the share of task $i$'s delta norm lying in task $j$'s top-16 right-singular basis. Rows sum to roughly the diagonal-free total of Table~\ref{tab:privacy}; the diagonal is left blank. Last two rows give the mean over the seven neighbours, per task.}
\label{tab:pairwise}
\setlength{\tabcolsep}{2.4pt}
\footnotesize
\begin{tabular}{@{}lcccccccc@{}}
\toprule
 & MNIST & EurS & GTS & SVHN & DTD & RES & Cars & SUN \\
\midrule
MNIST    & ---  & 33.1 & 48.5 & 59.9 & 32.0 & 31.9 & 32.2 & 30.7 \\
EuroSAT  & 31.4 & ---  & 35.6 & 34.6 & 34.9 & 45.1 & 30.4 & 34.6 \\
GTSRB    & 43.7 & 35.0 & ---  & 46.1 & 32.4 & 32.1 & 31.5 & 31.4 \\
SVHN     & 55.4 & 35.9 & 49.2 & ---  & 31.9 & 31.4 & 29.9 & 30.0 \\
DTD      & 26.9 & 30.9 & 28.1 & 27.2 & ---  & 33.1 & 29.5 & 33.4 \\
RESISC45 & 27.2 & 39.9 & 28.8 & 27.6 & 33.7 & ---  & 30.9 & 36.2 \\
Cars     & 25.7 & 25.7 & 26.7 & 24.9 & 28.2 & 28.6 & ---  & 31.3 \\
SUN397   & 23.8 & 27.5 & 25.4 & 23.8 & 29.7 & 31.2 & 29.3 & ---  \\
\midrule
Row mean & 38.4 & 35.2 & 36.1 & 37.7 & 29.9 & 32.0 & 27.3 & 27.2 \\
Col mean & 33.5 & 32.6 & 34.6 & 34.9 & 31.8 & 33.4 & 30.5 & 32.5 \\
\bottomrule
\end{tabular}
\end{table}

The matrix is not symmetric, and the asymmetry is informative. The row mean measures how much a task's delta lands in \emph{other} tasks' bases---how invasive it is---while the column mean measures how much of the other seven tasks lands in \emph{its} basis---how much territory it claims. MNIST and SVHN are the two most invasive tasks (row means 38.4 and 37.7); Cars and SUN397 are the two least (27.3 and 27.2). The column means are nearly flat, 30.5 to 34.9, so no task claims a disproportionate share of the shared volume; what differs is how much each task \emph{occupies} the others'. This distinction has a practical consequence that resolves an apparent puzzle in the main table. One might expect the most invasive tasks to be the ones most affected by the joint projection, since they are the ones whose removal hurts their neighbours. But the projection removes from each task only what lies in its peers' union, regardless of how invasive the task itself is, so invasiveness and vulnerability are different axes: MNIST is both invasive and strongly concentrated (83.7\% of its norm in its own top-16), and it survives the merge at 99.07\%, while SUN397 is neither invasive nor concentrated (50.1\%) and lands at 74.97\%. The tasks that score lowest after the merge (SUN397, Cars) are diffuse (own top-16 norm share 50.1\% and 59.2\%, Table~\ref{tab:privacy}) but the \emph{least} overlapped; across these tasks low accuracy tracks diffuseness, not overlap.

\paragraph{Own-basis share tracks accuracy across tasks.}
The captured fraction at $R{=}16$ correlates with final merged accuracy at $r{=}0.82$ across the eight tasks, and the own-top-16 share at $r{=}0.85$. Both are close to the correlation already reported for spectral concentration (Appendix~\ref{app:geometry}, $r{=}0.85$), and the difference between the two (own minus captured; not a private share, since the two bases differ in dimension) correlates \emph{positively} with accuracy at $r{=}0.73$.

\section{Rank Sweep, Per Rank}
\label{app:rankdetail}

\begin{figure}[t]
\centering
\begin{tikzpicture}[x=0.60cm,y=0.58cm,font=\scriptsize]
  \fill[oursblue!7] (1.85,0) rectangle (4.15,6.35);
  \draw[->,gray!70] (0.55,0) -- (5.55,0);
  \draw[->,gray!70] (0.55,0) -- (0.55,6.55);
  \node[below,black,inner sep=1pt] at (1.85,-1.12) {rank $r$};
  \node[basered,below,inner sep=1pt] at (4.05,-1.12) {seq.: $69.64$};
  \node[rotate=90,black,inner sep=1pt] at (-0.42,3.2) {avg.\ acc.\ (\%)};
  \foreach \v/\lab in {0/84, 2/86, 4/88, 6/90} {
    \draw[gray!40] (0.55,\v) -- (0.45,\v);
    \node[left,gray!70!black] at (0.42,\v) {\lab};
    \draw[gray!25,densely dotted] (0.55,\v) -- (5.45,\v);
  }
  \foreach \x/\lab in {1/8, 2/12, 3/16, 4/20, 5/32} {
    \draw[gray!40] (\x,0) -- (\x,-0.10);
    \node[below,gray!70!black] at (\x,-0.10) {\lab};
  }
  \draw[baseorange,thick,dashed,mark=square*,mark size=1.3pt]
    (1,1.00) -- (2,2.15) -- (3,2.51) -- (4,2.52) -- (5,0.94);
  \draw[oursblue,very thick,mark=*,mark size=1.6pt]
    (1,4.03) -- (2,5.30) -- (3,5.45) -- (4,5.22) -- (5,3.08);
  \node[above,oursblue] at (3,5.52) {89.45};
  \node[below right,baseorange] at (5.02,0.90) {84.94};
  \draw[basered,->,line width=0.9pt] (1.0,0.88) -- (1.0,0.22);
  \node[oursblue,inner sep=1.5pt] at (3.55,4.72) {\ours{}};
  \node[baseorange,inner sep=1.5pt] at (3.35,2.02) {shared};
  \node at (3.0,-2.02) {(a) accuracy versus rank};
\end{tikzpicture}\hspace{1pt}%
\begin{tikzpicture}[x=0.46cm,y=0.58cm,font=\scriptsize]
  \draw[->,gray!70] (0,0) -- (9.7,0);
  \draw[->,gray!70] (0,0) -- (0,6.55);
  \node[below,black,inner sep=1pt] at (4.8,-1.12) {retained delta norm (\%)};
  \foreach \v/\lab in {0/84, 2/86, 4/88, 6/90} {
    \draw[gray!40] (0,\v) -- (-0.12,\v);
    \node[left,gray!70!black] at (-0.15,\v) {\lab};
    \draw[gray!25,densely dotted] (0,\v) -- (9.4,\v);
  }
  \foreach \x/\lab in {0/25, 2/27, 4/29, 6/31, 8/33} {
    \draw[gray!40] (\x,0) -- (\x,-0.10);
    \node[below,gray!70!black] at (\x,-0.10) {\lab};
  }
  \draw[baseorange,thick,dashed,mark=square*,mark size=1.3pt]
    (9.04,2.15) -- (7.83,2.51) -- (7.01,2.52);
  \draw[oursblue,very thick,mark=*,mark size=1.6pt]
    (8.44,4.03) -- (6.73,5.30) -- (5.36,5.45) -- (4.16,5.22) -- (1.00,3.08);
  \node[anchor=north west,oursblue,inner sep=1.5pt] at (8.46,3.94) {$r{=}8$};
  \node[above right,oursblue,inner sep=1.5pt] at (5.36,5.45) {$r{=}16$};
  \node[below left,oursblue,inner sep=1.5pt] at (1.00,3.02) {$r{=}32$};
  \node[below left,baseorange,inner sep=1.2pt] at (9.04,2.15) {shared};
  \node at (4.8,-2.02) {(b) accuracy versus retained norm};
\end{tikzpicture}
\caption{\textbf{Rank and mechanism.} (a) Joint projection has a broad optimum around $r{=}16$ (shaded), leading the shared-basis control by 2.1--3.2 points at every rank; the sequential variant at matched rank collapses to 69.64. (b) The same runs on the same accuracy axis, placed against the fraction of aggregate delta norm they retain; the path runs $r{=}32 \to 20 \to 16 \to 12 \to 8$ left to right: accuracy peaks at intermediate retention, and the shared basis retains \emph{more} norm at every rank while scoring lower, so the score is not a function of how much is deleted.}
\label{fig:rank}
\end{figure}

\section{Order Dependence, Per Dataset}
\label{app:order}

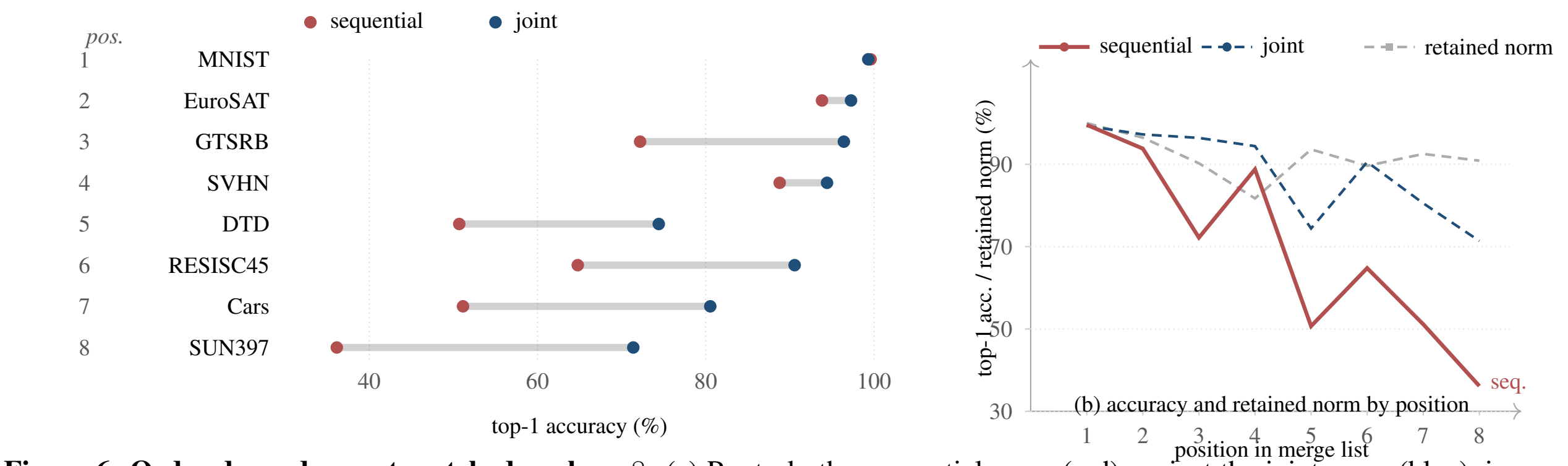
\begin{figure*}[t]
\centering
\begin{tikzpicture}[x=0.90cm,y=0.44cm,font=\scriptsize]
  \useasboundingbox (-2.55,-1.15) rectangle (8.15,9.05);
  \foreach \v/\lab in {1/40, 3/60, 5/80, 7/100} {
    \draw[gray!25,densely dotted] (\v,0.85) -- (\v,8.15);
    \node[below,gray!70!black] at (\v,0.82) {\lab};
  }
  \node[below,black,inner sep=6pt] at (3.5,-0.05) {top-1 accuracy (\%)};
  \foreach \i/\name/\seq/\joi/\p in {%
    0/MNIST/6.959/6.931/1, 1/EuroSAT/6.381/6.726/2, 2/GTSRB/4.219/6.641/3,
    3/SVHN/5.878/6.441/4, 4/DTD/2.069/4.441/5, 5/RESISC45/3.478/6.056/6,
    6/Cars/2.114/5.053/7, 7/SUN397/0.614/4.138/8} {
    \pgfmathsetmacro{\y}{8.15-\i}
    \draw[basegrey,line width=2.2pt,opacity=0.40] (\seq,\y) -- (\joi,\y);
    \fill[basered] (\seq,\y) circle (1.9pt);
    \fill[oursblue] (\joi,\y) circle (1.9pt);
    \node[left,inner sep=2pt] at (-0.10,\y) {\name};
    \node[anchor=west,inner sep=1pt,font=\scriptsize\color{gray!70!black}] at (-2.50,\y) {\p};
  }
  \node[anchor=west,font=\scriptsize\itshape,text=gray!70!black] at (-2.50,8.60) {pos.};
  \fill[basered] (0.3,9.05) circle (1.9pt);
  \node[right,inner sep=2pt] at (0.45,9.05) {sequential};
  \fill[oursblue] (2.5,9.05) circle (1.9pt);
  \node[right,inner sep=2pt] at (2.65,9.05) {joint};
  \node[below,font=\scriptsize,inner sep=8pt] at (3.5,-1.15) {(a) per-task, $r{=}8$};
\end{tikzpicture}\hspace{2pt}%
\begin{tikzpicture}[x=0.60cm,y=0.44cm,font=\scriptsize]
  \useasboundingbox (-0.95,-0.75) rectangle (9.6,9.05);
  \draw[->,gray!70] (0,0) -- (8.75,0);
  \draw[->,gray!70] (0,0) -- (0,8.55);
  \node[below,black,inner sep=2pt] at (4.3,-0.55) {position in merge list};
  \node[rotate=90,black,inner sep=2pt] at (-0.80,4.2) {top-1 acc.\ / retained norm (\%)};
  \foreach \v/\lab in {0/30, 2/50, 4/70, 6/90} {
    \draw[gray!40] (0,\v) -- (-0.10,\v);
    \node[left,gray!70!black] at (-0.12,\v) {\lab};
    \draw[gray!22,densely dotted] (0,\v) -- (8.55,\v);
  }
  \foreach \x in {1,...,8} {
    \draw[gray!40] (\x,0) -- (\x,-0.10);
    \node[below,gray!70!black] at (\x,-0.11) {\x};
  }
  \draw[gray!65,thick,densely dashed,mark=square*,mark size=1.2pt]
    (1,7.000) -- (2,6.647) -- (3,6.025) -- (4,5.172)
    -- (5,6.361) -- (6,5.964) -- (7,6.252) -- (8,6.088);
  \draw[oursblue,thick,densely dashed,mark=*,mark size=1.4pt]
    (1,6.931) -- (2,6.726) -- (3,6.641) -- (4,6.441)
    -- (5,4.441) -- (6,6.056) -- (7,5.053) -- (8,4.138);
  \draw[basered,very thick,mark=*,mark size=1.4pt]
    (1,6.959) -- (2,6.381) -- (3,4.219) -- (4,5.878)
    -- (5,2.069) -- (6,3.478) -- (7,2.114) -- (8,0.614);
  \draw[basered,very thick] (0.15,8.85) -- (1.05,8.85);
  \fill[basered] (0.60,8.85) circle (1.4pt);
  \node[right,inner sep=2pt] at (1.10,8.85) {sequential};
  \draw[oursblue,thick,densely dashed] (3.05,8.85) -- (3.95,8.85);
  \fill[oursblue] (3.50,8.85) circle (1.4pt);
  \node[right,inner sep=2pt] at (4.00,8.85) {joint};
  \draw[gray!65,thick,densely dashed] (5.95,8.85) -- (6.85,8.85);
  \fill[gray!65] (6.40,8.85) ++(-1.1pt,-1.1pt) rectangle ++(2.2pt,2.2pt);
  \node[right,inner sep=2pt] at (6.90,8.85) {retained norm};
  \node[right,basered,inner sep=1.5pt] at (8.10,0.62) {seq.};
  \node[above,font=\scriptsize,inner sep=8pt] at (4.3,-0.78) {(b) accuracy and retained norm by position};
\end{tikzpicture}
\caption{\textbf{Order dependence at matched rank $r{=}8$.} (a) Per task, the sequential score (red) against the joint score (blue), in merge-list order. Early tasks survive sequential merging nearly intact; later ones are progressively gutted, while the joint score does not depend on list position. (b) The same sequential accuracies as a function of list position (solid red), with the joint scores as a reference (blue dashed) and each task's \emph{retained} delta norm (grey, $\lVert\tilde\btau_i\rVert_F/\lVert\btau_i\rVert_F$, computed as one pooled ratio over all 53 tensors). Sequential retention stays between 81.7\% and 100\% at every position and shows no downward trend across the list (the last task retains 90.9\%, above the 81.7\% minimum), so the collapse cannot be read as ``later tasks lose their norm''. Joint projection retains \emph{less} on average (77.7--91.2\%) and scores 18.4 points higher, so the amount removed does not explain the collapse.}
\label{fig:order}
\end{figure*}

Section~\ref{sec:abl_seq} reports that the sequential variant collapses to 69.64\% at $r{=}8$ against 88.03\% for the joint merge. Figure~\ref{fig:order} and Table~\ref{tab:order} (below) decompose that collapse by task. The degradation grows with the position of the task in the merge list. The first two tasks are barely touched (99.59 and 93.81), the fourth still holds 88.78, and the losses then climb steadily through the tail of the list: 23.7, 25.8, 29.4, and 35.2 points for DTD, RESISC45, Cars, and SUN397. The one exception to the trend is GTSRB, which sits third in the list and drops to 72.19, 24.2 points below the joint merge; its retained norm (90.3\%) does not explain the drop, and we do not attribute it to a specific overlap. The pattern is consistent with Proposition~\ref{prop:sequential}: each task is projected against the union of every task that came before it, so the removed subspace grows monotonically, up to 56 directions for the last task at $r{=}8$; at the same time, early tasks are never filtered against later ones (Appendix~\ref{app:proof_sequential}).

\begin{table}[t]
\centering
\caption{\textbf{Sequential versus joint merging at $r{=}8$} (ViT-B/16, \%). The sequential column merges tasks in the order listed, projecting each against the basis accumulated so far; the joint column projects every task against the same seven neighbours.}
\label{tab:order}
\setlength{\tabcolsep}{4pt}
\small
\begin{tabular}{@{}lcccc@{}}
\toprule
Task & Position & Sequential & Joint & $\Delta$ \\
\midrule
MNIST    & 1 & 99.59 & 99.31 & $+0.28$ \\
EuroSAT  & 2 & 93.81 & 97.26 & $-3.45$ \\
GTSRB    & 3 & 72.19 & 96.41 & $-24.22$ \\
SVHN     & 4 & 88.78 & 94.41 & $-5.63$ \\
DTD      & 5 & 50.69 & 74.41 & $-23.72$ \\
RESISC45 & 6 & 64.78 & 90.56 & $-25.78$ \\
Cars     & 7 & 51.14 & 80.53 & $-29.39$ \\
SUN397   & 8 & 36.14 & 71.38 & $-35.24$ \\
\midrule
Average  & & 69.64 & 88.03 & $-18.39$ \\
Retained & & 35.90 & 33.44 & \\
\bottomrule
\end{tabular}
\end{table}

The retained-norm row is the interesting one. The sequential merge keeps \emph{more} of the delta norm (35.90\% versus 33.44\%) and performs far worse. So the amount deleted does not explain the collapse. Each task loses only its overlap with \emph{earlier} tasks; early tasks are never filtered against later ones, so their overlap with later tasks enters the sum unprojected. The joint construction filters every task against the same set of neighbours, which makes the per-task profile in Figure~\ref{fig:order} independent of list position. This comparison suggests that which components are removed matters more than how much, but does not isolate it.

\section{What the Projection Removes}
\label{app:retention}

Table~\ref{tab:retention} records the aggregate retained delta ratio across the rank sweep for both variants. The aggregate ratio is $\norm{\sum_i\tilde{\btau}_i}\,/\,\sum_i\norm{\btau_i}$, computed per tensor and averaged over the 53 tensors; it is the quantity reported in the main text. It is \emph{not} the per-task ratio $\norm{\tilde{\btau}_i}/\norm{\btau_i}$, which is much higher (77.8\% at $r{=}16$ versus 30.4\%). The aggregate ratio is low mainly because the eight deltas are nearly orthogonal, so their sum is already well below the sum of their norms before any projection (about $0.45$; Appendix~\ref{app:worked}); projection then lowers it further to $0.30$. Table~\ref{tab:pertask_ret} separates the two and Appendix~\ref{app:geometry} measures the cross-task overlap. Two regularities across the rank sweep are discussed after the module breakdown below.

\begin{table}[t]
\centering
\caption{\textbf{Aggregate retained fraction of the delta norm}, $\norm{\sum_i\tilde{\btau}_i}/\sum_i\norm{\btau_i}$, at each rank (ViT-B/16, \%). The joint merge retains slightly less than the shared control at every rank and scores higher at every rank. Per-task retention is in Table~\ref{tab:pertask_ret}.}
\label{tab:retention}
\setlength{\tabcolsep}{4pt}
\small
\begin{tabular}{@{}lccccc@{}}
\toprule
Variant & $r{=}8$ & $r{=}12$ & $r{=}16$ & $r{=}20$ & $r{=}32$ \\
\midrule
Joint             & 33.44 & 31.73 & 30.36 & 29.16 & 25.98 \\
Shared basis      & 35.47 & 34.04 & 32.83 & 32.01 & 29.98 \\
\bottomrule
\end{tabular}
\end{table}

\begin{table}[t]
\centering
\caption{\textbf{Per-task retained fraction of the delta norm}, $\norm{\tilde{\btau}_i}/\norm{\btau_i}$, under joint projection (ViT-B/16, \%). Each cell is the mean over the 53 merged tensors of the per-tensor ratio $\norm{\tilde{\btau}_i^{(k)}}/\norm{\btau_i^{(k)}}$. This averages per-tensor ratios and is therefore not the pooled ratio used for the grey curve of Figure~\ref{fig:order}, nor the aggregate ratio of Table~\ref{tab:retention}.}
\label{tab:pertask_ret}
\setlength{\tabcolsep}{4pt}
\small
\begin{tabular}{@{}lcccc@{}}
\toprule
Task & $r{=}8$ & $r{=}12$ & $r{=}16$ & $r{=}20$ \\
\midrule
MNIST    & 79.8 & 75.3 & 71.8 & 68.8 \\
EuroSAT  & 84.2 & 80.0 & 76.3 & 73.1 \\
GTSRB    & 83.5 & 79.5 & 76.0 & 72.9 \\
SVHN     & 79.6 & 75.1 & 71.6 & 68.5 \\
DTD      & 87.8 & 84.4 & 81.3 & 78.5 \\
RESISC45 & 86.0 & 82.1 & 78.7 & 75.6 \\
Cars     & 89.6 & 86.5 & 83.7 & 81.0 \\
SUN397   & 89.1 & 85.8 & 82.8 & 80.1 \\
\midrule
Mean     & 84.9 & 81.1 & 77.8 & 74.8 \\
\bottomrule
\end{tabular}
\end{table}

\paragraph{Retention is not a difficulty score.}
Per-task retention spans 71.6 (SVHN) to 83.7 (Cars) at $r{=}16$, and the ordering is close to the opposite of what an ``easy tasks are less disturbed'' intuition would predict: the two tasks with the lowest retention, MNIST and SVHN, are among the \emph{highest}-scoring, while SUN397 is among the highest-retention tasks and scores lowest. The correlation between retention and accuracy across the eight tasks is $-0.78$. The variable that explains the ordering is the spectral concentration of the update, $\phi$: retention and $\phi$ correlate at $-0.965$ at $r{=}16$, and the relation is stable across the rank sweep ($-0.963$, $-0.964$, $-0.965$, $-0.965$ at $r{=}8,12,16,20$). This is consistent with concentrated updates pointing in similar directions, so that more of a concentrated delta falls inside the other tasks' union, while a diffuse delta puts more of its energy in directions nobody else claims. The same $\phi$ correlates with accuracy positively ($+0.85$, Appendix~\ref{app:geometry}); retention and accuracy are both strongly correlated with $\phi$ across these eight tasks.

\paragraph{Where the removal lands.}
Aggregated by module type at $r{=}16$, retention is highest in the MLP output projection (84.0\%) and the attention output projection (80.2\%), and lowest in the embedding and projection tensors (62.1\%)---a 21.9-point spread. The embedding group is the lowest at every rank in the table below, but the attention output projection, not the MLP output, is highest at $r{=}8$, so the ordering among the other modules is not stable across ranks. We do not test a mechanism for the embedding result. The per-block profile is flat (Figure~\ref{fig:layer}a), and we did not test a per-module rank.

\begin{table}[t]
\centering
\caption{\textbf{Per-task retention by module type} under joint projection (ViT-B/16, \%). Each cell averages the retained per-task ratio $\norm{\tilde\btau_i}/\norm{\btau_i}$ over the eight tasks and over every tensor of that module type. The final row is the spread between the best- and worst-preserved module at each rank.}
\label{tab:module_ret}
\setlength{\tabcolsep}{4.0pt}
\footnotesize
\begin{tabular}{@{}lcccc@{}}
\toprule
Module type & $r{=}8$ & $r{=}12$ & $r{=}16$ & $r{=}20$ \\
\midrule
MLP output projection (\texttt{c\_proj})        & 87.7 & 85.7 & 84.0 & 82.5 \\
Attention output projection (\texttt{out\_proj}) & 88.1 & 84.0 & 80.2 & 76.8 \\
Attention input projections (\texttt{in\_proj})  & 84.6 & 80.3 & 76.5 & 73.1 \\
MLP input projection (\texttt{c\_fc})            & 85.0 & 80.7 & 76.9 & 73.5 \\
Embeddings and visual/text projections            & 71.4 & 65.9 & 62.1 & 58.9 \\
\midrule
Spread (max $-$ min) & 16.7 & 19.8 & 21.9 & 23.6 \\
\bottomrule
\end{tabular}
\end{table}

Table~\ref{tab:module_ret} gives the full profile. Every module type loses ground as $r$ grows, but the rate of loss is unequal: the embedding and projection group falls from 71.4 to 58.9, while the MLP output projection falls only from 87.7 to 82.5, so the spread grows by about 40\%, from 16.7 to 23.6 points, between $r{=}8$ and $r{=}20$. The attention input, attention output and MLP input projections each fall by 11.3--11.5 points over the same range. The embedding and projection group is lowest at every rank; whether a per-module rank would help in accuracy is untested; the depth profile below suggests the room for improvement is limited, because interference is spread across depth rather than concentrated.

First, the joint merge always removes more than the shared control---by 2.03, 2.31, 2.47, 2.85 and 4.00 points at $r{=}8$, $12$, $16$, $20$ and $32$---and yet always scores higher. If raw excision were the mechanism, the more aggressive method would be the more damaging one; it is not, which is consistent with the choice of \emph{which} directions to remove mattering more than \emph{how many}. Second, the score is not monotone in the retained fraction: retention falls from 33.44 to 30.36 between $r{=}8$ and $r{=}16$ while accuracy rises 1.42 points, and falls a further 1.2 points to 29.16 at $r{=}20$ while accuracy falls 0.23. A plausible reading, not tested directly, is that the rank trades off covering the interfering directions (which needs large $r$) against deleting a task's own content (which grows with $r$), with the optimum where the two balance.

\section{The Geometry of the Task Subspaces}
\label{app:geometry}

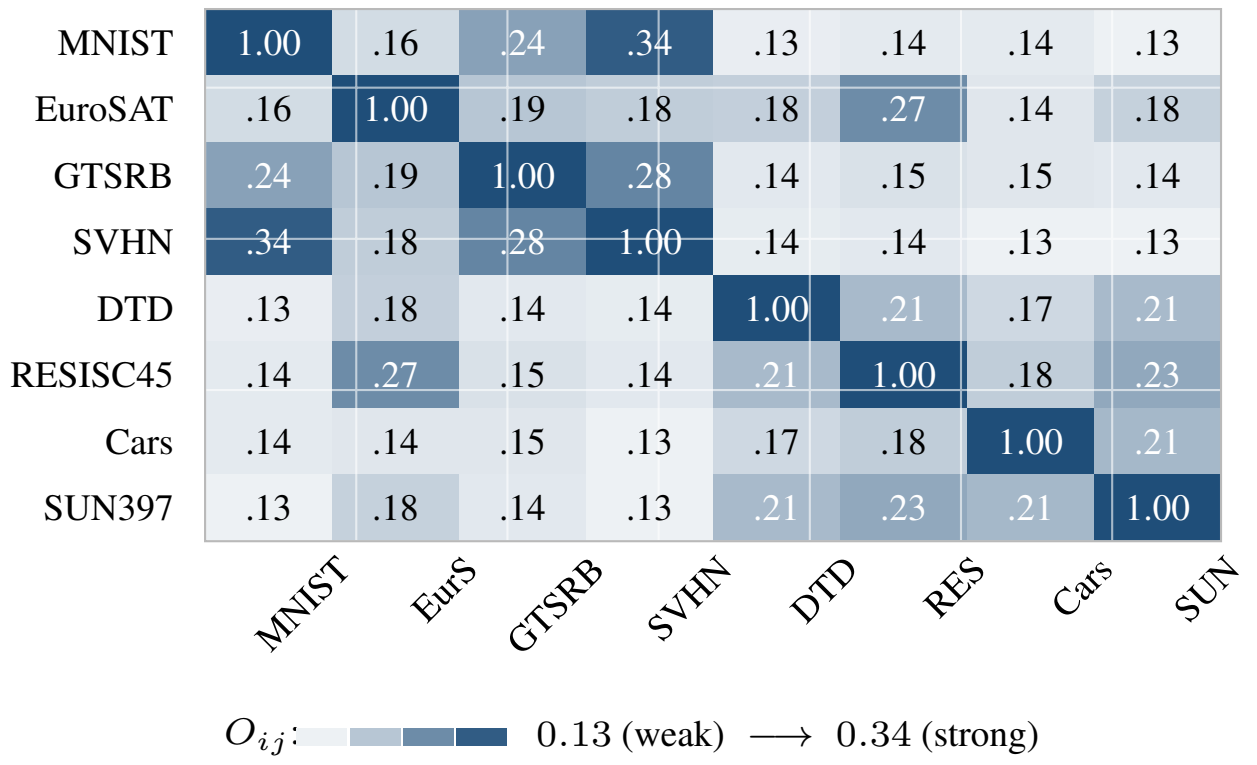
\begin{figure}[t]
\centering
\begin{tikzpicture}[x=0.84cm,y=0.44cm]
  \fill[oursblue!100] (0,7) rectangle ++(1,1);
  \fill[oursblue!20] (1,7) rectangle ++(1,1);
  \fill[oursblue!53] (2,7) rectangle ++(1,1);
  \fill[oursblue!92] (3,7) rectangle ++(1,1);
  \fill[oursblue!10] (4,7) rectangle ++(1,1);
  \fill[oursblue!12] (5,7) rectangle ++(1,1);
  \fill[oursblue!12] (6,7) rectangle ++(1,1);
  \fill[oursblue!8] (7,7) rectangle ++(1,1);
  \fill[oursblue!20] (0,6) rectangle ++(1,1);
  \fill[oursblue!100] (1,6) rectangle ++(1,1);
  \fill[oursblue!32] (2,6) rectangle ++(1,1);
  \fill[oursblue!28] (3,6) rectangle ++(1,1);
  \fill[oursblue!27] (4,6) rectangle ++(1,1);
  \fill[oursblue!65] (5,6) rectangle ++(1,1);
  \fill[oursblue!14] (6,6) rectangle ++(1,1);
  \fill[oursblue!26] (7,6) rectangle ++(1,1);
  \fill[oursblue!53] (0,5) rectangle ++(1,1);
  \fill[oursblue!32] (1,5) rectangle ++(1,1);
  \fill[oursblue!100] (2,5) rectangle ++(1,1);
  \fill[oursblue!69] (3,5) rectangle ++(1,1);
  \fill[oursblue!13] (4,5) rectangle ++(1,1);
  \fill[oursblue!17] (5,5) rectangle ++(1,1);
  \fill[oursblue!14] (6,5) rectangle ++(1,1);
  \fill[oursblue!13] (7,5) rectangle ++(1,1);
  \fill[oursblue!92] (0,4) rectangle ++(1,1);
  \fill[oursblue!28] (1,4) rectangle ++(1,1);
  \fill[oursblue!69] (2,4) rectangle ++(1,1);
  \fill[oursblue!100] (3,4) rectangle ++(1,1);
  \fill[oursblue!11] (4,4) rectangle ++(1,1);
  \fill[oursblue!13] (5,4) rectangle ++(1,1);
  \fill[oursblue!8] (6,4) rectangle ++(1,1);
  \fill[oursblue!8] (7,4) rectangle ++(1,1);
  \fill[oursblue!10] (0,3) rectangle ++(1,1);
  \fill[oursblue!27] (1,3) rectangle ++(1,1);
  \fill[oursblue!13] (2,3) rectangle ++(1,1);
  \fill[oursblue!11] (3,3) rectangle ++(1,1);
  \fill[oursblue!100] (4,3) rectangle ++(1,1);
  \fill[oursblue!39] (5,3) rectangle ++(1,1);
  \fill[oursblue!22] (6,3) rectangle ++(1,1);
  \fill[oursblue!39] (7,3) rectangle ++(1,1);
  \fill[oursblue!12] (0,2) rectangle ++(1,1);
  \fill[oursblue!65] (1,2) rectangle ++(1,1);
  \fill[oursblue!17] (2,2) rectangle ++(1,1);
  \fill[oursblue!13] (3,2) rectangle ++(1,1);
  \fill[oursblue!39] (4,2) rectangle ++(1,1);
  \fill[oursblue!100] (5,2) rectangle ++(1,1);
  \fill[oursblue!28] (6,2) rectangle ++(1,1);
  \fill[oursblue!49] (7,2) rectangle ++(1,1);
  \fill[oursblue!12] (0,1) rectangle ++(1,1);
  \fill[oursblue!14] (1,1) rectangle ++(1,1);
  \fill[oursblue!14] (2,1) rectangle ++(1,1);
  \fill[oursblue!8] (3,1) rectangle ++(1,1);
  \fill[oursblue!22] (4,1) rectangle ++(1,1);
  \fill[oursblue!28] (5,1) rectangle ++(1,1);
  \fill[oursblue!100] (6,1) rectangle ++(1,1);
  \fill[oursblue!40] (7,1) rectangle ++(1,1);
  \fill[oursblue!8] (0,0) rectangle ++(1,1);
  \fill[oursblue!26] (1,0) rectangle ++(1,1);
  \fill[oursblue!13] (2,0) rectangle ++(1,1);
  \fill[oursblue!8] (3,0) rectangle ++(1,1);
  \fill[oursblue!39] (4,0) rectangle ++(1,1);
  \fill[oursblue!49] (5,0) rectangle ++(1,1);
  \fill[oursblue!40] (6,0) rectangle ++(1,1);
  \fill[oursblue!100] (7,0) rectangle ++(1,1);
  \draw[white,line width=0.4pt,opacity=0.6] (0,0) grid (8,8);
  \draw[gray!55,line width=0.4pt] (0,0) rectangle (8,8);
  \node[font=\scriptsize,inner sep=0pt,text=white] at (0.5,7.5) {1.00};
  \node[font=\scriptsize,inner sep=0pt,text=black] at (1.5,7.5) {.16};
  \node[font=\scriptsize,inner sep=0pt,text=white] at (2.5,7.5) {.24};
  \node[font=\scriptsize,inner sep=0pt,text=white] at (3.5,7.5) {.34};
  \node[font=\scriptsize,inner sep=0pt,text=black] at (4.5,7.5) {.13};
  \node[font=\scriptsize,inner sep=0pt,text=black] at (5.5,7.5) {.14};
  \node[font=\scriptsize,inner sep=0pt,text=black] at (6.5,7.5) {.14};
  \node[font=\scriptsize,inner sep=0pt,text=black] at (7.5,7.5) {.13};
  \node[font=\scriptsize,inner sep=0pt,text=black] at (0.5,6.5) {.16};
  \node[font=\scriptsize,inner sep=0pt,text=white] at (1.5,6.5) {1.00};
  \node[font=\scriptsize,inner sep=0pt,text=black] at (2.5,6.5) {.19};
  \node[font=\scriptsize,inner sep=0pt,text=black] at (3.5,6.5) {.18};
  \node[font=\scriptsize,inner sep=0pt,text=black] at (4.5,6.5) {.18};
  \node[font=\scriptsize,inner sep=0pt,text=white] at (5.5,6.5) {.27};
  \node[font=\scriptsize,inner sep=0pt,text=black] at (6.5,6.5) {.14};
  \node[font=\scriptsize,inner sep=0pt,text=black] at (7.5,6.5) {.18};
  \node[font=\scriptsize,inner sep=0pt,text=white] at (0.5,5.5) {.24};
  \node[font=\scriptsize,inner sep=0pt,text=black] at (1.5,5.5) {.19};
  \node[font=\scriptsize,inner sep=0pt,text=white] at (2.5,5.5) {1.00};
  \node[font=\scriptsize,inner sep=0pt,text=white] at (3.5,5.5) {.28};
  \node[font=\scriptsize,inner sep=0pt,text=black] at (4.5,5.5) {.14};
  \node[font=\scriptsize,inner sep=0pt,text=black] at (5.5,5.5) {.15};
  \node[font=\scriptsize,inner sep=0pt,text=black] at (6.5,5.5) {.15};
  \node[font=\scriptsize,inner sep=0pt,text=black] at (7.5,5.5) {.14};
  \node[font=\scriptsize,inner sep=0pt,text=white] at (0.5,4.5) {.34};
  \node[font=\scriptsize,inner sep=0pt,text=black] at (1.5,4.5) {.18};
  \node[font=\scriptsize,inner sep=0pt,text=white] at (2.5,4.5) {.28};
  \node[font=\scriptsize,inner sep=0pt,text=white] at (3.5,4.5) {1.00};
  \node[font=\scriptsize,inner sep=0pt,text=black] at (4.5,4.5) {.14};
  \node[font=\scriptsize,inner sep=0pt,text=black] at (5.5,4.5) {.14};
  \node[font=\scriptsize,inner sep=0pt,text=black] at (6.5,4.5) {.13};
  \node[font=\scriptsize,inner sep=0pt,text=black] at (7.5,4.5) {.13};
  \node[font=\scriptsize,inner sep=0pt,text=black] at (0.5,3.5) {.13};
  \node[font=\scriptsize,inner sep=0pt,text=black] at (1.5,3.5) {.18};
  \node[font=\scriptsize,inner sep=0pt,text=black] at (2.5,3.5) {.14};
  \node[font=\scriptsize,inner sep=0pt,text=black] at (3.5,3.5) {.14};
  \node[font=\scriptsize,inner sep=0pt,text=white] at (4.5,3.5) {1.00};
  \node[font=\scriptsize,inner sep=0pt,text=white] at (5.5,3.5) {.21};
  \node[font=\scriptsize,inner sep=0pt,text=black] at (6.5,3.5) {.17};
  \node[font=\scriptsize,inner sep=0pt,text=white] at (7.5,3.5) {.21};
  \node[font=\scriptsize,inner sep=0pt,text=black] at (0.5,2.5) {.14};
  \node[font=\scriptsize,inner sep=0pt,text=white] at (1.5,2.5) {.27};
  \node[font=\scriptsize,inner sep=0pt,text=black] at (2.5,2.5) {.15};
  \node[font=\scriptsize,inner sep=0pt,text=black] at (3.5,2.5) {.14};
  \node[font=\scriptsize,inner sep=0pt,text=white] at (4.5,2.5) {.21};
  \node[font=\scriptsize,inner sep=0pt,text=white] at (5.5,2.5) {1.00};
  \node[font=\scriptsize,inner sep=0pt,text=black] at (6.5,2.5) {.18};
  \node[font=\scriptsize,inner sep=0pt,text=white] at (7.5,2.5) {.23};
  \node[font=\scriptsize,inner sep=0pt,text=black] at (0.5,1.5) {.14};
  \node[font=\scriptsize,inner sep=0pt,text=black] at (1.5,1.5) {.14};
  \node[font=\scriptsize,inner sep=0pt,text=black] at (2.5,1.5) {.15};
  \node[font=\scriptsize,inner sep=0pt,text=black] at (3.5,1.5) {.13};
  \node[font=\scriptsize,inner sep=0pt,text=black] at (4.5,1.5) {.17};
  \node[font=\scriptsize,inner sep=0pt,text=black] at (5.5,1.5) {.18};
  \node[font=\scriptsize,inner sep=0pt,text=white] at (6.5,1.5) {1.00};
  \node[font=\scriptsize,inner sep=0pt,text=white] at (7.5,1.5) {.21};
  \node[font=\scriptsize,inner sep=0pt,text=black] at (0.5,0.5) {.13};
  \node[font=\scriptsize,inner sep=0pt,text=black] at (1.5,0.5) {.18};
  \node[font=\scriptsize,inner sep=0pt,text=black] at (2.5,0.5) {.14};
  \node[font=\scriptsize,inner sep=0pt,text=black] at (3.5,0.5) {.13};
  \node[font=\scriptsize,inner sep=0pt,text=white] at (4.5,0.5) {.21};
  \node[font=\scriptsize,inner sep=0pt,text=white] at (5.5,0.5) {.23};
  \node[font=\scriptsize,inner sep=0pt,text=white] at (6.5,0.5) {.21};
  \node[font=\scriptsize,inner sep=0pt,text=white] at (7.5,0.5) {1.00};
  \foreach \c/\nm in {0/MNIST,1/EurS,2/GTSRB,3/SVHN,4/DTD,5/RES,6/Cars,7/SUN} {
    \node[rotate=45,anchor=north east,font=\scriptsize,inner sep=1pt] at (\c+1.02,-0.05) {\nm};}
  \foreach \r/\nm in {0/MNIST,1/EuroSAT,2/GTSRB,3/SVHN,4/DTD,5/RESISC45,6/Cars,7/SUN397} {
    \node[anchor=east,font=\scriptsize] at (-0.10,7.5-\r) {\nm};}
  \begin{scope}[shift={(0,-3.05)}]
    \node[anchor=west,font=\scriptsize] at (0,0.11) {$O_{ij}$:};
    \foreach \s/\x in {8/0.72,32/1.14,65/1.56,92/1.98} {
      \fill[oursblue!\s] (\x,-0.02) rectangle ++(0.40,0.26);}
    \node[anchor=west,font=\scriptsize] at (2.46,0.10) {$0.13$ (weak) $\;\longrightarrow\;$ $0.34$ (strong)};
  \end{scope}
\end{tikzpicture}
\caption{\textbf{Cross-task subspace overlap} on ViT-B/16, Eq.~\eqref{eq:overlap} averaged over the 53 merged tensors ($r{=}16$). The diagonal is $1$ by construction. Independent random subspaces at these widths would sit at $0.018$ on average; every off-diagonal cell is at least seven times that, and the mean is $0.179$, or $9.9$ times chance. The darkest pairs share an input modality or a visual statistic.}
\label{fig:overlap}
\end{figure}

The merge is a function of one measured object: how much two tasks' subspaces overlap. For tasks $i,j$ and a single tensor with orthonormal bases $\mV_i,\mV_j \in \R^{n \times r}$, define
\begin{equation}
  O_{ij} \;=\; \frac{1}{r}\bigl\lVert \mV_i^{\top}\mV_j \bigr\rVert_F^2 ,
  \label{eq:overlap}
\end{equation}
the mean squared cosine between the two subspaces. It equals $1$ when they coincide, $0$ when they are orthogonal, and $r/n$ in expectation for two subspaces drawn independently in $\R^n$ ($16/768 = 0.021$ in a $768$-wide tensor, and smaller where the tensor is wider). Averaged over the 53 merged tensors of ViT-B/16, whose widths range from 512 to 3072, that expectation is $0.018$. Figure~\ref{fig:overlap} reports $O_{ij}$ averaged over the same 53 tensors.

Three facts stand out. First, no pair is anywhere near orthogonal: the smallest entry, $O_{\text{MNIST},\text{SUN397}}{=}0.130$, is still $7.2$ times the matched random baseline of $0.018$. Fine-tuning the same CLIP encoder on eight unrelated datasets does not produce independent row spaces: every task's dominant directions overlap with every other task's well beyond chance. This is the quantitative sense in which the projection touches every task.

Second, the spread across pairs is modest, a factor of $2.6$ from weakest to strongest, so interference is broadly distributed rather than concentrated in a few pathological pairings. The three strongest pairs are MNIST--SVHN ($0.338$), GTSRB--SVHN ($0.278$) and EuroSAT--RESISC45 ($0.272$): digits with house numbers, two sign-and-number datasets, and two aerial-imagery datasets. The weakest are MNIST--SUN397 and SVHN--Cars (both $0.130$).

Third, the union bound is not the binding constraint. Seven neighbours contribute at most $7r{=}112$ directions out of $768$, so the projection can remove at most $15\%$ of the input dimensions; yet each task loses $22\%$ of its delta norm on average (per-task retention $77.8\%$, Table~\ref{tab:pertask_ret}), more than the $8\%$ an isotropic delta would lose, so the removed directions are ones the deltas concentrate on. (The aggregate merged norm, $30.4\%$ of the summed norms, is a different quantity; Appendix~\ref{app:retention}.)

\paragraph{Overlap is not just shared magnitude.}
A natural objection is that $O_{ij}$ measures what it measures because all eight deltas look alike---if every fine-tuned model moved its weights in a similar overall direction, any two rank-$r$ subspaces drawn from them would overlap. The right control is the entrywise cosine between the flattened deltas, $\langle \mathrm{vec}\,\Delta_i, \mathrm{vec}\,\Delta_j\rangle / (\norm{\Delta_i}\norm{\Delta_j})$, which is $1$ for identical deltas and averages $0$ for independent random matrices of the same size. Averaged over the 53 tensors, the off-diagonal entrywise cosine is $0.083$ (Table~\ref{tab:cosine}), comfortably above zero but far below the subspace overlap of $0.179$. The two matrices are strongly related---the correlation across the 28 distinct pairs is $0.843$---but they are not the same measurement, and the pairs that separate are informative. MNIST--SVHN has both the largest entrywise cosine ($0.214$) and the largest subspace overlap ($0.338$); DTD--SUN397 has a near-average entrywise cosine ($0.078$) yet an overlap of $0.206$, more than twice EuroSAT--Cars ($0.144$) whose entrywise cosine is only slightly lower ($0.068$). Entrywise similarity explains much of the ordering and not all of it; the remainder reflects row-space structure that an entrywise cosine does not measure.

\begin{table}[t]
\centering
\caption{\textbf{Entrywise versus subspace similarity} (ViT-B/16). Upper triangle: entrywise cosine between flattened task deltas, averaged over the 53 tensors. Lower triangle: subspace overlap $O_{ij}$ of Eq.~\eqref{eq:overlap} at $r{=}16$. Independent random spaces of these widths would sit at $0.018$ for the subspace overlap; independent random matrices at $0$ for the entrywise cosine. The two measures correlate at $0.843$ across the 28 distinct pairs.}
\label{tab:cosine}
\setlength{\tabcolsep}{2.6pt}
\footnotesize
\begin{tabular}{@{}lcccccccc@{}}
\toprule
 & MNIST & EurS & GTS & SVHN & DTD & RES & Cars & SUN \\
\midrule
MNIST  & --    & 0.077 & 0.109 & 0.214 & 0.074 & 0.074 & 0.072 & 0.069 \\
EuroSAT& 0.156 & --    & 0.075 & 0.080 & 0.078 & 0.104 & 0.068 & 0.072 \\
GTSRB  & 0.241 & 0.189 & --    & 0.122 & 0.073 & 0.073 & 0.071 & 0.071 \\
SVHN   & 0.338 & 0.180 & 0.278 & --    & 0.071 & 0.072 & 0.067 & 0.068 \\
DTD    & 0.134 & 0.177 & 0.142 & 0.138 & --    & 0.082 & 0.073 & 0.078 \\
RESISC & 0.140 & 0.272 & 0.152 & 0.142 & 0.207 & --    & 0.071 & 0.084 \\
Cars   & 0.139 & 0.144 & 0.145 & 0.130 & 0.165 & 0.179 & --    & 0.074 \\
SUN397 & 0.130 & 0.176 & 0.143 & 0.132 & 0.206 & 0.231 & 0.209 & -- \\
\bottomrule
\end{tabular}
\end{table}

\paragraph{Depth profile.}
Table~\ref{tab:layer} gives the aggregate retained ratio per transformer block, and Figure~\ref{fig:layer}a plots it. The profile varies between $0.255$ and $0.308$ over blocks $0$--$11$, a spread of 5.3 points, somewhat larger than the 4.3-point change in the aggregate ratio between $r{=}8$ and $r{=}20$; no single block stands apart, with blocks 0 and 11 lowest. We find no evidence that interference is concentrated in a few layers that could be left unmerged. The one outlier is the group of projection and embedding tensors, which retains $0.442$---much higher, but they carry about $2\%$ of the aggregate delta norm (Appendix~\ref{app:setup}) and cannot move the merged score appreciably in either direction.

\begin{table}[t]
\centering
\caption{\textbf{Aggregate retained ratio per block} (ViT-B/16, $r{=}16$, \%). Each entry is $\norm{\sum_i\tilde\btau_i}/\sum_i\norm{\btau_i}$ pooled over the four tensors of that block. No block stands apart.}
\label{tab:layer}
\setlength{\tabcolsep}{1.4pt}
\scriptsize
\begin{tabular}{@{}lcccccccccccc|c@{}}
\toprule
Block & 0 & 1 & 2 & 3 & 4 & 5 & 6 & 7 & 8 & 9 & 10 & 11 & emb \\
\midrule
Retained & 25.5 & 28.7 & 28.0 & 28.7 & 28.7 & 29.2 & 29.5 & 30.1 & 30.4 & 30.8 & 30.3 & 27.2 & 44.2 \\
\bottomrule
\end{tabular}
\end{table}

\paragraph{Why the sum shrinks more than the parts.}
The per-task and aggregate retention figures measure different things, and the gap is the mechanism. Per task at $r{=}16$, $\norm{\tilde\btau_i}/\norm{\btau_i}$ averages $0.78$ (range $0.72$ on MNIST and SVHN to $0.84$ on Cars): each task keeps most of its own update. Per tensor, the merged sum $\norm{\sum_i \tilde\btau_i}$ is only $30.4\%$ of $\sum_i\norm{\btau_i}$ (Table~\ref{tab:retention}). If the eight projected deltas were mutually orthogonal, a per-task retention of $0.778$ would give $0.778/\sqrt{8} = 27.5\%$; the measured $30.4\%$ is $2.9$ points above that. This is consistent with the projection removing most of the coherent cross-task accumulation while each task keeps most of its own update.

Figure~\ref{fig:layer} localizes this in depth. Retention varies modestly across the twelve blocks, from $0.255$ at block $0$ to $0.308$ at block $9$ and $0.272$ at block $11$. The five projection and embedding tensors retain $0.442$, higher than any block, but they carry about $2\%$ of the aggregate delta norm (Appendix~\ref{app:setup}) and contribute little either way.

\begin{figure}[t]
\centering
\begin{tikzpicture}[x=1cm,y=1cm,font=\scriptsize]
  \begin{scope}[shift={(1.05,0)}]
    \foreach \v/\lab in {0/0, 1.2/20, 2.4/40, 3.6/60} {
      \draw[gray!30] (0,\v) -- (6.30,\v);
      \node[left,font=\tiny,gray!55!black] at (-0.06,\v) {\lab};}
    \foreach \i/\v in {0/.255,1/.287,2/.280,3/.287,4/.287,5/.292,6/.295,7/.301,8/.304,9/.308,10/.303,11/.272} {
      \fill[oursblue!70] (\i*0.48+0.06,0) rectangle ++(0.34,\v*6);
      \node[below,font=\tiny,gray!45!black,rotate=90,anchor=east,inner sep=1pt]
        at (\i*0.48+0.23,-0.04) {\i};}
    \fill[baseorange!85] (6.02,0) rectangle ++(0.34,0.442*6);
    \node[below,font=\tiny,rotate=90,anchor=east,inner sep=1pt] at (6.19,-0.04) {proj/emb};
    \draw[gray!60] (0,0) -- (6.55,0);
    \node[font=\tiny,gray!55!black] at (3.15,-1.02) {(a) retained fraction, by block (\%)};
    \node[font=\tiny,gray!55!black,rotate=90] at (-0.72,1.15) {retained};
  \end{scope}
  \begin{scope}[shift={(2.35,-4.55)}]
    \foreach \i/\nm/\v/\sd/\lab in {0/SUN397/.263/oursblue!35/26.3,
        1/Cars/.341/oursblue!35/34.1, 2/DTD/.367/oursblue!45/36.7,
        3/RESISC45/.401/oursblue!55/40.1, 4/EuroSAT/.546/oursblue!75/54.6,
        5/GTSRB/.551/oursblue!75/55.1, 6/SVHN/.624/oursblue!90/62.4,
        7/MNIST/.656/oursblue!90/65.6} {
      \fill[\sd] (0,\i*0.36) rectangle ++(\v*5.2,0.26);
      \node[left,font=\tiny] at (-0.06,\i*0.36+0.13) {\nm};
      \node[right,font=\tiny,gray!40!black] at (\v*5.2+0.05,\i*0.36+0.13) {\lab};}
    \draw[gray!60] (0,-0.08) -- (0,2.90);
    \foreach \v/\x in {0/0, .2/20, .4/40, .6/60} {
      \draw[gray!30] (\v*5.2,-0.06) -- (\v*5.2,2.86);
      \node[below,font=\tiny,gray!55!black] at (\v*5.2,-0.10) {\x};}
    \node[font=\tiny,gray!55!black] at (1.72,-1.02) {(b) top-16 energy share, per task (\%)};
  \end{scope}
\end{tikzpicture}
\caption{\textbf{Where the removal lands.} (a) Retained fraction per transformer block under the aggregate definition of Table~\ref{tab:retention} ($r{=}16$, $\%$); no block stands apart. (b) Fraction of each task's delta energy captured by its own top-16 right-singular directions. Datasets whose updates concentrate in few directions score higher; the two are correlated at $r{=}0.85$ across the eight tasks.}
\label{fig:layer}
\end{figure}

\paragraph{Spectral concentration and accuracy.}
Figure~\ref{fig:layer}(b) reports the share of each task's delta energy that its own top-16 right-singular directions carry, from $0.263$ on SUN397 to $0.656$ on MNIST. Across the eight datasets this share correlates with final accuracy at Pearson $r{=}0.85$. The method does not truncate a delta to its own top-16 directions, so $\phi$ is a description of the update, not of what the merge keeps. SUN397 is both the least concentrated and our weakest task ($74.97\%$), and the rank does not help it: raising $r$ from $16$ to $32$ buys SUN397 $0.8$ points while costing $2.4$ on the eight-task average (Table~\ref{tab:rank_full}). With eight tasks the correlation is suggestive, not a causal finding.

The correlation does not extend to the margin: against TIES-Merging, concentration correlates only weakly, at $r{=}-0.31$. Concentration is associated with the accuracy a merge reaches, not with the distance between methods. The gain reported in Table~\ref{tab:main} is therefore not a restatement of ``some tasks are easier than others.''

\section{The Role of the Singular Basis, in Detail}
\label{app:role}

Section~\ref{sec:method_role} states the objection of ESM~\cite{li2026esm}---that the singular value ordering of $\Delta_i$ need not track functional importance---and answers it briefly. This appendix gives the expanded version.

\paragraph{Compression versus projection.}
The objection describes a \emph{compression}: $\Delta_i$ is replaced by $\Delta_i\bm{V}_i\bm{V}_i^\top$, so everything outside its own top-$r$ space is deleted, and a low-$\sigma_i$ direction with high functional value is lost without recourse. Our re-implementation of the compression-based TSV-M loses 1.8 points against its published score (Section~\ref{sec:repro_gap}), whether this compression accounts for the gap is not tested here. In~\eqref{eq:proj} the basis plays a different role: $\mathcal{S}_i$ is never used to project $\Delta_i$---it is excluded from $\mU^{(i)}_\cup$---and the only thing deleted is $\Delta_i$'s component inside \emph{other} tasks' top-$r$ spaces, which can include part of $\mathcal{S}_i$. A direction that is low-energy for task $i$ and not dominant for any neighbour is retained in full, at any $\sigma_i$, so the singular-value ordering never decides what task $i$ keeps.

\paragraph{What the objection does reach.}
It survives for one configuration: a direction that is low-$\sigma_i$, dominant for exactly one neighbour $j$, and at the same time carries real function for task $i$. Equations~\eqref{eq:svd}--\eqref{eq:proj} cannot tell that case apart from a genuinely shared direction, and the broad rank plateau of Figure~\ref{fig:rank}---while consistent with the operator removing mostly redundancy---is not a functional probe. Two claims should therefore be kept apart. \emph{Verified}: the operator does not truncate $\Delta_i$ to $\mathcal{S}_i$; the part of $\mathcal{S}_i$ lying in the peer union is, however, deleted (Proposition~\ref{prop:blockdiag}). \emph{Unverified}: whether the deleted directions carry function for the task that loses them, and whether an activation-space basis of the kind ESM fits would select a better deletion set. Settling the second would need per-example forward passes over task data, which is exactly the cost data-freedom avoids.

\section{Qualitative Results}
\label{app:qual}

Figure~\ref{fig:qual} (Section~\ref{sec:main_results}) shows the single-image comparison; the numbered results below refer to it.

\paragraph{The overlap is large and one-sided.} Of 87{,}432 evaluation images, 7{,}102 satisfy the condition that \ours{} is correct while both entrywise baselines err; the reverse condition---both baselines correct, \ours{} wrong---holds on 995. The ratio is 7.1 to 1. Table~\ref{tab:overlap} breaks this down by dataset. The ratio is highest where the accuracy gap is largest (EuroSAT 29.5, GTSRB 195) and lowest on MNIST and SUN397 (3.6 and 2.9), where the baselines are already close to their ceiling or where all methods struggle. GTSRB is extreme because both baselines score far below \ours{} on that dataset (71.1 and 71.9 versus 96.3) while almost never being right where \ours{} is wrong; we do not test why.

\begin{table}[t]
\centering
\caption{\textbf{Per-image agreement with the two entrywise baselines} (ViT-B/16, counts of evaluation images). ``Ours only'' counts images where \ours{} is correct and both TIES-Merging and Task Arithmetic err; ``baselines only'' is the reverse.}
\label{tab:overlap}
\setlength{\tabcolsep}{4pt}
\small
\begin{tabular}{@{}lcccr@{}}
\toprule
Dataset & Ours only & Baselines only & $n$ & Ratio \\
\midrule
MNIST    & 51   & 14  & 10{,}000 & 3.6 \\
EuroSAT  & 354  & 12  & 2{,}700  & 29.5 \\
GTSRB    & 2{,}340 & 12 & 12{,}630 & 195.0 \\
SVHN     & 765  & 138 & 26{,}032 & 5.5 \\
DTD      & 349  & 24  & 1{,}880  & 14.5 \\
RESISC45 & 702  & 79  & 6{,}300  & 8.9 \\
Cars     & 962  & 171 & 8{,}040  & 5.6 \\
SUN397   & 1{,}579 & 545 & 19{,}850 & 2.9 \\
\midrule
Total    & 7{,}102 & 995 & 87{,}432 & 7.1 \\
\bottomrule
\end{tabular}
\end{table}

\paragraph{The error structure differs between the two families.} On Cars and DTD, TIES-Merging and Task Arithmetic both predict a class adjacent to the ground truth in the fine-grained hierarchy---a different make of car, a similar texture category---whereas \ours{} is correct. On EuroSAT and RESISC45 the baselines tend to predict a plausible same-domain class (urban versus industrial land use, a sports field versus a meadow), These are observations on the sampled images of Figure~\ref{fig:qual}; we do not test which directions carry the decision boundary.

\section{Per-Class Behaviour}
\label{app:perclass}

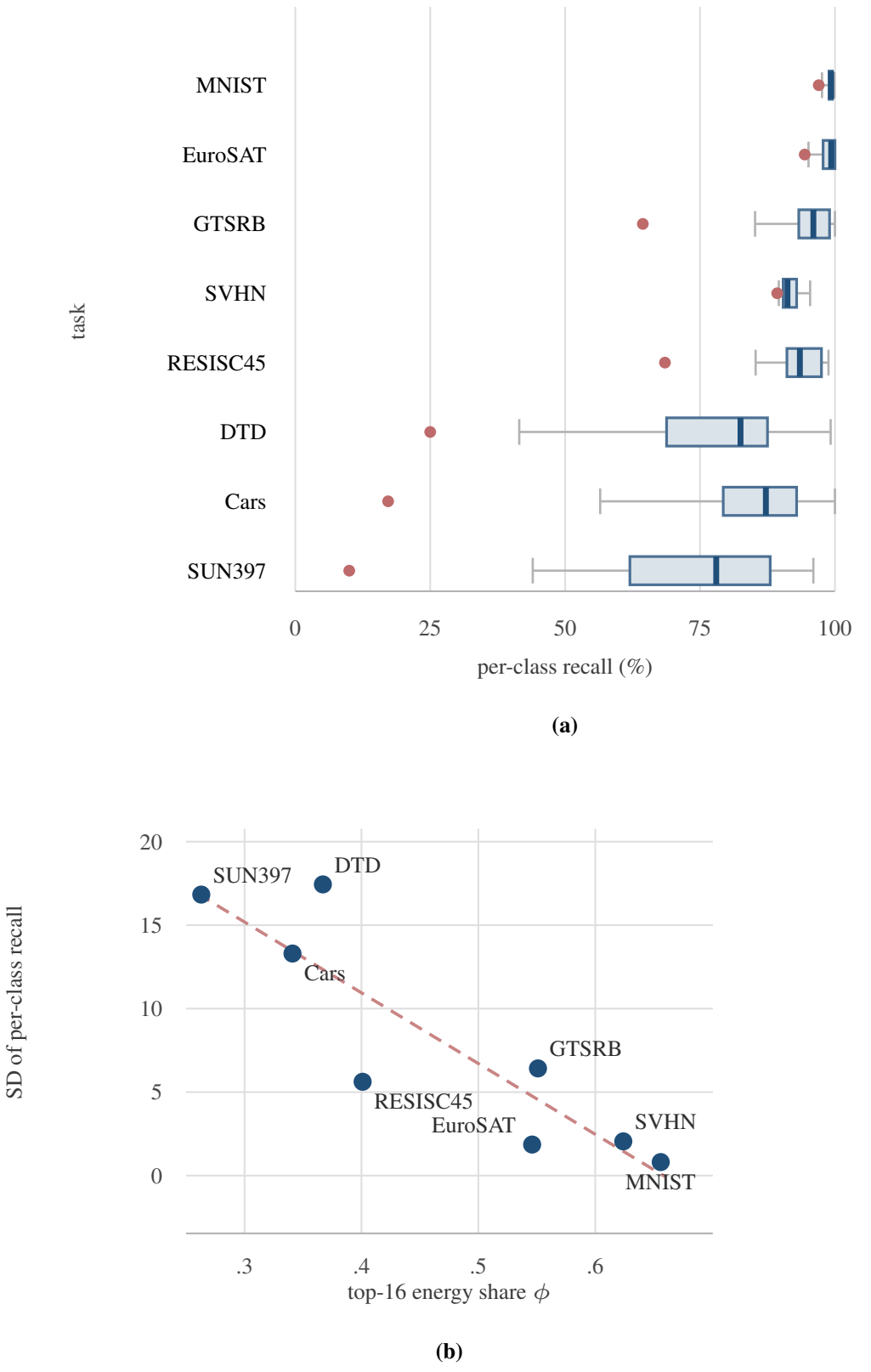
\begin{figure}[t]
\centering
\begin{tikzpicture}[font=\scriptsize]
  \begin{scope}[shift={(1.95,0.60)}]
    \foreach \v/\lab in {0/0, 25/25, 50/50, 75/75, 100/100} {
      \draw[gray!25] (\v*0.042,0) -- (\v*0.042,4.55);
      \node[below,font=\tiny,gray!55!black] at (\v*0.042,-0.10) {\lab};}
    \draw[gray!60] (0,0) -- (4.20,0);
    \foreach \i/\nm/\lo/\qone/\med/\qthree/\hi/\mn in {
        0/SUN397/44.0/62.0/78.0/88.0/96.0/10.0,
        1/Cars/56.5/79.3/87.2/92.9/100.0/17.2,
        2/DTD/41.5/68.8/82.5/87.5/99.2/25.0,
        3/RESISC45/85.3/91.1/93.5/97.5/98.8/68.5,
        4/SVHN/89.6/90.4/91.2/92.9/95.4/89.3,
        5/GTSRB/85.2/93.3/96.0/99.0/100.0/64.4,
        6/EuroSAT/95.1/97.8/99.3/100.0/100.0/94.4,
        7/MNIST/97.6/98.9/99.3/99.5/99.9/97.0} {
      \pgfmathsetmacro{\y}{\i*0.54+0.16}
      \draw[gray!60,line width=0.5pt] (\lo*0.042,\y) -- (\hi*0.042,\y);
      \draw[gray!60,line width=0.5pt] (\lo*0.042,\y-0.10) -- (\lo*0.042,\y+0.10);
      \draw[gray!60,line width=0.5pt] (\hi*0.042,\y-0.10) -- (\hi*0.042,\y+0.10);
      \fill[oursblue!16] (\qone*0.042,\y-0.11) rectangle (\qthree*0.042,\y+0.11);
      \draw[oursblue!80,line width=0.6pt] (\qone*0.042,\y-0.11) rectangle (\qthree*0.042,\y+0.11);
      \draw[oursblue,line width=1.3pt] (\med*0.042,\y-0.11) -- (\med*0.042,\y+0.11);
      \fill[basered!85] (\mn*0.042,\y) circle (1.3pt);
      \node[left,font=\tiny] at (-0.10,\y) {\nm};}
    \node[font=\tiny,gray!45!black] at (2.10,-0.60) {per-class recall (\%)};
    \node[font=\tiny,gray!55!black,rotate=90] at (-1.70,2.10) {task};
    \node[font=\tiny\bfseries] at (2.10,-1.05) {(a)};
  \end{scope}
  \begin{scope}[shift={(1.10,-3.95)}]
    \foreach \v/\lab in {0/0, 5/5, 10/10, 15/15, 20/20} {
      \draw[gray!25] (0,\v*0.13) -- (4.10,\v*0.13);
      \node[left,font=\tiny,gray!55!black] at (-0.06,\v*0.13) {\lab};}
    \foreach \v/\lab in {0.3/.3, 0.4/.4, 0.5/.5, 0.6/.6} {
      \pgfmathsetmacro{\tx}{(\v-0.25)*9.1}
      \draw[gray!25] (\tx,-0.45) -- (\tx,2.70);
      \node[below,font=\tiny,gray!55!black] at (\tx,-0.52) {\lab};}
    \draw[gray!60] (0,-0.45) -- (4.10,-0.45);
    \pgfmathsetmacro{\ax}{(0.26-0.25)*9.1}
    \pgfmathsetmacro{\ay}{(-42.38*0.26+27.90)*0.13}
    \pgfmathsetmacro{\bx}{(0.66-0.25)*9.1}
    \pgfmathsetmacro{\by}{(-42.38*0.66+27.90)*0.13}
    \draw[basered!70,densely dashed,line width=0.7pt] (\ax,\ay) -- (\bx,\by);
    \foreach \c/\s/\nm/\an/\dx/\dy in {
        0.656/0.81/MNIST/north/0/-1,
        0.546/1.86/EuroSAT/south east/-2/1,
        0.624/2.05/SVHN/south west/1/1,
        0.551/6.42/GTSRB/south west/1/1,
        0.401/5.62/RESISC45/north west/1/-1,
        0.367/17.44/DTD/south west/1/1,
        0.341/13.30/Cars/north west/1/-1,
        0.263/16.83/SUN397/south west/1/1} {
      \pgfmathsetmacro{\px}{(\c-0.25)*9.1}
      \pgfmathsetmacro{\py}{\s*0.13}
      \fill[oursblue] (\px,\py) circle (2.0pt);
      \node[anchor=\an,font=\tiny,inner sep=1.4pt,gray!35!black,shift={(\dx pt,\dy pt)}]
        at (\px,\py) {\nm};}
    \node[font=\tiny,gray!45!black] at (2.05,-0.92) {top-16 energy share $\phi$};
    \node[font=\tiny,gray!45!black,rotate=90] at (-1.32,1.35) {SD of per-class recall};
    \node[font=\tiny\bfseries] at (2.05,-1.38) {(b)};
  \end{scope}
\end{tikzpicture}
\caption{\textbf{Where the error lives.} (a) Per-class recall for each task: box spans the interquartile range, thick bar the median, whiskers the 5th/95th percentiles, and the red dot the worst class. (b) The standard deviation of per-class recall against the concentration $\phi$ of the task's update (the top-16 energy share of Figure~\ref{fig:layer}b). Spread falls almost linearly in $\phi$ ($r{=}-0.89$, eight tasks).}
\label{fig:perclass}
\end{figure}

The per-dataset averages hide how the error is distributed over categories. Table~\ref{tab:perclass} reports the distribution of per-class recall for each task, computed from the merged model's predictions on every evaluation image, and Figure~\ref{fig:perclass}(a) shows the same distributions as box plots.

\begin{table}[t]
\centering
\caption{\textbf{Distribution of per-class recall} for \ours{} at $r{=}16$ (ViT-B/16), over the classes of each dataset. Mean is the unweighted mean of per-class recall (macro average): it coincides with the accuracy of Table~\ref{tab:main} to two decimals on MNIST, EuroSAT, DTD and SUN397 and is lower on the other four (e.g.\ 94.61 versus 96.26 on GTSRB). SD is the population standard deviation over classes and quartiles are linearly interpolated; the last two columns give the fraction of classes at or above $90\%$ recall and below $50\%$.}
\label{tab:perclass}
\setlength{\tabcolsep}{3.2pt}
\footnotesize
\begin{tabular}{@{}lcccccccc@{}}
\toprule
Dataset & \#cls & Mean & SD & Min & Med & $Q_1$/$Q_3$ & $\ge90$ & $<50$ \\
\midrule
MNIST    & 10  & 99.07 & 0.81  & 97.0 & 99.3 & 98.9/99.5   & 1.000 & 0.000 \\
EuroSAT  & 10  & 98.44 & 1.86  & 94.4 & 99.3 & 97.8/100.0  & 1.000 & 0.000 \\
GTSRB    & 43  & 94.61 & 6.42  & 64.4 & 96.0 & 93.3/99.0   & 0.907 & 0.000 \\
SVHN     & 10  & 91.90 & 2.05  & 89.3 & 91.2 & 90.4/92.9   & 0.800 & 0.000 \\
RESISC45 & 45  & 93.10 & 5.62  & 68.5 & 93.5 & 91.1/97.5   & 0.844 & 0.000 \\
Cars     & 196 & 84.11 & 13.30 & 17.2 & 87.2 & 79.3/92.9   & 0.378 & 0.026 \\
DTD      & 47  & 77.23 & 17.44 & 25.0 & 82.5 & 68.8/87.5   & 0.191 & 0.106 \\
SUN397   & 397 & 74.97 & 16.83 & 10.0 & 78.0 & 62.0/88.0   & 0.244 & 0.081 \\
\bottomrule
\end{tabular}
\end{table}

\paragraph{The deficit is a tail, not a shift.}
For MNIST, EuroSAT and GTSRB, per-class recall is tightly clustered: standard deviations of $0.81$, $1.86$ and $6.42$ points, worst classes at $97.0$, $94.4$ and $64.4$. No class falls below $50\%$. The average is therefore close to what every class achieves. The picture changes on the fine-grained tasks. On DTD, $10.6\%$ of classes fall below $50\%$ recall while $19.1\%$ sit at or above $90\%$; the median class sits at $82.5\%$. SUN397 is similar ($8.1\%$ below $50\%$, $24.4\%$ at or above $90\%$, median $78.0\%$); Cars has a long lower tail (worst class $17.2\%$) with only $2.6\%$ of its 196 classes below $50\%$.

A single dataset average therefore conceals two different regimes: what a merge must preserve for a coarse task is a small decision boundary over well-separated classes. For a fine-grained task it must preserve hundreds of boundaries of uneven difficulty, and the aggregate is dominated by the ones it loses.

\paragraph{The spread is predicted by concentration.}
Figure~\ref{fig:perclass}(b) plots the per-class standard deviation against the spectral concentration $\phi$ of Figure~\ref{fig:layer}(b). The relation is close to linear over the full range---$\mathrm{SD}\approx -42.4\,\phi + 27.9$, $r{=}-0.893$, $R^2{=}0.80$---and is a stronger statement than the correlation between $\phi$ and accuracy already noted in Appendix~\ref{app:geometry}. Accuracy tells us that low-concentration tasks score worse; the spread tells us \emph{how} they score worse. The most concentrated delta (MNIST, $\phi{=}0.656$) has the most uniform per-class recall, the least concentrated (SUN397, $\phi{=}0.263$) one of the most spread. We do not test a per-class mechanism; that would require per-class task vectors.

\begin{table}[t]
\centering
\caption{\textbf{Most frequent confusions} of \ours{} at $r{=}16$, as (predicted, true) class-index pairs with the number of affected evaluation images. Indices follow each dataset's native label order, so adjacency in the table is not semantic adjacency; the table localises where errors concentrate rather than naming the confusion. The largest single pair on every dataset is listed, together with the next two.}
\label{tab:confusion}
\setlength{\tabcolsep}{4pt}
\small
\begin{tabular}{@{}llll@{}}
\toprule
Dataset & Pair 1 & Pair 2 & Pair 3 \\
\midrule
MNIST    & (4,9) 17   & (7,2) 15   & (7,9) 9 \\
EuroSAT  & (0,5) 8    & (1,2) 5    & (6,5) 5 \\
GTSRB    & (16,10) 18 & (40,42) 18 & (14,38) 14 \\
SVHN     & (1,4) 106  & (4,1) 92   & (2,1) 82 \\
RESISC45 & (13,22) 27 & (27,7) 13  & (23,11) 12 \\
Cars     & (83,82) 20 & (70,63) 20 & (12,21) 19 \\
DTD      & (30,10) 13 & (31,28) 11 & (0,22) 8 \\
SUN397   & (287,202) 25 & (217,168) 20 & (17,16) 17 \\
\bottomrule
\end{tabular}
\end{table}

\paragraph{Errors are pairwise, not diffuse.}
Table~\ref{tab:confusion} lists the three most frequent (predicted, true) pairs for each task. Two regimes again appear. On MNIST and EuroSAT the largest pair concerns fewer than 20 of the $1{,}000$ and $270$ test images per class, so no single boundary accounts for the residual error; the remaining mistakes are scattered across many pairs. On SVHN, by contrast, a single unordered pair between classes 4 and 1 accounts for 198 errors---both directions, at comparable counts ($106$ and $92$)---far more than the third-ranked pair (82). On Cars the two largest pairs, (83,82) and (70,63), each account for 20 errors. We do not test whether these confusions arise from the merge or are already present in the expert.

The concentration and confusion analyses together locate where a data-free merge loses accuracy, but neither identifies a repair within the family of linear projections of Section~\ref{sec:method}.

\section{Per-Task Confidence Intervals}
\label{app:ci}

Table~\ref{tab:ci} reports the bootstrap intervals behind the aggregates in Section~\ref{sec:exp}. The merge is deterministic, so the variability shown is evaluation-sampling variability under the held-out protocol, not seed variance over training. The intervals are 2{,}000-resample percentile intervals, and because the eight task estimates are computed on disjoint image sets they can be combined into an average interval by resampling each task independently.

\begin{table}[t]
\centering
\caption{\textbf{Per-dataset bootstrap intervals} for \ours{} at $r{=}16$ (ViT-B/16). Intervals are 95\% percentile intervals from 2{,}000 resamples of the evaluation set.}
\label{tab:ci}
\setlength{\tabcolsep}{4pt}
\small
\begin{tabular}{@{}lccc@{}}
\toprule
Dataset & Acc.\ (\%) & 95\% interval & $n$ \\
\midrule
MNIST    & 99.07 & $[98.88,\,99.26]$ & 10{,}000 \\
EuroSAT  & 98.44 & $[97.96,\,98.93]$ & 2{,}700 \\
GTSRB    & 96.26 & $[95.91,\,96.58]$ & 12{,}630 \\
SVHN     & 92.16 & $[91.83,\,92.47]$ & 26{,}032 \\
DTD      & 77.23 & $[75.32,\,79.10]$ & 1{,}880 \\
RESISC45 & 93.22 & $[92.57,\,93.81]$ & 6{,}300 \\
Cars     & 84.25 & $[83.46,\,85.06]$ & 8{,}040 \\
SUN397   & 74.97 & $[74.37,\,75.55]$ & 19{,}850 \\
\midrule
Average  & 89.45 & $[89.17,\,89.75]$ & 87{,}432 \\
\bottomrule
\end{tabular}
\end{table}

\begin{figure}[t]
\centering
\begin{tikzpicture}[font=\scriptsize]
  \begin{scope}[shift={(1.55,0.30)}]
    \foreach \v in {70,75,80,85,90,95,100} {
      \pgfmathsetmacro{\gx}{(\v-70)*0.235}
      \draw[gray!22] (\gx,0) -- (\gx,5.05);
      \node[below,font=\tiny,gray!55!black] at (\gx,-0.08) {\v};}
    \draw[gray!60] (0,0) -- (7.15,0);
    \foreach \i/\nm/\mOne/\mLo/\mHi/\tOne/\tLo/\tHi in {
        0/SUN397/74.97/74.37/75.55/74.85/74.25/75.44,
        1/Cars/84.25/83.46/85.06/84.85/84.09/85.63,
        2/RESISC/93.22/92.57/93.81/92.70/92.05/93.32,
        3/DTD/77.23/75.32/79.10/75.90/73.94/77.77,
        4/SVHN/92.16/91.83/92.47/91.43/91.08/91.77,
        5/GTSRB/96.26/95.91/96.58/96.03/95.68/96.37,
        6/EuroSAT/98.44/97.96/98.93/98.63/98.19/99.07,
        7/MNIST/99.07/98.88/99.26/99.23/99.06/99.39} {
      \pgfmathsetmacro{\y}{\i*0.60+0.36}
      \pgfmathsetmacro{\xOne}{(\mOne-70)*0.235}
      \pgfmathsetmacro{\xLo}{(\mLo-70)*0.235}
      \pgfmathsetmacro{\xHi}{(\mHi-70)*0.235}
      \pgfmathsetmacro{\xTOne}{(\tOne-70)*0.235}
      \pgfmathsetmacro{\xTLo}{(\tLo-70)*0.235}
      \pgfmathsetmacro{\xTHi}{(\tHi-70)*0.235}
      \draw[baseorange!85,line width=0.9pt] (\xTLo,\y-0.13) -- (\xTHi,\y-0.13);
      \draw[baseorange!85,line width=0.9pt] (\xTLo,\y-0.17) -- (\xTLo,\y-0.09);
      \draw[baseorange!85,line width=0.9pt] (\xTHi,\y-0.17) -- (\xTHi,\y-0.09);
      \fill[baseorange] (\xTOne,\y-0.13) circle (1.7pt);
      \draw[oursblue,line width=0.9pt] (\xLo,\y+0.13) -- (\xHi,\y+0.13);
      \draw[oursblue,line width=0.9pt] (\xLo,\y+0.09) -- (\xLo,\y+0.17);
      \draw[oursblue,line width=0.9pt] (\xHi,\y+0.09) -- (\xHi,\y+0.17);
      \fill[oursblue] (\xOne,\y+0.13) circle (1.7pt);
      \node[left,font=\tiny] at (-0.10,\y) {\nm};}
    \node[font=\tiny,gray!45!black] at (3.52,-0.62) {top-1 accuracy (\%)};
    \begin{scope}[shift={(0,5.30)}]
      \draw[oursblue,line width=0.9pt] (0,0.06) -- (0.42,0.06);
      \fill[oursblue] (0.21,0.06) circle (1.7pt);
      \node[right,font=\tiny] at (0.46,0.06) {own checkpoints};
      \draw[baseorange!85,line width=0.9pt] (2.55,0.06) -- (2.97,0.06);
      \fill[baseorange] (2.76,0.06) circle (1.7pt);
      \node[right,font=\tiny] at (3.01,0.06) {TALL release};
    \end{scope}
  \end{scope}
\end{tikzpicture}
\caption{\textbf{Bootstrap intervals, both checkpoint releases.} Point estimates and 95\% percentile intervals over the evaluation set ($2{,}000$ resamples) for the eight tasks, our ViT-B/16 checkpoints and the independently trained TALL release. Intervals are narrow where the test set is large and the task is saturated, and widest on DTD---the smallest test set (1{,}880 images) at a mid-range accuracy.}
\label{fig:forest}
\end{figure}
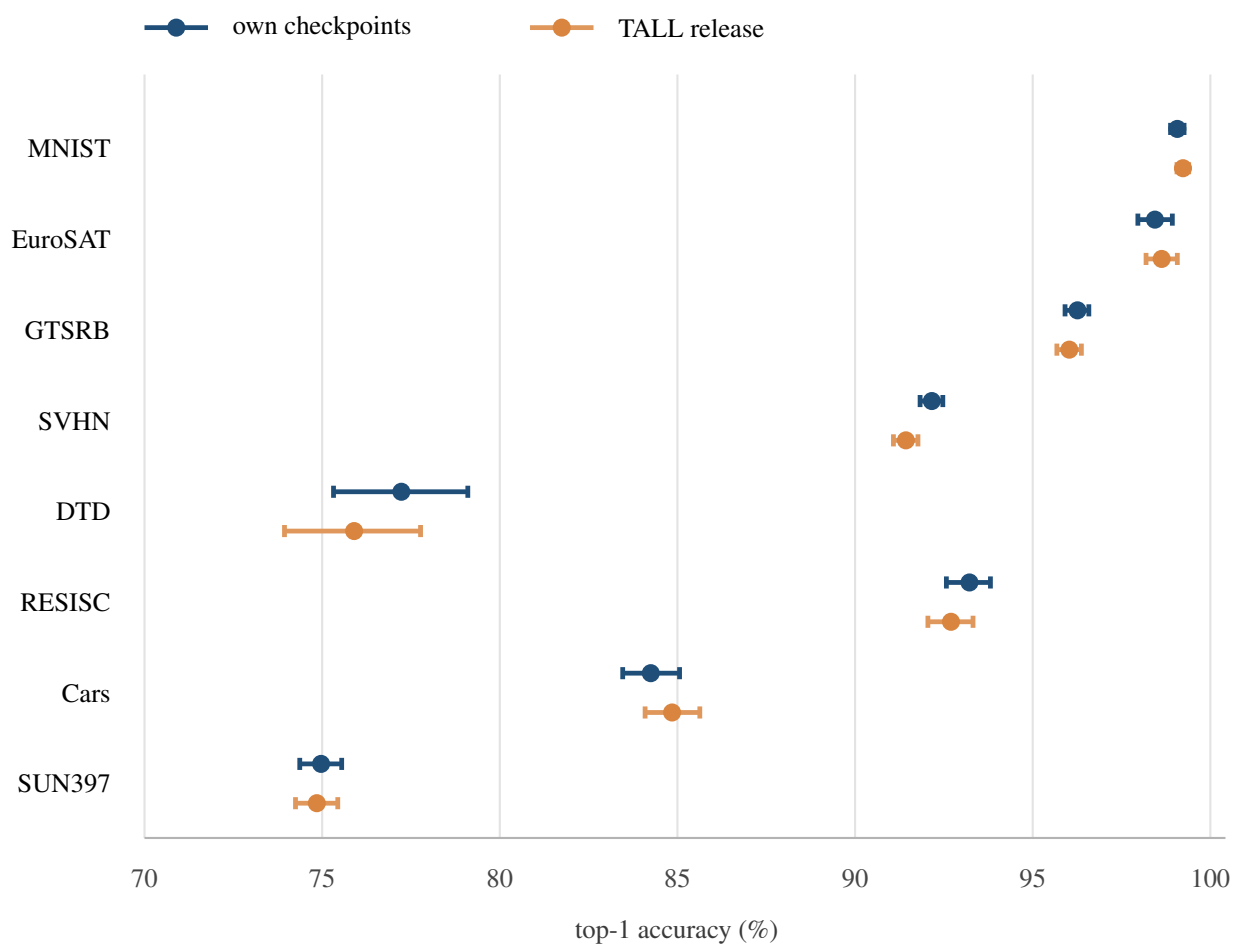

Figure~\ref{fig:forest} plots the intervals rather than tabulating them, so the two releases can be compared row by row. The same protocol applied to the independently trained TALL ViT-B/16 release gives 89.20 with interval $[88.90,\,89.51]$. The two intervals overlap over most of their range, as expected if the difference between releases (0.25 points) is smaller than the evaluation noise, and the TIES-Merging point estimate of 77.57 lies below both.

\paragraph{What the intervals do and do not measure.}
The widths track the test-set size, not the difficulty of the merge. DTD has the widest interval ($\pm1.9$ points) because it has the smallest evaluation set ($1{,}880$ images); SUN397, with ten times as many images, is accurate to about $\pm0.6$ despite scoring 2.3 points below DTD. WUDI's 88.47, 0.98 points below ours, lies outside both average intervals; we report no interval for WUDI. The TIES, Task Arithmetic and DARE scores of Table~\ref{tab:main} differ from ours by more than our per-dataset interval half-width on every one of the eight tasks. The intervals also do not speak to checkpoint-to-checkpoint variation during fine-tuning; the two-release comparison is our substitute for that, since each release comes from a different pipeline.

\section{Paired Significance Over Shared Test Items}
\label{app:mcnemar}

The intervals of Appendix~\ref{app:ci} treat each method's accuracy as an independent estimate. It is not: every method is evaluated on the same test images, so the relevant question is not whether two marginal intervals overlap but whether the two methods disagree on the same items in a consistent direction. We therefore report exact McNemar tests, on the $b$ items \ours{} gets right and the baseline gets wrong and the $c$ items the baseline gets right and \ours{} gets wrong. Under the null hypothesis $b$ and $c$ are draws from a fair coin, and the exact two-sided $p$-value is the binomial tail with $n = b+c$. Table~\ref{tab:mcnemar} gives the result per task for both training-free baselines.

\begin{table}[t]
\centering
\caption{\textbf{Paired McNemar tests on shared test items} (ViT-B/16). $b$: items \ours{} classifies correctly and the baseline does not. $c$: the reverse. $\Delta$ is the accuracy difference in percentage points and $p$ the exact two-sided binomial test. Pooling adds the discordant counts across all eight tasks; the pooled $\Delta$ is $(b-c)/87{,}432$, an item-weighted difference, and therefore differs from the difference of the equal-weight averages in Table~\ref{tab:main}.}
\label{tab:mcnemar}
\setlength{\tabcolsep}{1.35pt}
\scriptsize
\begin{tabular}{@{}lcccccccc@{}}
\toprule
 & \multicolumn{4}{c}{vs.\ Task Arithmetic} & \multicolumn{4}{c}{vs.\ TIES-Merging} \\
\cmidrule(lr){2-5}\cmidrule(lr){6-9}
Task & $b$ & $c$ & $\Delta$ & $p$ & $b$ & $c$ & $\Delta$ & $p$ \\
\midrule
MNIST    & 209  & 17  & $+1.92$  & $3\times10^{-43}$  & 58   & 22  & $+0.36$  & $7.0\times10^{-5}$ \\
EuroSAT  & 463  & 14  & $+16.63$ & $2\times10^{-117}$ & 436  & 13  & $+15.67$ & $6\times10^{-111}$ \\
GTSRB    & 3205 & 24  & $+25.19$ & $<10^{-300}$       & 3106 & 24  & $+24.40$ & $<10^{-300}$ \\
SVHN     & 2567 & 187 & $+9.14$  & $<10^{-300}$       & 863  & 391 & $+1.81$  & $3\times10^{-41}$ \\
DTD      & 494  & 30  & $+24.68$ & $2\times10^{-109}$ & 396  & 34  & $+19.26$ & $2\times10^{-79}$ \\
RESISC45 & 1092 & 102 & $+15.71$ & $7\times10^{-210}$ & 859  & 95  & $+12.13$ & $1\times10^{-154}$ \\
Cars     & 1369 & 257 & $+13.83$ & $3\times10^{-183}$ & 1291 & 233 & $+13.16$ & $1\times10^{-177}$ \\
SUN397   & 2186 & 861 & $+6.68$  & $2\times10^{-131}$ & 2437 & 796 & $+8.27$  & $1\times10^{-191}$ \\
\midrule
Pooled   & 11585 & 1492 & $+11.54$ & $<10^{-300}$ & 9446 & 1608 & $+8.96$ & $<10^{-300}$ \\
\bottomrule
\end{tabular}
\end{table}

The paired view sharpens the argument in two places where the marginal intervals are uninformative. On MNIST, \ours{} leads TIES by only 0.36 points, less than the full width of our own interval (0.38); the paired test still separates them at $7\times10^{-5}$ because the disagreement is asymmetric, 58 items to 22. Conversely, on SUN397 a moderate margin against TIES (8.27, against 24.40 on GTSRB) coexists with the largest discordant mass (3{,}233 items). Margin and significance are therefore not interchangeable summaries, and the pooled row is the honest global statement: over 87{,}432 shared test items, \ours{} is correct on 11{,}585 items where Task Arithmetic fails and 1{,}492 where it succeeds, and 9{,}446 against 1{,}608 for TIES.

\paragraph{Where the baseline still wins.}
The $c$ column is not noise to be dismissed. TIES beats \ours{} on 391 SVHN items and 233 Cars items, and the SVHN count is large enough to be a systematic difference rather than a tail; one possible cause, not tested here, is TIES's sign election. That is consistent with TIES remaining within two points of \ours{} on SVHN in the main table while trailing by 8.27 to 24.40 points on the six tasks other than MNIST and SVHN. No single merge rule dominates item by item, and a method that combined per-task sign election with per-task subspace protection might recover these items; we did not pursue this because the SVHN margin is small compared with the deficits the same step creates elsewhere.

\paragraph{Multiplicity.}
Sixteen tests are reported across the two panels. A Bonferroni correction at family-wise $\alpha{=}0.05$ requires $p < 3.1\times10^{-3}$; the largest of the sixteen $p$-values is $7.0\times10^{-5}$ (MNIST versus TIES), and every cell is below the threshold. The conclusion does not depend on the correction.

\section{Backbone Scaling, Per Dataset}
\label{app:backbones}

Table~\ref{tab:backbones} of the main text reports eight rows per backbone. Tables~\ref{tab:b32} and~\ref{tab:l14} repeat \ours{} and the three parameter-wise baselines for ViT-B/32 and ViT-L/14. Summed over the merged tensors and the eight tasks, the task-vector Frobenius norm is 136.28 for ViT-B/32 (52 tensors) and 304.18 for ViT-L/14 (101 tensors); the larger figure tracks the wider hidden dimension rather than a larger interference problem. All numbers come from the TALL release, so cross-method comparisons within a table are apples-to-apples.

\begin{table}[t]
\centering
\caption{\textbf{ViT-B/32, full per-dataset results} (TALL release, \%). \ours{} uses $r{=}16$, untuned.}
\label{tab:b32}
\setlength{\tabcolsep}{2.0pt}
\footnotesize
\begin{tabular}{@{}lccccccccc@{}}
\toprule
Method & MNIST & EurS & GTS & SVHN & DTD & RES & Cars & SUN & Avg \\
\midrule
\ours{} & \textbf{98.91} & \textbf{92.52} & \textbf{90.36} & \textbf{87.04} & \textbf{68.03} & \textbf{85.89} & \textbf{72.21} & \textbf{68.40} & \textbf{82.92} \\
TIES    & 97.17 & 69.74 & 71.96 & 81.93 & 51.65 & 71.06 & 60.50 & 58.46 & 70.31 \\
TA      & 90.00 & 66.74 & 63.44 & 69.20 & 49.26 & 69.05 & 60.41 & 62.39 & 66.31 \\
DARE    & 90.06 & 65.56 & 61.63 & 68.37 & 48.72 & 68.92 & 59.94 & 62.08 & 65.66 \\
\bottomrule
\end{tabular}
\end{table}

\begin{table}[t]
\centering
\caption{\textbf{ViT-L/14, full per-dataset results} (TALL release, \%). \ours{} uses $r{=}16$, untuned.}
\label{tab:l14}
\setlength{\tabcolsep}{2.0pt}
\footnotesize
\begin{tabular}{@{}lccccccccc@{}}
\toprule
Method & MNIST & EurS & GTS & SVHN & DTD & RES & Cars & SUN & Avg \\
\midrule
\ours{} & \textbf{99.52} & \textbf{97.96} & \textbf{97.49} & \textbf{95.23} & \textbf{78.62} & \textbf{92.67} & \textbf{89.39} & \textbf{76.09} & \textbf{90.87} \\
TIES    & 98.88 & 87.37 & 82.67 & 88.98 & 61.22 & 83.32 & 81.21 & 70.16 & 81.73 \\
TA      & 97.74 & 82.48 & 79.41 & 81.91 & 60.64 & 79.89 & 79.91 & 70.38 & 79.05 \\
DARE    & 97.69 & 82.15 & 79.26 & 82.09 & 60.32 & 79.71 & 79.64 & 70.13 & 78.87 \\
\bottomrule
\end{tabular}
\end{table}

Three patterns hold across capacities. On all three backbones \ours{} places first and the training-free baselines keep the same order: TIES above Task Arithmetic and DARE, which are within 0.7 points of each other on every backbone, with the subspace family (TSV-M, Iso-C) between TIES and \ours{}. The margin over TIES is 10.84, 12.61, and 9.14 points on B/16, B/32, and L/14 respectively. Per dataset, \ours{} on B/32 is below B/16 on all eight datasets, by 0.3 points on MNIST to 12.6 on Cars; L/14 averages 1.7 points above B/16 and is higher on six of the eight datasets, with EuroSAT and RESISC45 lower.

\section{Data-Using Baselines and Post-Merge Adaptation}
\label{app:datausing}

Table~\ref{tab:main} contains only data-free rows. Table~\ref{tab:fda} adds the synthetic-anchor FDA rows and two data-free cascades that apply DARE or TIES after the \ours{} projection. This section also reports the one experiment in the paper that uses real data after the merge: distillation of the eight experts into the \ours{} merge, which is the data-using counterpart of the data-free operator and is not a row of Table~\ref{tab:main}.

\paragraph{WUDI's merge scope.} WUDI's released operator merges the four attention and MLP projection weights of each of the twelve blocks---48 tensors---and leaves five two-dimensional tensors at their pretrained values: the text token embedding, the two positional embeddings, the visual projection, and the text projection. Across the eight tasks these five carry between 0.78\% (RESISC45) and 1.81\% (GTSRB) of the squared Frobenius norm of the 53-tensor delta---at most 1.9\% on every task---so its 88.47\% is obtained on a smaller scope than the 53-tensor scope used by every other row of Table~\ref{tab:main}. We do not test whether this scope difference favours either method.

\begin{table}[t]
\centering
\caption{\textbf{FDA and cascade variants, per dataset} (ViT-B/16, \%). The FDA and WUDI-init FDA rows generate synthetic anchors per task and adapt the merged model on them (from the pretrained model and from the WUDI merge, respectively); neither sees a real image. The \ours{}$\to$DARE and \ours{}$\to$TIES rows apply DARE or TIES to the projected task vectors and are data-free. WUDI is the released data-free consensus merge, scored without adaptation. The rightmost column is the eight-task average; bold marks the column maximum.}
\label{tab:fda}
\setlength{\tabcolsep}{2.0pt}
\footnotesize
\resizebox{\columnwidth}{!}{%
\begin{tabular}{@{}lccccccccc@{}}
\toprule
Method & MNIST & EurS & GTS & SVHN & DTD & RES & Cars & SUN & Avg \\
\midrule
WUDI & \textbf{99.44} & 98.00 & \textbf{96.55} & \textbf{95.30} & 73.62 & 90.13 & 81.58 & 73.13 & 88.47 \\
\midrule
FDA & 99.38 & 98.26 & 93.43 & 94.31 & 69.84 & 87.83 & 80.56 & 71.80 & 86.93 \\
\ours{}$\to$DARE & 98.8 & 97.4 & 94.2 & 91.0 & 73.5 & 91.7 & 77.7 & 71.7 & 86.99 \\
\ours{}$\to$TIES & 89.5 & 86.7 & 60.3 & 73.3 & 59.3 & 84.7 & 78.3 & 72.3 & 75.55 \\
WUDI-init FDA & 99.40 & \textbf{98.70} & 96.32 & 94.58 & 74.26 & 91.11 & 82.69 & 74.11 & 88.90 \\
\midrule
\ours{} & 99.07 & 98.44 & 96.26 & 92.16 & \textbf{77.23} & \textbf{93.22} & \textbf{84.25} & \textbf{74.97} & \textbf{89.45} \\
\bottomrule
\end{tabular}}
\end{table}

The comparison is not matched in data or compute: FDA adapts the merged model on synthetic anchors, while \ours{} uses no images and no adaptation. Against plain FDA, \ours{} is ahead on six of eight datasets. FDA leads on MNIST (by 0.31) and SVHN (by 2.15). \ours{}'s margin is largest on DTD (77.23 versus 69.84) and RESISC45 (93.22 versus 87.83), where synthetic anchors may capture coarse class structure but not fine distinctions. The \ours{}$\to$TIES row applies TIES-style trimming and sign election after the \ours{} projection, and averages 75.55 (\ours{}$\to$DARE, the analogous cascade with DARE, averages 86.99); its weakest datasets are GTSRB (60.3) and DTD (59.3). WUDI's data-free merge reaches 88.47, and starting the same adaptation from it adds 0.43 points (88.90). Against this stronger variant, WUDI-init FDA leads on MNIST (by 0.33), EuroSAT (by 0.26), GTSRB (by 0.06) and SVHN (by 2.42), while \ours{} leads on DTD (by 2.97), RESISC45 (by 2.11), Cars (by 1.56) and SUN397 (by 0.86).

\paragraph{Post-merge adaptation with real samples.}
The FDA rows adapt on synthetic anchors and never see a real image. As a data-using reference point, we take the \ours{} merge at $r{=}16$ and distill the eight fine-tuned experts into it: the student is the merged model with every parameter trainable, the teachers are the eight expert encoders with their frozen heads, and the loss is a KL divergence to the teachers' soft targets (temperature 2) on $4\times64 = 256$ real training images per task, drawn from each task's held-out-from-test split. Training uses AdamW at learning rate $10^{-6}$ with a cosine schedule for 600 steps, and accuracy is evaluated at 200, 400 and 600 steps on the official test sets. Table~\ref{tab:kd} gives the result on all three backbones.

\begin{table}[t]
\centering
\caption{\textbf{Post-merge expert distillation} (\%). Student initialized from the data-free \ours{} merge at $r{=}16$, then trained on 256 real images per task against the eight expert teachers. All numbers are our own runs on the same evaluator as Table~\ref{tab:main}. CE is cross-entropy on ground-truth labels instead of distillation. The ViT-B/32 and ViT-L/14 rows use the TALL release; the B/32 student starts from the re-implementation-harness merge (82.91; Appendix~\ref{app:setup}).}
\label{tab:kd}
\setlength{\tabcolsep}{3.5pt}
\footnotesize
\begin{tabular}{@{}lcccc@{}}
\toprule
Backbone & \ours{} init & $+$distill (600) & $+$CE (600) & $+$distill (1800) \\
\midrule
ViT-B/16 & 89.45 & \textbf{91.17} & 90.85 & 91.40 \\
ViT-B/32 & 82.91 & \textbf{87.95} & 87.77 & --- \\
ViT-L/14 & 90.87 & \textbf{92.62} & 92.57 & --- \\
\bottomrule
\end{tabular}
\end{table}

Distillation lifts every backbone, by $+1.72$ (B/16), $+5.04$ (B/32) and $+1.75$ (L/14) points, and the B/16 run improves further to 91.41 at 1200 steps and 91.40 at 1800, so 600 steps is not saturating the budget. Two qualifications keep this out of the head-to-head comparison. First, the gain is mostly supervised learning, not the teachers: replacing the KL term with plain cross-entropy on the same images gives 90.85 (B/16), 87.77 (B/32) and 92.57 (L/14), within 0.32 of the distillation rows. Second, the method uses real training-split images (and, in the CE variant, their labels), which the data-free baselines of Table~\ref{tab:main} do not; it is reported as the ceiling a small amount of real data reaches from this initialization, not as a training-free merge. Read together with the FDA rows, the comparison is consistent: synthetic adaptation moves the merge by at most 0.43 points and stays below it (88.90), whereas 256 real images per task move it by 1.7--5.0 points. A supervised signal on a few hundred images reduces the remaining error further.

\paragraph{Does the starting point matter?} Running the same recipe on ViT-B/16 (600 steps) from other data-free merges, a better start stays ahead but the gap narrows. From a WUDI merge at ridge $3\times10^{-3}$ (88.75; a different setting from Table~\ref{tab:main}) distillation reaches 90.71; from AlphaMerging ($r_1{=}16$, $k{=}64$; 87.11) it reaches 90.31; from unscaled Task Arithmetic (26.54) only 77.98. \ours{}'s 0.70-point lead over the WUDI start shrinks to 0.46 after distillation, so most of the distillation gain does not depend on the initialization, but among the starting points we tried \ours{} ends highest.

\section{Where the Margin Comes From, Per Task}
\label{app:gain}
\begin{figure}[t]
\centering
\begin{tikzpicture}[font=\scriptsize,x=1cm,y=1cm]

\begin{scope}[shift={(1.62,0)}]
  \draw[gray!45,line width=0.4pt] (0,-0.24) -- (0,2.34);
  \foreach \v in {0,10,20}{
    \draw[gray!45,line width=0.4pt] ({0.125*\v},-0.24) -- ({0.125*\v},-0.30);
    \node[font=\fontsize{5.4}{6}\selectfont,text=black!65,anchor=north] at ({0.125*\v},-0.32) {\v};}
  \node[font=\fontsize{5.6}{6.4}\selectfont,text=black!70,anchor=north] at (1.62,-0.54) {points};
  \foreach \i/\name/\val in {%
    0/GTSRB/24.40, 1/DTD/19.25, 2/EuroSAT/15.66, 3/Cars/13.16,
    4/RESISC45/12.12, 5/SUN397/8.27, 6/SVHN/1.81, 7/MNIST/0.36}{
    \pgfmathsetmacro{\y}{(7-\i)*0.30}
    \pgfmathsetmacro{\w}{0.125*\val}
    \fill[oursblue!80] (0,{\y}) rectangle ({\w},{\y+0.19});
    \node[font=\fontsize{5.4}{6}\selectfont,text=oursblue!85!black,anchor=west]
      at ({\w+0.04},{\y+0.095}) {\val};
    \node[font=\fontsize{5.8}{6.6}\selectfont,anchor=east] at (-0.10,{\y+0.095}) {\name};}
  \node[font=\fontsize{5.8}{7}\selectfont\bfseries,text=black!80,anchor=south] at (1.40,2.44)
    {vs.\ best entrywise};
\end{scope}

\begin{scope}[shift={(5.30,0)}]
  \draw[gray!45,line width=0.4pt] ({0.36*(-2.2)},-0.24) -- ({0.36*5.2},-0.24);
  \draw[gray!70,line width=0.5pt] (0,-0.24) -- (0,2.34);
  \foreach \v in {-2,0,2,4}{
    \draw[gray!45,line width=0.4pt] ({0.36*\v},-0.24) -- ({0.36*\v},-0.30);
    \node[font=\fontsize{5.4}{6}\selectfont,text=black!65,anchor=north] at ({0.36*\v},-0.32) {\v};}
  \node[font=\fontsize{5.6}{6.4}\selectfont,text=black!70,anchor=north] at (0.60,-0.54) {points};
  \foreach \i/\val in {%
    0/2.30, 1/0.16, 2/1.66, 3/0.98, 4/1.06, 5/-0.47, 6/-1.77, 7/-0.20}{
    \pgfmathsetmacro{\y}{(7-\i)*0.30}
    \pgfmathsetmacro{\w}{0.36*\val}
    \ifdim\w pt>0pt
      \fill[baseorange!85] (0,{\y}) rectangle ({\w},{\y+0.19});
      \node[font=\fontsize{5.4}{6}\selectfont,text=baseorange!80!black,anchor=west]
        at ({\w+0.04},{\y+0.095}) {\val};
    \else
      \fill[basered!55] ({\w},{\y}) rectangle (0,{\y+0.19});
      \node[font=\fontsize{5.4}{6}\selectfont,text=basered!80!black,anchor=east]
        at ({\w-0.04},{\y+0.095}) {\val};
    \fi}
  \node[font=\fontsize{5.8}{7}\selectfont\bfseries,text=black!80,anchor=south] at (0.60,2.44)
    {vs.\ best subspace};
\end{scope}

\end{tikzpicture}
\caption{\textbf{Where the margin comes from.} Per-task difference between \ours{} ($r{=}16$) and a competing method, on our ViT-B/16 checkpoints (Table~\ref{tab:main}). Left: against TIES-Merging, the strongest entrywise baseline on average; the gain is small only on the two tasks the entrywise rule already handles well (MNIST, SVHN). Right: against the best of TSV-M, Iso-C, Iso-CTS and AlphaMerging on each task, a much tighter comparison; \ours{} leads on five of eight tasks and trails on three, by at most 1.77 points, and the mean per-task advantage is 0.47 points. The right-panel opponent differs from task to task, so its mean is not a difference of two table averages.}
\label{fig:gain}
\end{figure}
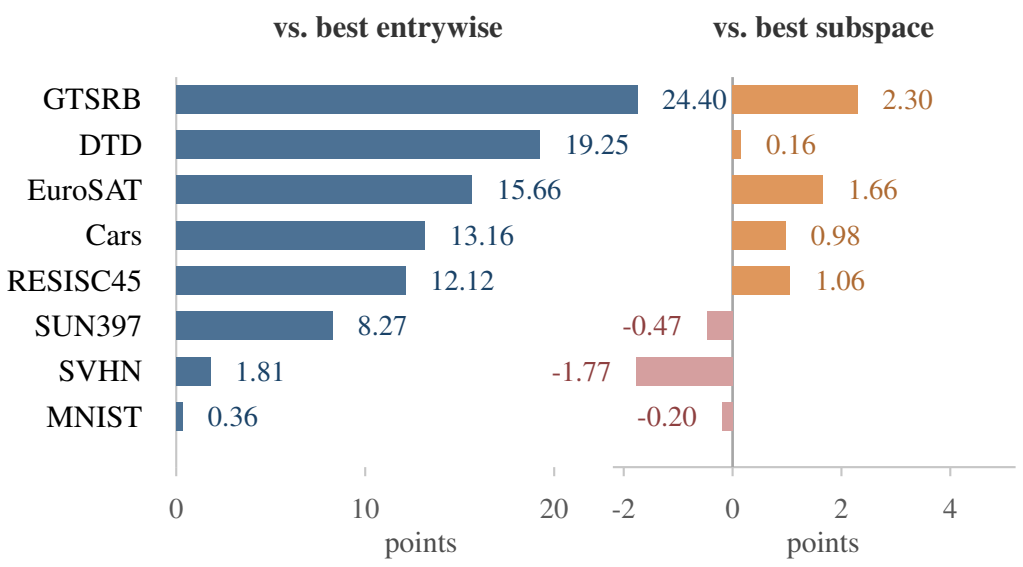

Table~\ref{tab:main} reports averages; Figure~\ref{fig:gain} gives the per-task version. The left panel is the gap to TIES-Merging, the strongest entrywise baseline on average; the right is the gap to the strongest of the four subspace baselines (TSV-M, Iso-C, Iso-CTS, AlphaMerging) on each task, so the second margin is measured against a different opponent in each row and is not a difference of two averages. Measured that way the mean per-task advantage is 0.47 points, against the 2.23 difference between \ours{}'s average and the best subspace average (TSV-M and AlphaMerging, 87.22); the larger figure is the right one for comparing against any single method, the smaller one for the claim that \ours{} leads the subspace family task by task. On DTD the strongest subspace baseline is AlphaMerging (77.07 against 77.23 for \ours{}); excluding AlphaMerging, the DTD margin would be 4.62 and the mean 1.02.

\section{Published versus Reproduced Subspace Baselines}
\label{app:repro}

Table~\ref{tab:repro} records, for each published subspace baseline, the value reported by its authors, our re-implementation on our own ViT-B/16 checkpoint, and our re-implementation on the TALL ViT-B/16 release. For TSV-M, Iso-C and Iso-CTS the two reproduction columns agree to within 0.31 points, which is why Section~\ref{sec:repro_gap} rules out the checkpoint as the source of the discrepancy for these three. AlphaMerging was run on TALL only at its best own-checkpoint setting ($r_1{=}32$, $k{=}64$), without re-sweeping; at that setting it scores 87.22 on our checkpoint and 85.40 on TALL.

\begin{table}[t]
\centering
\caption{\textbf{Published versus reproduced scores}, eight-task ViT-B/16 average accuracy (\%). ``Own'' and ``TALL'' are the same operator run on two independent checkpoint releases; ``Pub.'' is the number quoted from the original paper: TSV-M from~\cite{gargiulo2025tasksingular} and AlphaMerging from~\cite{bazarvaani2026alphamerging}. Iso-C and Iso-CTS~\cite{marczak2025isocts} have no published value in our table. We did not generate the published values. $^\ast$AlphaMerging is run at the best setting of its own-checkpoint sweep ($r_1{=}32$, $k{=}64$) on both releases. All methods in the table are data-free.}
\label{tab:repro}
\setlength{\tabcolsep}{3pt}
\small
\begin{tabular}{@{}lccccc@{}}
\toprule
Method & Data & Own & TALL & Pub. & TALL$-$Pub. \\
\midrule
TSV-M     & --  & 87.22 & 87.25 & 89.01 & $-1.76$ \\
AlphaMerging & -- & 87.22$^\ast$ & 85.40$^\ast$ & 89.9 & $-4.50$ \\
Iso-C     & --  & 87.15 & 87.46 & ---   & --- \\
Iso-CTS   & --  & 86.59 & 86.62 & ---   & --- \\
\midrule
\ours{}   & --  & \textbf{89.45} & \textbf{89.20} & --- & --- \\
\bottomrule
\end{tabular}
\end{table}

\paragraph{Reading of the table.} Two things should be kept apart. The first is that our harness is internally consistent: TSV-M, Iso-C and Iso-CTS score within a third of a point on two checkpoint releases produced by different pipelines. The second is that this consistency does not extend to the published values. For TSV-M we do not know why the numbers differ, and we do not assert a cause; the candidate explanation (the authors' own fine-tuning recipe) is recorded in Section~\ref{sec:repro_gap} as speculation, not as a finding. AlphaMerging's 89.9 is quoted from its paper; its TALL run, at the best own-checkpoint setting, falls 1.82 points below the same setting on our checkpoint, so for it the checkpoint is not ruled out.

\paragraph{What we did not do.} We did not substitute a different setting and report it as the published one, and we did not drop the baselines from Table~\ref{tab:main} on the grounds that they were inconvenient to reproduce. Every row in that table is a number our code produced under one protocol, with the reproduction status marked.

\section{Language-Model Details}
\label{app:nlu}

Table~\ref{tab:nlu_full} gives the per-task numbers behind Section~\ref{sec:negative}. RoBERTa-large~\cite{liu2019roberta} is a BERT-style~\cite{devlin2019bert} encoder with 24 transformer layers.

\begin{table}[t]
\centering
\caption{\textbf{Per-task GLUE results} for the language experiment (RoBERTa-large). Columns are the four configurations discussed in Section~\ref{sec:negative}, all on the same 146 tensors and with no merge coefficient ($\lambda{=}1$): Joint is \ours{}, Direct the unscaled sum of task vectors (task arithmetic at $\lambda{=}1$), Global the shared-basis control, and CR the consensus--residual variant at $c{=}8$. Metrics ($\times100$): Matthews correlation for CoLA, the mean of accuracy and F1 for MRPC and QQP, the mean of Pearson and Spearman correlation for STS-B, and accuracy otherwise; a negative STS-B value means the merged model's predictions are anticorrelated with the gold scores.}
\label{tab:nlu_full}
\setlength{\tabcolsep}{4pt}
\footnotesize
\begin{tabular}{@{}lcccc@{}}
\toprule
Task & Joint & Direct & Global & CR ($c{=}8$) \\
\midrule
CoLA   & 6.6  & $-2.1$ & 47.7 & 53.8 \\
SST-2  & 51.5 & 50.1 & 91.9 & 93.6 \\
MRPC   & 22.3 & 59.0 & 75.8 & 75.7 \\
STS-B  & $-13.9$ & $-11.8$ & 84.0 & 78.1 \\
QQP    & 49.3 & 80.4 & 21.7 & 31.5 \\
MNLI   & 38.6 & 44.3 & 49.0 & 70.4 \\
QNLI   & 49.4 & 48.2 & 70.1 & 82.8 \\
RTE    & 47.7 & 49.8 & 66.1 & 69.7 \\
\midrule
Avg.   & 31.4 & 39.8 & 63.3 & \textbf{69.5} \\
\bottomrule
\end{tabular}
\end{table}

Two details are worth noting. First, the joint projection is not uniformly bad on language: on QQP it reaches 49.3, against 21.7 for the global control, although it stays below the 80.4 of the unscaled direct sum; on MRPC, by contrast, it loses 36.7 points to the direct sum (22.3 versus 59.0). Relative to the global control the joint projection loses most on STS-B, MRPC, CoLA and SST-2 (40--98 points) and least on MNLI (10.4); QQP is the only task on which it is higher. We do not test why. Second, the consensus--residual repair changes the pattern: it improves over the global control on CoLA, SST-2, QQP, MNLI, QNLI and RTE, matches it on MRPC (75.7 versus 75.8) and is lower on STS-B (78.1 versus 84.0); on QQP it rises only from 21.7 to 31.5, still far below the joint projection's 49.3. We have no explanation for QQP's behaviour and flag it as unresolved.

\paragraph{Consensus--residual decomposition.}
For every two-dimensional backbone tensor with smallest dimension at least 64 (the 144 attention and FFN projection matrices plus the word and position embeddings; LayerNorm, biases and token-type embeddings stay at $\btheta_0$, and each task keeps its own classifier head), let $\mV_c\in\R^{n\times c}$ hold the top-$c$ right-singular vectors of the summed delta $\sum_{j=1}^{N}\btau_j$ and $\mP_c=\mV_c\mV_c^\top$. Each task vector splits exactly as $\btau_i=\btau_i\mP_c+\delta\btau_i$, with $\delta\btau_i=\btau_i(\mI-\mP_c)$. The shared component is the unweighted mean of the per-task consensus parts, $\btau^{c}=\tfrac{1}{N}\sum_i\btau_i\mP_c$. Let $\mV^{\delta}_j$ hold the top-$r$ ($r{=}16$) right-singular vectors of $\delta\btau_j$, and let $\mQ_i$ be an orthonormal (QR) basis of $[\mV^{\delta}_j]_{j\ne i}$. Each residual is filtered as $\tilde{\delta\btau}_i=\delta\btau_i(\mI-\mQ_i\mQ_i^\top)$, and CR--\ours{} merges as $\btheta=\btheta_0+\btau^{c}+\sum_i\tilde{\delta\btau}_i$, with the consensus added once and no merge coefficient. Because every $\delta\btau_j$ is orthogonal to $\mV_c$, the filter leaves the consensus directions untouched. The sweep over $c\in\{2,4,8,16,32\}$ gives 65.52, 66.94, 69.45, 68.13 and 66.97. The peak is at $c{=}8$; $c{=}16$ is 1.32 points lower, $c{=}4$ and $c{=}32$ lose 2.5, and $c{=}2$ loses 3.9.

\paragraph{Why the repair is not a contribution.}
CR--\ours{} introduces a second hyperparameter $c$, requires a rank sweep to find, and is specific to settings where a large consensus component exists. It improves on the shared-basis control on average (69.45 versus 63.28) but not on every task: it is lower on STS-B (78.1 versus 84.0), and on QQP it rises only from 21.7 to 31.5, still 48.9 points below the unscaled direct sum (80.4). We report it because it bounds the failure diagnosis: separating a consensus component accounts for part, but not all, of the language result.

\section{Scope, Threats to Validity, and Negative Evidence}
\label{app:limits}

Section~\ref{sec:negative} and the Conclusion state the limitations briefly. This appendix expands them and records what we could not establish. We separate three kinds of shortfall: limits on the \emph{scope} of what was tested, threats to the \emph{validity} of the conclusions within that scope, and \emph{negative} results that bound how far the geometric story can be pushed.

\subsection{Scope}

\paragraph{Task type.}
Every positive result in this paper is image classification with a per-task linear head. The method never merges the heads, so a merged model has no single classifier; evaluating it requires knowing which task an input belongs to, which is the standard protocol in this literature but is also an assumption a deployment would have to satisfy. Nothing in the projection depends on the head being linear, but nothing we measured exercises a head that is not.

\paragraph{Objective.}
All eight experts are CLIP-initialized encoders fine-tuned with cross-entropy on a labelled training set. We did not test contrastive fine-tuning, self-supervised adaptation, LoRA-style low-rank experts, or experts obtained by continued pretraining. This matters for interpretation: one untested reading of our results is that the directions removed in the vision encoder are less essential to other tasks than those removed in the language encoder; Appendix~\ref{app:privacy} shows that the vision deltas themselves are substantially shared, not mostly private, and we have not tested whether this reading depends on the supervised cross-entropy objective. The language result of Section~\ref{sec:negative} suggests the vision behaviour is not universal across objectives and architectures, and we have no controlled experiment isolating which factor is responsible.

\paragraph{Scale.}
The largest model is ViT-L/14 with 101 merged tensors. We did not test models where the merged tensors are numerous enough for the per-tensor SVD cost to dominate---the method is closed-form but not free, and at large scale the $O(KNmn^2)$ cost may become a practical objection even though it is small at the scales studied here. We also did not test models whose hidden dimension is not rectangular (e.g.\ factorized attention or mixture-of-experts layers), where the row-space argument of Section~\ref{sec:interference} would need restating.

\paragraph{Data-freedom is exact but narrow.}
The merge uses no inputs, labels, or forward passes. It does use the fine-tuned checkpoints themselves, and therefore any information that fine-tuning absorbed from its training data. ``Data-free'' here means no access to the data \emph{at merge time}, not that the merge is independent of the data's influence; a reader who interprets it the stronger way would be overcrediting the method.

\subsection{Threats to Validity}

\paragraph{Single training run per task.}
The eight experts come from one fine-tuning run each. The two-release comparison in Appendix~\ref{app:ci} substitutes for a seed study, but two releases are not a distribution: they share the same data, the same evaluation splits, and much of the same pipeline. Variance of a point or two would leave the 11.88-point gap to TIES intact, but not the 0.98-point gap to WUDI or the 2.23-point gap to the subspace re-implementations, and would make the 2.94-point joint-versus-shared ablation less secure than it reads. We did not run the fine-tuning sweep that would settle this, and we flag it as the largest unaddressed gap in the evidence.

\paragraph{The evaluation split.}
Each dataset is evaluated on the split distributed with the model release. We did not re-split, and we cannot rule out that these splits were used during model selection elsewhere. The baselines share the split and the protocol, so comparisons within Table~\ref{tab:main} are matched, but absolute numbers may not transfer to a different split of the same data.

\paragraph{Rank selection.}
We use $r{=}16$ everywhere, on the basis of the sweep in Appendix~\ref{app:rank}. The plateau is broad---89.30, 89.45, 89.22 at $r{=}12,16,20$---so the choice is not delicate on this task set, but it was still chosen by looking at merged accuracy on the eight tasks, which is the quantity being reported. A genuinely held-out task set would be the clean check, and we did not have one.

\paragraph{Baseline tuning.}
We swept the merge coefficient for every training-free baseline and report each one's peak, which is generous to them. We did not sweep the baselines' other hyperparameters exhaustively: TIES density is fixed at $0.2$ and DARE's drop rate at $p{=}0.9$. A full joint sweep might move a baseline's peak by a point or two. The margin to the strongest parameter-wise baseline is 11.88 points, so such a change would not affect that ordering; the 0.98-point margin to WUDI's data-free consensus merge is not secure against a change of that size.

\paragraph{What the shared-basis control does and does not isolate.}
The control projects every task delta against one basis, the top-$r$ right singular vectors of the summed delta. It matches the joint merge in rank, tensor set, and projection algebra, and differs only in whose subspace is removed. That is the right control for the claim that per-task peer unions matter. It is not a control for the choice of SVD at all: both variants use the same rank-$r$ right singular basis, so the comparison cannot tell us whether a different low-rank summary---random projections, gradient-based subspaces, or a learned basis---would do better. The ablations in the main text are a comparison \emph{within} the SVD family.

\subsection{Negative Evidence}

\paragraph{Entrywise baselines are not weak at what they measure.}
TIES reaches 90.35 on SVHN and 98.71 on MNIST---within two points of \ours{}; on MNIST the gap is only 0.36, though the paired test still separates them ($p{=}7\times10^{-5}$, Appendix~\ref{app:mcnemar}). TIES loses most on GTSRB (24.4), DTD (19.3) and EuroSAT (15.7), and this per-task deficit does not track a task's mean subspace overlap with the others ($r{=}-0.27$ across the eight tasks, Figure~\ref{fig:overlap}). We therefore do not claim that entrywise methods fail where deltas overlap most.

\paragraph{The projection is not universally beneficial.}
At $r{=}32$ the joint merge falls to 87.08, below its own $r{=}12$ result, and on RESISC45 at that rank the shared-basis control is ahead by 0.33 points. Projection removes real signal as well as interference, and the method has a failure mode at large rank that we characterize but do not fix. A rank-selection rule that adapts per task---for instance by measuring per-task retained energy and stopping when it drops below a threshold---would be a natural extension; we did not implement one, and the sweep suggests the threshold would be task-dependent.

\paragraph{Language is a genuine failure.}
Section~\ref{sec:negative} reports that the method is worse than the baselines it was meant to improve on for RoBERTa-large. The consensus--residual repair recovers the merge to 69.45, but is still 48.9 points below direct task arithmetic on QQP (31.5 versus 80.4), and its consensus rank $c$ is itself swept. We report it as a repair, not a contribution, and we do not know whether a decomposition of this kind exists for architectures where no single dominant consensus direction can be found.

\paragraph{What we would need to be wrong about.}
The mechanism claim is that the gain comes from removing cross-task interference in input space, not from norm reduction, sparsity, or incidental regularization. Two pieces of evidence support it: the shared-basis control removes a comparable fraction of the delta norm and scores three points lower (Appendix~\ref{app:retention}); and sequential projection, which also removes overlap but asymmetrically, collapses (Appendix~\ref{app:order}). The baselines' per-task deficit does not track subspace overlap (first paragraph of this subsection), so it is not offered as support. None of these rules out the possibility that some \emph{other} property correlated with cross-task overlap---for example, that overlapping tasks also have noisier fine-tuning runs---produces the same pattern. A controlled experiment that corrupted one task's delta with isotropic noise, holding its overlap fixed, would separate these; we did not run it.

\section{Runtime, Artifacts, and Reproducibility}
\label{app:runtime}

\paragraph{Compute.}
The merge computes a thin SVD of each task's delta for every merged tensor, followed by a QR of the stacked neighbour bases and a projection. On CPU it takes about 61 seconds for the 53 tensors of ViT-B/16 and does not touch a GPU; the corresponding figures for ViT-B/32 and ViT-L/14 were not measured. Evaluation is the only GPU step. The full comparison in Table~\ref{tab:main} therefore requires no training, no fine-tuning, and no data loading during the merge itself.

\paragraph{Artifacts retained.}
The merged weight files of the ViT-B/16 rank sweep are retained, named by variant and rank (\texttt{merged\_joint\_r8.pt} through \texttt{merged\_joint\_r32.pt}, and \texttt{merged\_global\_r8.pt} through \texttt{merged\_global\_r32.pt}); the sequential file is \texttt{merged\_sequential\_r8.pt}.

\paragraph{Scripts.}
The vision merges take the rank, the variant, and the model and data locations as arguments, write the merged checkpoint, and evaluate it on the eight test splits. The re-implemented subspace baselines of Tables~\ref{tab:main} and~\ref{tab:backbones} share one merge-and-evaluate harness with the merge rule as an argument, so they share the evaluator and tensor scope (Appendix~\ref{app:setup}). The language results use a separate driver.

\paragraph{Numerical audit of the closed form.}
The claims of Section~\ref{sec:method} are algebraic, so they can be checked against the artifacts rather than argued. We ran four checks on ViT-B/16 at $r{=}16$, reported in Table~\ref{tab:audit}.

\begin{table}[h]
\centering
\caption{\textbf{Numerical audit of the closed-form merge} (ViT-B/16, $r{=}16$). Relative errors are Frobenius norms over the 53 merged tensors unless a tensor is named.}
\label{tab:audit}
\setlength{\tabcolsep}{4.0pt}
\footnotesize
\begin{tabular}{@{}lr@{}}
\toprule
Check & Result \\
\midrule
Recomputed \texttt{global} merge vs.\ archived file & $2.8\times10^{-6}$ \\
\quad same, worst tensor & $8.4\times10^{-6}$ (abs.) \\
Aggregate retention, joint $r{=}16$ (target $0.3036$) & $0.3036$ \\
Wall-clock, 53 tensors, CPU (target $61$ s) & $61.0$ s \\
\midrule
Permutation of the task list, three orders & $1.9$, $1.7$, $1.7\times10^{-3}$ \\
\quad worst tensor (\texttt{token\_embedding}) & $2.3\times10^{-2}$ \\
\quad attention/MLP tensors only & $\le 1.3\times10^{-6}$ \\
\quad same test repeated in float64 & $\le 3.3\times10^{-11}$ \\
Sign flip, $\btau_i \mapsto -\btau_i$ & $7.9\times10^{-8}$ \\
\bottomrule
\end{tabular}
\end{table}

The first block checks that the code reproduces the artifact. Re-running the shared-basis variant from the eight task vectors reproduces the archived \texttt{merged\_global\_r16.pt} file to $2.8\times10^{-6}$ relative error over all 53 tensors, with a worst-tensor absolute deviation of $8.4\times10^{-6}$. The difference is float32 storage rounding in the file, not a discrepancy in the merge; the retained-norm ratio and the wall-clock time reproduce the reported figures exactly.

The second block checks Proposition~\ref{prop:invariance}, and it is worth being precise about what it shows. Permuting the task list changes nothing mathematically---$\mU^{(i)}_\cup$ depends on the \emph{set} of neighbours---but it changes the column order fed to the QR, and therefore the floating-point value of the projector. Across three permutations the merged delta moves by $1.7$--$1.9\times10^{-3}$ relative. The movement is not uniform across tensors. Attention and MLP tensors agree to $\le 1.3\times10^{-6}$, while three tensors---the token embedding ($2.3\times10^{-2}$), the learned positional embedding ($1.8\times10^{-2}$), and the text projection ($1.6\times10^{-3}$)---carry essentially all of it. Repeating the identical test in float64 drives every one of them below $3.3\times10^{-11}$, which identifies the cause as float32 roundoff rather than an asymmetry in the merge.

The roundoff is amplified in exactly the tensors where the projection is least stable, for two compounding reasons. First, their rank-16 cutoff sits on a near-tied singular value: the ratio $s_{15}/s_{16}$ of the task's own delta spectrum is $1.003$ for the token embedding and $1.004$ for the text projection, so a rounding-level perturbation of $\Delta_i$ can swap which direction enters $\mathcal{S}_i$. Second, the positional embedding's eight deltas cancel strongly: the merged delta retains only $0.052$ of the summed norms, so an absolute error of fixed size becomes a large relative error after cancellation. We did not re-evaluate accuracy for each order; the merged deltas differ by at most $1.9\times10^{-3}$ in relative terms over all tensors. It does mean that a bitwise-identical merge should not be expected across task orders, and a reader reproducing the numbers should compare accuracies, not weights.

The sign-flip check is exact to float roundoff: negating one task vector shifts the merged delta by $-2\btau_i(\mI-\mU^{(i)}_\cup\mU^{(i)\top}_\cup)=-2\tilde\btau_i$ to $7.9\times10^{-8}$ relative, confirming that a task's contribution enters linearly through its own vector and not through its basis.

\paragraph{What is not reproduced here.}
Two numbers in the literature could not be reproduced from released artifacts and are therefore not used as comparison points: a task-arithmetic score of 73.94 sometimes cited for this task set, and the per-dataset Task Arithmetic columns of the TALL paper, which use a different evaluation split. All numbers in the \emph{Avg} columns are our own runs under the protocol of Appendix~\ref{app:setup}; the \emph{Pub.} column of Table~\ref{tab:main} quotes published values.


\end{document}